\documentclass[runningheads]{llncs}

\usepackage[final,year=2026]{eccv} 

\usepackage{eccvabbrv}
\usepackage{graphicx}
\usepackage{booktabs}
\usepackage{tikz}
\usetikzlibrary{positioning,spy}
\usetikzlibrary{calc}
\usepackage{svg}
\usepackage{seqsplit}
\usepackage{amsmath}
\usepackage{amssymb}
\usepackage{textcomp}
\usepackage[accsupp]{axessibility}
\usepackage[breaklinks,colorlinks,citecolor=eccvblue]{hyperref}

\newlength{\rowimgheight}
\newlength{\inputimgheight}
\newlength{\imageheight}
\title{SplatGuide: Geometric Priors from 3D Gaussians for Pose-Free Novel View Synthesis}
\titlerunning{SplatGuide}

\author{Yejun Zhang\inst{1}$^{\star}$ \and
Zihan Wang\inst{1}$^{\star}$ \and
Xu Ji\inst{1}$^{\star}$ \and
Yihao Wang\inst{1} \and
Yuxin Hou\inst{2} \and
Junyuan Fang\inst{1} \and
Juho-Matti Kilpel\"ainen\inst{1} \and
Arno Solin\inst{1, 3} \and
Hamed Rezazadegan Tavakoli\inst{4} \and
Esa Rahtu\inst{5} \and
Juho Kannala\inst{1, 6}}
\authorrunning{Y.\ Zhang et al.}
\institute{
$^1$Aalto University, Finland \enspace $^2$Deep~Render, UK \enspace $^3$ELLIS Institute Finland \enspace
$^4$Nokia~Technologies, Finland \enspace $^5$Tampere~University, Finland \enspace $^6$University of Oulu, Finland \\
\email{\{firstname.lastname, zihan.1.wang, xu.1.ji\}@aalto.fi} \quad
\email{hamed.rezazadegan\_tavakoli@nokia.com} \\
\email{esa.rahtu@tuni.fi} \quad
\email{yuxin.hou@deeprender.ai}
}
\begin{document}
\maketitle
{\renewcommand{\thefootnote}{}\footnotetext{$^{\star}$~Equal contribution.}}

\begin{center}
\usetikzlibrary{arrows.meta}
\definecolor{tred}{RGB}{230,35,30}
\definecolor{tblue}{RGB}{30,55,200}
\definecolor{tviolet}{RGB}{126,91,218}
\definecolor{tgreen}{RGB}{60,165,85}
\definecolor{recpeach}{RGB}{252,226,200}
\definecolor{gsblue}{RGB}{53,142,235}
\definecolor{gsorange}{RGB}{255,132,48}
\definecolor{gspink}{RGB}{238,91,164}
\definecolor{gsgreen}{RGB}{78,193,119}

\tikzset{
  frus/.pic={
    \draw[pic actions] (0,0) -- (-0.20,0.38) (0,0) -- (0.02,0.50)
                       (0,0) -- (0.24,0.42) (0,0) -- (0.06,0.34);
    \draw[pic actions] (-0.20,0.38) -- (0.02,0.50) -- (0.24,0.42)
                       -- (0.06,0.34) -- cycle;
  },
  fatarrow/.style={-{Triangle[length=5pt,width=10pt]},line width=3.6pt},
}

\resizebox{0.8\textwidth}{!}{%
\begin{tikzpicture}[x=1cm,y=1cm]
  \path[use as bounding box] (0,0.85) rectangle (14.33,7.80);

  \node[font=\small\bfseries,text=tred]  at (1.95,7.60) {Images};
  \node[font=\small\bfseries,text=tblue] at (4.275,7.60) {Latent Image};

  \node[anchor=south west,inner sep=0pt,draw=gsblue,line width=0.9pt]
    at (0.85,6.04) {\includegraphics[width=1.35cm,height=1.35cm]{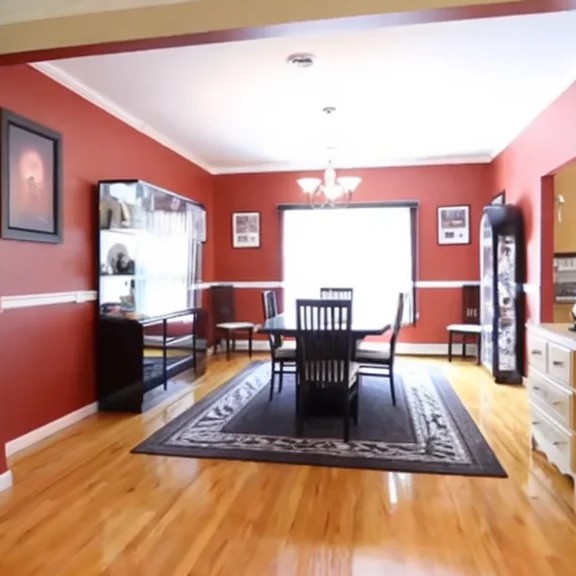}};
  \node[anchor=south west,inner sep=0pt,draw=gsorange,line width=0.9pt]
    at (1.27,5.82) {\includegraphics[width=1.35cm,height=1.35cm]{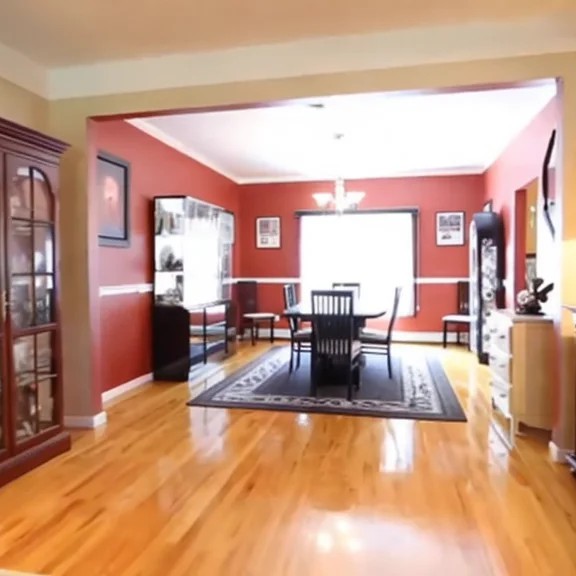}};
  \node[anchor=south west,inner sep=0pt,draw=gsgreen,line width=0.9pt]
    at (1.69,5.60) {\includegraphics[width=1.35cm,height=1.35cm]{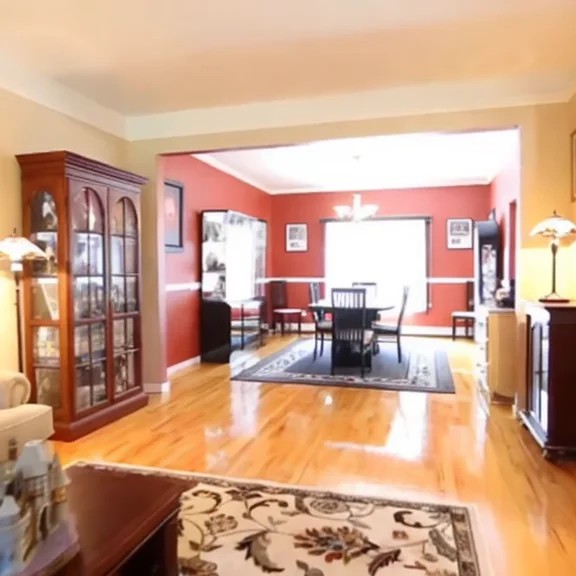}};
  \node[anchor=south west,inner sep=0pt,draw=black!30,line width=0.4pt]
    at (3.60,5.95) {\includegraphics[width=1.35cm,height=1.35cm]{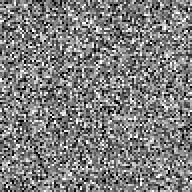}};

  \pic[tred,line width=0.55pt,rotate=8]    at (1.10,4.85) {frus};
  \pic[tred,line width=0.55pt,rotate=-8]   at (1.95,4.85) {frus};
  \pic[tred,line width=0.55pt,rotate=6]    at (2.80,4.85) {frus};
  \pic[tblue,line width=0.55pt,rotate=10]  at (4.275,4.85) {frus};

  \draw[fatarrow,tred]  (1.10,4.66) -- (1.10,4.30);
  \draw[fatarrow,tred]  (1.95,4.66) -- (1.95,4.30);
  \draw[fatarrow,tred]  (2.80,4.66) -- (2.80,4.30);
  \draw[fatarrow,tblue] (4.275,4.66) -- (4.275,4.30);

  \fill[black!15,rounded corners=2pt] (0,3.50) rectangle (5.65,4.25);
  \node[font=\normalsize\bfseries] at (2.825,3.875) {Multi-view Diffusion};

  \draw[fatarrow,tblue] (2.825,3.44) -- (2.825,3.08);

  \node[anchor=south west,inner sep=0pt,draw=black!30,line width=0.4pt]
    at (2.15,1.67) {\includegraphics[width=1.35cm,height=1.35cm]{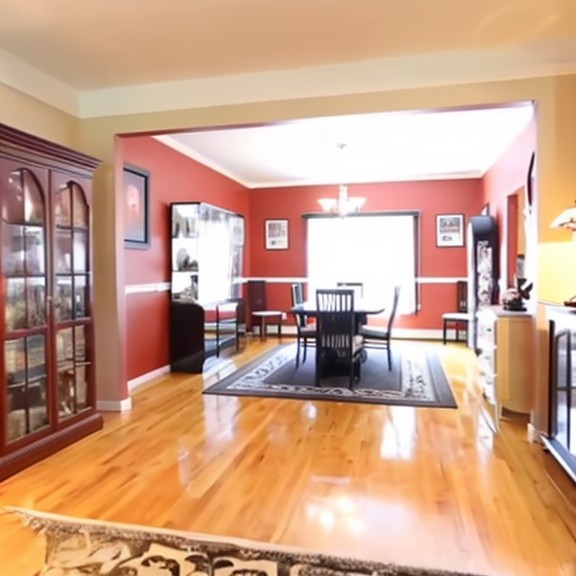}};

  \fill[blue!7] (0,0.95) rectangle (5.65,1.55);
  \node[font=\normalsize] at (2.825,1.25) {(a) Baseline};

  \draw[dashed,black!35,line width=0.5pt] (5.78,0.95) -- (5.78,7.65);

  \node[font=\small\bfseries,text=tred] at (7.30,7.60) {Images};

  \node[anchor=south west,inner sep=0pt,draw=gsblue,line width=0.9pt]
    at (6.15,6.25) {\includegraphics[width=1.15cm,height=1.15cm]{figures/scene3/input2.jpg}};
  \node[anchor=south west,inner sep=0pt,draw=gsorange,line width=0.9pt]
    at (6.51,6.03) {\includegraphics[width=1.15cm,height=1.15cm]{figures/scene3/input1.jpg}};
  \node[anchor=south west,inner sep=0pt,draw=gsgreen,line width=0.9pt]
    at (6.87,5.81) {\includegraphics[width=1.15cm,height=1.15cm]{figures/scene3/input.jpg}};

  \draw[fatarrow,tred] (8.10,6.67) -- (8.34,6.67);

  \draw[draw=black!45,fill=recpeach,line width=0.8pt,rounded corners=2pt]
    (8.37,5.95) rectangle (9.87,7.40);
  \node[font=\scriptsize\bfseries] at (9.12,6.67) {FF Recon.};
  \draw[fatarrow,tred] (9.90,6.67) -- (10.14,6.67);

  \draw[draw=black!35,fill=black!2,line width=0.5pt,rounded corners=2pt]
    (10.22,5.95) rectangle (11.67,7.40);
  \node[font=\small\bfseries] at (10.945,7.60) {3D-GS};
  \begin{scope}[shift={(10.945,6.675)},scale=0.55]
    \fill[gsblue,opacity=.66,rotate=22] (-0.62,0.55) ellipse (0.58 and 0.22);
    \fill[gsorange,opacity=.60,rotate=-12] (0.18,0.67) ellipse (0.66 and 0.24);
    \fill[gspink,opacity=.56,rotate=18] (-0.36,0.03) ellipse (0.70 and 0.25);
    \fill[gsgreen,opacity=.58,rotate=-31] (0.48,0.14) ellipse (0.61 and 0.23);
    \fill[gsblue,opacity=.66,rotate=-14] (0.43,-0.58) ellipse (0.64 and 0.24);
    \fill[gsgreen,opacity=.58,rotate=33] (-0.49,-0.77) ellipse (0.50 and 0.19);
  \end{scope}
  \draw[fatarrow,tblue] (11.75,6.67) -- (11.99,6.67);

  \node[font=\scriptsize\bfseries,text=tblue] at (13.20,7.58) {Guidance};

  \draw[draw=black!40,fill=black!1,line width=0.55pt,rounded corners=2pt]
    (12.07,5.95) rectangle (14.33,7.40);
  \node[anchor=west,font=\scriptsize] at (12.13,7.10) {\textbf{1}~View selection};
  \node[anchor=west,font=\scriptsize] at (12.13,6.675) {\textbf{2}~Rendering};
  \node[anchor=west,font=\scriptsize] at (12.13,6.25) {\textbf{3}~Tokens};

  \pic[tred,line width=0.55pt,rotate=6]    at (6.30,4.85) {frus};
  \pic[tred,line width=0.55pt,rotate=-6]   at (7.085,4.85) {frus};
  \pic[tred,line width=0.55pt,rotate=-10]  at (7.87,4.85) {frus};
  \pic[tblue,line width=0.55pt,rotate=8]   at (13.20,4.85) {frus};

  \draw[fatarrow,tred]  (6.30,4.66) -- (6.30,4.30);
  \draw[fatarrow,tred]  (7.085,4.66) -- (7.085,4.30);
  \draw[fatarrow,tred]  (7.87,4.66) -- (7.87,4.30);
  \draw[fatarrow,tblue] (13.20,4.66) -- (13.20,4.30);

  \fill[black!15,rounded corners=2pt] (5.90,3.50) rectangle (14.33,4.25);
  \node[font=\normalsize\bfseries] at (10.115,3.875) {Multi-view Diffusion};

  \draw[fatarrow,tblue] (10.115,3.44) -- (10.115,3.08);

  \node[anchor=south west,inner sep=0pt,draw=black!30,line width=0.4pt]
    at (9.44,1.67) {\includegraphics[width=1.35cm,height=1.35cm]{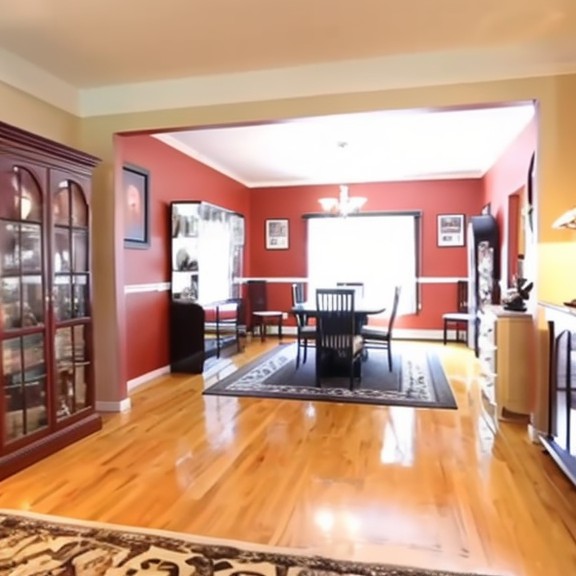}};

  \fill[blue!7] (5.90,0.95) rectangle (14.33,1.55);
  \node[font=\normalsize] at (10.115,1.25) {(b) \textbf{Ours}: GS-guided diffusion};
\end{tikzpicture}%
}

  \captionof{figure}{\textbf{SplatGuide} vs.\ diffusion baseline. (a) The baseline passes camera poses directly to a multi-view diffusion model. (b) SplatGuide first uses a feed-forward reconstruction (FF Recon.) model to build a 3DGS scene from unposed images, then reuses it to supply rendered images, visibility-aware view selection, and feature tokens as geometric guidance.}
  \label{fig:teaser}
\end{center}
\begin{abstract}
Generating photorealistic novel views from unposed images requires both 3D geometric understanding and the ability to synthesize unseen content.
A natural strategy combines feed-forward 3DGS reconstruction with multi-view diffusion. Yet prior pipelines extract at most one signal from the reconstruction, either pixel rendering or learned features, while none exploits per-Gaussian visibility for occlusion-aware reference selection.
This \emph{information disconnect} leaves renderable geometry, visibility cues, and learned features unused.
SplatGuide closes this disconnect by reusing a single 3DGS scene across three complementary roles.
Rendered images provide pixel-aligned geometric conditioning. Per-Gaussian source-view indices are rendered into a target-view voting map for occlusion-aware reference selection. Reconstruction tokens supply feature-level guidance via cross-attention.
All three signals derive from the same reconstruction forward pass.
Across RealEstate10K, DL3DV, Tanks-and-Temples, and Mip-NeRF~360, SplatGuide achieves state-of-the-art pose-free novel view synthesis. On RealEstate10K, with a moderate number of input views, it surpasses the ground-truth-pose baseline.
\end{abstract}

\section{Introduction}
\label{sec:intro}

Imagine synthesizing a photorealistic walkthrough of a room from a handful of casual phone snapshots, with no camera calibration, no controlled capture, and no pose annotations. This is the goal of pose-free novel view synthesis (NVS), and it demands two capabilities that today live in separate worlds. Feed-forward 3D reconstruction models~\cite{jiang2025anysplat,ye2024no,dust3r_cvpr24,wang2025vggt} recover camera poses and scene geometry in a single forward pass, yet they can only interpolate between observed viewpoints and cannot hallucinate content in unobserved regions. Multi-view diffusion models~\cite{cat3d,zhou2025stable,wang2024motionctrl,ren2025gen3c,shi2023mvdream,Cao_2025_CVPR,lu2025matrix3d,zhang2025spatialcrafter} generate photorealistic images at novel viewpoints, yet they depend on accurate, pre-computed poses, a requirement that is rarely met outside controlled benchmarks. Combining reconstruction with diffusion is therefore a natural strategy: reconstruction grounds the geometry, and diffusion synthesizes what the reconstruction cannot see.

Existing combinations, however, pass the reconstruction output to the generator through a single narrow channel.
Pose-only methods~\cite{zhou2025stable,cat3d,zhang2025spatialcrafter} forward the estimated camera parameters and discard the reconstructed scene entirely, causing novel-view quality to fall well below the ground-truth-pose ceiling.
Pixel-level methods~\cite{yu2024viewcrafter,mark2025trajectorycrafter,chan2023genvs,zhou2023sparsefusion,wu2024difix3d} render the reconstructed geometry into images that condition a video diffusion backbone, anchoring the spatial layout but restricting generation to trajectory interpolation and tying the pipeline to a specific architecture.
Feature-level methods~\cite{ren2025gen3c,wu2025geometryforcing} inject dense latent features via cross-attention or alignment losses, improving spatial consistency but requiring dense features from models like VGGT~\cite{wang2025vggt} at prohibitive computational cost.
Moreover, all of these pipelines select reference views by pose distance or temporal recency, heuristics that ignore occlusion and break down under tight context budgets.

The common blind spot is not the absence of a particular module, but a structural \emph{information disconnect}: feed-forward reconstruction already produces a complete 3D Gaussian Splatting (3DGS)~\cite{kerbl20233d} scene that encodes renderable geometry, per-Gaussian source-view ownership, and learned feature representations, yet existing pipelines extract at most one of these signals and discard the rest.
The cost of this disconnect surfaces at both ends of the pipeline. On the conditioning side, restoring the discarded renderings and features to a predicted-pose baseline markedly improves both fidelity and perceptual quality (\cref{tab:ablation_structure}). On the selection side, reference-view selection hinges on exactly the per-Gaussian visibility that existing pipelines throw away, and the choice of selection policy alone separates the strongest strategies from the weakest by a wide margin (\cref{tab:selection-main}). Pose-free NVS is thus bottlenecked not by what reconstruction fails to provide, but by what existing pipelines fail to use.

We present SplatGuide, which closes the information disconnect by reusing a single 3DGS scene across three complementary roles (\cref{fig:teaser}). Rather than inventing new modules, SplatGuide recovers information that the reconstruction already provides but that existing pipelines throw away:
(1)~\emph{Geometric signal}: rendering the 3DGS scene at target and reference poses produces pixel-aligned images that provide direct geometric conditioning for the diffusion model.
(2)~\emph{Visibility signal}: each Gaussian naturally records the index of its source view; rendering these indices into a target-view voting map yields an occlusion-aware reference selector that clearly outperforms pose-based and recency-based strategies under tight context budgets.
(3)~\emph{Feature signal}: camera and register tokens extracted from the reconstruction backbone are injected via cross-attention, supplying scene-level context that pixel-aligned renderings cannot convey.
All three signals derive from a single reconstruction forward pass, require no additional 3D primitives, and keep the backbone frozen.

Our contributions are:
\begin{itemize}
    \item We identify the information disconnect as the structural bottleneck of pose-free NVS and propose SplatGuide, which closes it by bridging rendered images, per-Gaussian visibility, and reconstruction tokens from a single reconstructed 3DGS scene to the diffusion generator.
    \item We introduce a visibility-aware view selector that renders per-Gaussian source-view indices into a target-view voting map, delivering occlusion-aware reference selection at negligible cost and large gains over pose-based and recency-based strategies under tight context budgets.
    \item We achieve state-of-the-art pose-free NVS on four benchmarks, surpassing the ground-truth-pose baseline on RealEstate10K given sufficient input views, and demonstrate modularity: the reconstruction backbone can be substituted zero-shot without retraining, so the framework benefits directly from advances in feed-forward reconstruction.
\end{itemize}

\section{Related Work}
\label{sec:related}

\paragraph{Pose-Free Novel View Synthesis.}
NeRF~\cite{mildenhall2021nerf} and 3DGS~\cite{kerbl20233d} achieve photorealistic novel view synthesis but require accurate camera poses, typically obtained from Structure-from-Motion (SfM)~\cite{schonberger2016structure} pipelines. Feed-forward reconstruction models remove this dependency: DUSt3R~\cite{dust3r_cvpr24}, MASt3R~\cite{mast3r}, and VGGT~\cite{wang2025vggt} jointly predict camera poses and 3D point clouds from input images, while InstantSplat~\cite{instantsplat} accelerates convergence with sparse-view priors.

NoPo-Splat~\cite{ye2024no} marks a turning point by reconstructing a 3DGS scene and synthesizing novel views in a single forward pass without any pose input, opening the direction of pose-free novel view synthesis. Subsequent works follow two routes. On the reconstruction side, AnySplat~\cite{jiang2025anysplat}, WorldMirror~\cite{liu2025worldmirror}, and RayZer~\cite{jiang2025rayzer} scale feed-forward 3DGS to larger and more diverse scenes, yet they can only interpolate between observed viewpoints and cannot hallucinate unobserved content. On the generation side, end-to-end multi-view diffusion models such as Matrix3D~\cite{lu2025matrix3d} and Fillerbuster~\cite{weber2025fillerbuster} jointly predict poses and generate novel views, but lack explicit 3D geometric constraints, leading to inconsistencies in observed regions.

Combining reconstruction with diffusion is therefore a natural strategy: ViewCrafter~\cite{yu2024viewcrafter} conditions a video diffusion model on DUSt3R-predicted poses and rendered point maps, while CAT3D~\cite{cat3d} and SEVA~\cite{zhou2025stable} accept predicted poses directly but degrade severely because they are trained on ground-truth poses. These pipelines extract at most one signal from the reconstruction output and discard the rest; SplatGuide instead reuses the full 3DGS scene across three complementary conditioning signals.\sloppy

\paragraph{Conditioning Diffusion with 3D Reconstruction Priors.}
Existing methods that inject reconstruction priors into diffusion each bridge only one of the three information signals.
For the geometric signal, ViewCrafter~\cite{yu2024viewcrafter} and its follow-ups~\cite{mark2025trajectorycrafter,zhang2025spatialcrafter} render reconstructed point clouds at target poses and feed the resulting images into video diffusion models~\cite{liu2024reconxreconstructscenesparse,Cao_2025_CVPR,yin2025gsfixer}, providing a strong geometric anchor but tying the pipeline to a video backbone and limiting generation to sparse trajectory interpolation; earlier reprojection-based methods~\cite{chan2023genvs,zhou2023sparsefusion,wu2024difix3d} share this single signal under further constraints such as per-scene optimization or single-frame inpainting.
For the feature signal, Gen3C~\cite{ren2025gen3c} injects dense latent features via cross-attention, and Geometry Forcing~\cite{wu2025geometryforcing} aligns diffusion representations with geometric foundation model features; both improve spatial consistency but require dense features from models like VGGT at prohibitive computational cost.
Neither paradigm exploits the visibility signal already encoded in the reconstructed scene; SplatGuide bridges all three signals from a single reconstruction pass.

\paragraph{View Selection for Multi-View Generation.}
Scaling multi-view diffusion to large view sets hinges on selecting informative reference views. Temporally local conditioning~\cite{dfot,yu2025gamefactory,wang2024motionctrl,jason2023long,rombach2021geometry} suits contiguous video but not general multi-view settings. Spatial strategies rank candidates by pose distance~\cite{cat3d,zhou2025stable} or field-of-view overlap~\cite{worldmem,contextasmem}, lacking explicit occlusion reasoning, while VMem~\cite{vmem} models visibility with surfel-indexed memory but requires aggressive downsampling of 3D primitives, yielding coarse geometric discrimination. Our selector instead reads per-Gaussian source-view indices already produced by the reconstruction backbone, providing occlusion-aware selection at negligible cost without additional 3D primitives.

\begin{figure*}[!t]
    \centering
    \includegraphics[width=\linewidth]{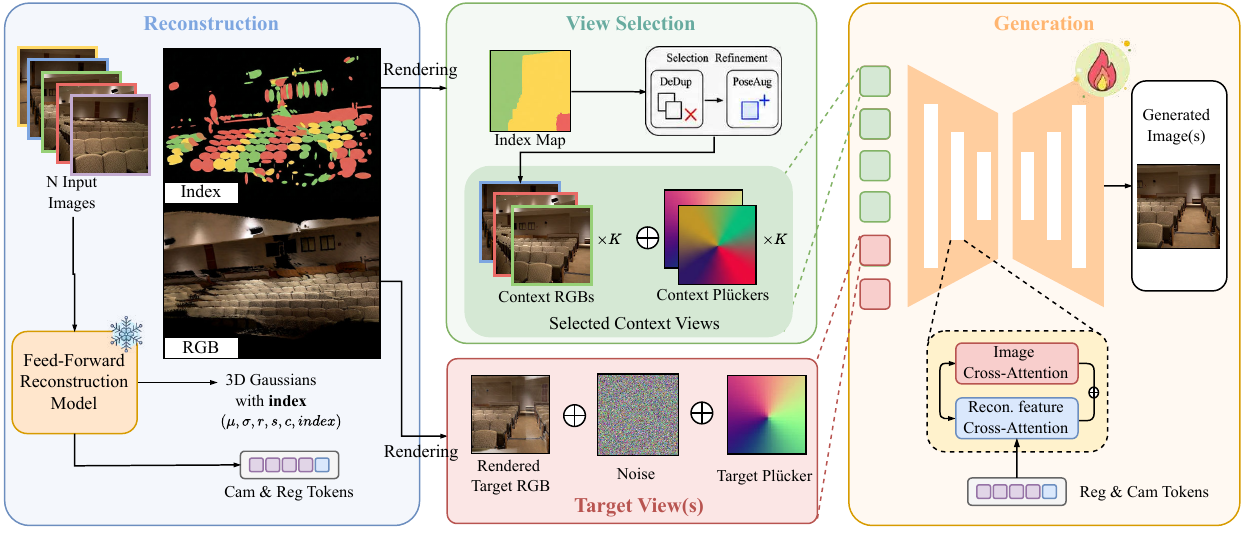}
    \caption{\textbf{Overview of SplatGuide.} \textit{Reconstruction:} A feed-forward model takes $N$ unposed images and produces 3D Gaussians with per-Gaussian source-view indices $(\mu, \sigma, r, s, c, \mathit{index})$, along with camera and register tokens. \textit{View Selection:} The per-Gaussian indices are rendered into a target-view index map; pixel-wise voting retrieves the top-$K$ reference views, whose RGBs and Pl\"ucker coordinates form the selected context views. \textit{Generation:} The rendered target RGB, noise, and target Pl\"ucker are concatenated with the context views and fed into the diffusion model, which is further conditioned on reconstruction tokens via cross-attention to produce the final photorealistic image.}
    \label{fig: main pipeline}
\end{figure*}










\section{Method}


\paragraph{Problem Formulation.}
Given a set of $N$ unposed RGB reference images ${\cal I}^{\rm ref} = \{ {\bf I}_i^{\rm ref} \}_{i=1}^N$ capturing the same static scene, our goal is to synthesize $M$ target views ${\cal I}^{\rm tgt} = \{ {\bf I}_j^{\rm tgt} \}_{j=1}^M$ at the corresponding query camera poses ${\cal P}^{\rm tgt} = \{ {\bf P}_j^{\rm tgt} \}_{j=1}^M$. For each target view, we model the generation as
\begin{equation}
p(\mathbf{I}_j^{\mathrm{tgt}} \mid \mathcal{I}^{\mathrm{ref}}, {\cal C}, \mathbf{P}_j^{\mathrm{tgt}}),
\end{equation}
where ${\cal C} = f_{\rm recon}({\cal I}^{\rm ref})$ denotes target-independent reconstruction conditioning from the reference images, comprising estimated reference poses ${\cal P}^{\rm ref}$, a reconstructed 3DGS scene ${\cal G}$, and reconstruction features.

\subsection{Model Architecture}

\paragraph{Overview.}
As illustrated in \cref{fig: main pipeline}, SplatGuide follows a three-stage pipeline of reconstruction, selection, and generation. A feed-forward backbone estimates the reference camera poses and reconstructs a 3DGS scene ${\cal G}$ from the unposed inputs. The resulting renderings, source-view indices, and reconstruction tokens respectively provide pixel-level conditioning, visibility-aware reference selection, and feature-level guidance. Together, these signals guide $f_{\mathrm{diff}}$ to synthesize the target views.

\paragraph{Reconstruction Model.}
\label{sec:recon-model}
The reconstruction model recovers camera poses and scene geometry from the unposed input. The feed-forward model $f_{\mathrm{recon}}$ processes the reference set ${\cal I}^{\rm ref}$ and outputs both poses and an explicit 3DGS scene:
\begin{equation}
(\mathcal{P}^{\mathrm{ref}}, \mathcal{G}) = f_{\mathrm{recon}}(\mathcal{I}^{\mathrm{ref}}).
\end{equation}
We adopt backbones that predict \emph{pixel-aligned} Gaussians, \ie, every pixel of every reference view regresses one Gaussian. Each $g_i \in \mathcal{G}$ therefore carries a source-view index $v(i) \in \{1,\dots,V\}$ that is determined by construction at reconstruction time: no Gaussian is fused or merged across views, so the index requires neither extra supervision nor any tie-breaking heuristic. This property is shared by all reconstruction backbones we evaluate, and it is the only requirement a backbone must satisfy to be used as a drop-in replacement.
From the reconstructed scene $\mathcal{G}$, we render coarse but geometrically consistent images at both target poses $\mathcal{P}^{\mathrm{tgt}}$ to obtain $\hat{\mathcal{I}}^{\mathrm{tgt}}$ and estimated reference poses $\mathcal{P}^{\mathrm{ref}}$ to obtain $\hat{\mathcal{I}}^{\mathrm{ref}}$, via standard 3DGS rendering~\cite{kerbl20233d}:
\begin{equation}
\hat{\mathbf{I}}_{\mathbf{P}} = \sum_{k \in \mathcal{N}(\mathbf{P})} \mathbf{c}_k \alpha_k' \prod_{l=1}^{k-1} (1 - \alpha_l'),
\label{eq:3dgs_general}
\end{equation}
where ${\bf c}_k$ and $\alpha_k'$ are the learned color and projected opacity of the $k$-th Gaussian, and $\mathcal{N}(\mathbf{P})$ denotes the depth-sorted set of Gaussians visible from pose $\mathbf{P}$. Setting $\mathbf{P} = \mathbf{P}_j^{\mathrm{tgt}}$ or $\mathbf{P} = \mathbf{P}_i^{\mathrm{ref}}$ yields the rendered target images $\hat{\mathbf{I}}^{\mathrm{tgt}}$ and reference images $\hat{\mathbf{I}}^{\mathrm{ref}}$, respectively.
These rendered images constitute the primary geometric anchor of our pipeline: unlike pure 2D conditioning, they encode both geometry and appearance from the reconstructed scene and inherently maintain multi-view consistency. Our ablation (\cref{tab:ablation_structure}) confirms that the rendered image is the largest single contributor to generation quality.


\paragraph{Diffusion Model.}
The diffusion-based generator $f_{\mathrm{diff}}$, built on SEVA~\cite{zhou2025stable}, refines the coarse rendered images $\hat{\mathbf{I}}^{\mathrm{tgt}}$ into photorealistic novel views $\mathbf{I}^{\mathrm{tgt}}$, receiving geometric guidance at two complementary levels: pixel-level conditioning via rendered images and feature-level conditioning via injected reconstruction tokens.

\paragraph{Rendering as Geometric Conditioning.}
The primary geometric signal comes from the rendered images produced by $f_{\mathrm{recon}}$. We choose channel-wise concatenation for injection because it preserves the pixel-aligned spatial correspondence between the rendered geometry and the generated content. Following the SEVA conditioning framework, each reference image $\mathbf{I}^{\mathrm{ref}}$ is encoded into a latent $\mathbf{z} = \mathcal{E}(\mathbf{I}^{\mathrm{ref}})$ via the VAE encoder $\mathcal{E}(\cdot)$ and concatenated with its Pl\"ucker ray embedding and a binary mask distinguishing reference from target views. For target views, we replace the latent with the noisy state $\mathbf{z}_t$.
We extend this by encoding the rendered images $\hat{\mathbf{I}}^{\mathrm{ref}}$ and $\hat{\mathbf{I}}^{\mathrm{tgt}}$ into the latent space and concatenating them as an additional channel group $\mathbf{z}_{\text{render}} \in \mathbb{R}^{H \times W \times C}$:
\begin{equation}
\mathbf{z}_{\text{cond}} = [\mathbf{z}_t, \mathbf{z}_{\text{render}}, \mathbf{e}_{\text{plk}}, \mathbf{m}] \in \mathbb{R}^{H \times W \times (2C + C_{\text{plk}} + 1)},
\end{equation}
where $\mathbf{e}_{\text{plk}} \in \mathbb{R}^{H \times W \times C_{\text{plk}}}$ encodes the 6D Pl\"ucker coordinates for each pixel ray and $\mathbf{m} \in \{0,1\}^{H \times W \times 1}$ indicates valid regions. To accommodate the additional $C$ channels, we expand the first convolutional layer of the U-Net from $C + C_{\text{plk}} + 1$ to $2C + C_{\text{plk}} + 1$ input channels, initializing the new weights to zeros so that the pre-trained model behavior is preserved at the start of training.

\paragraph{Tokens as Feature Conditioning.}
\label{sec:token-injection}
Rendered images provide pixel-aligned geometric conditioning but cannot convey scene-level context beyond the visible surfaces, such as global structure and texture statistics. We therefore inject reconstruction tokens from $f_{\rm recon}$ into the diffusion model.
For each reference view $i$ we extract from the last layer of the reconstruction backbone one camera token $\mathbf{t}_i^{\text{cam}} \in \mathbb{R}^{d_r}$ used for pose prediction, and four register tokens $\{\mathbf{t}_{i,l}^{\text{reg}}\}_{l=1}^{4} \in \mathbb{R}^{4 \times d_r}$ ($d_r{=}1024$), which aggregate non-local information across patches~\cite{darcet2023vision}. A learned linear projection maps each token to the cross-attention dimension of the diffusion U-Net, and the tokens are injected through dedicated cross-attention layers, analogous to how SEVA~\cite{zhou2025stable} injects CLIP~\cite{radford2021learning} features. Our ablation (\cref{tab:ablation_structure}) confirms that these feature cues are complementary to rendered images, further improving both fidelity and perceptual quality.

The final conditioning $\mathcal{C}$ thus comprises reference latents, rendered latents, Pl\"ucker ray embeddings, the binary view-type mask, and the camera and register tokens.

\paragraph{Training Objective.}
We train only the diffusion model while keeping the reconstruction backbone frozen. Following standard latent diffusion practice~\cite{song2020denoising}, we minimize the noise-prediction objective:
\begin{equation}
\mathcal{L} = \mathbb{E}_{\mathbf{z}_0, \boldsymbol{\epsilon}, t}
\left[
\left\| \boldsymbol{\epsilon} - \boldsymbol{\epsilon}_\theta(\mathbf{z}_t, t, \mathcal{C}) \right\|_2^2
\right],
\end{equation}
where $\mathbf{z}_0 = \mathcal{E}(\mathbf{I}^{\mathrm{tgt}})$ is the VAE-encoded target, $\boldsymbol{\epsilon} \sim \mathcal{N}(\mathbf{0}, \mathbf{I})$, $t \sim \mathcal{U}(1, T)$, and $\mathbf{z}_t$ is the noised latent at timestep $t$. 

\subsection{Gaussians as Context Views Selector}
\label{sec:gaussian-selection}

\begin{figure}[!t]
    \centering
    \includegraphics[width=\linewidth]
    {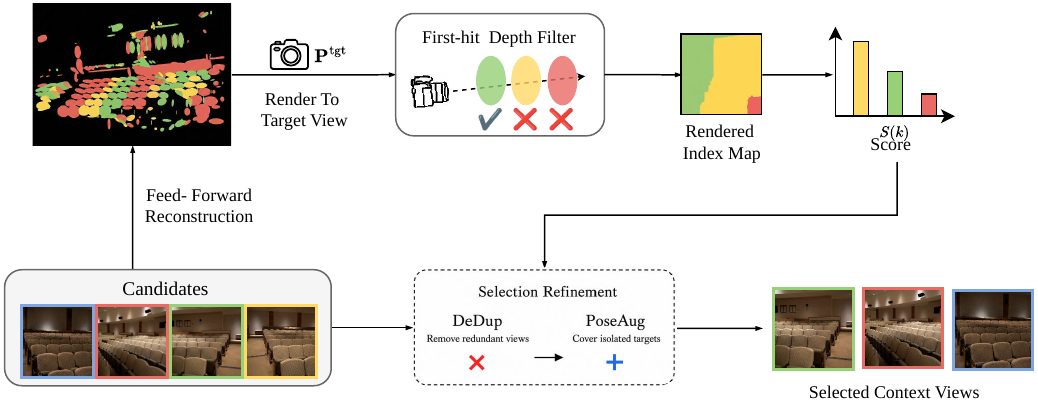}
    \caption{Overview of the proposed view selection pipeline. We reuse the reconstructed 3D Gaussians with per-Gaussian source-view indices, render an occlusion-aware target-view index map via first-hit visibility, and aggregate pixel votes into per-view visibility scores $S(k)$ to determine Top-K Indices. The final selection is refined by DeDup, which filters spatially redundant candidates, and PoseAug, which augments the context set with pose-proximal views for isolated targets, ensuring a complete and diverse set of $K$ reference images.
    }
    \label{fig:view-selection-pipeline}
\end{figure}

Scaling pose-free NVS to practical settings requires handling candidate pools that far exceed the context budget of the diffusion model; our experiments show that the selection policy alone accounts for a 7~dB range in output quality (\cref{tab:selection-main}), making view selection a first-order design decision.
We design a hybrid strategy that combines scene-based visibility reasoning with a lightweight pose-based augmentation. As illustrated in~\cref{fig:view-selection-pipeline}, the selector (1) reconstructs a 3D Gaussian proxy with source-view indices, (2) renders an occlusion-aware view-index map in the target view, and (3) selects reference views via visibility-based scoring, directly reusing the 3DGS scene $\mathcal{G}$ already produced by the reconstruction backbone.

Importantly, spatial proximity between cameras does not necessarily imply shared visible content due to occlusions. Given a target camera $\mathbf{P}^{\mathrm{tgt}}$ and reconstructed Gaussians $\mathcal{G}=\{g_i\}_{i=1}^G$ with their source-view indices $v(i)$ from \cref{sec:recon-model}, we estimate how much of the surface visible from $\mathbf{P}^{\mathrm{tgt}}$ is explained by each candidate view. Since selection only requires coarse visibility, we downsample the Gaussian set to reduce computational overhead. Reusing the projection and rasterization operations of 3D Gaussian Splatting~\cite{kerbl20233d}, we render a \emph{view-index map} under a hard first-hit depth test: at each pixel $p$, only the nearest Gaussian is kept and writes a distinct palette color $c_{v(i)}$ encoding its source view, while pixels with no valid Gaussian are ignored. Whereas $v(i)$ is unambiguous per Gaussian, a target ray typically intersects Gaussians originating from several different views; the first-hit test resolves this competition by awarding the pixel to the nearest surface only, and palette colors are never alpha-blended, which keeps index recovery exact. Each target pixel thus votes for the source view that best explains its visible surface, and the view index $\hat{v}(p)$ is recovered by nearest-color lookup. We then define a visibility score
\begin{equation}
S(k) = \sum_{p} \mathbb{I}\bigl[\hat{v}(p) = k\bigr],
\end{equation}
which counts the target pixels dominated by geometry from view $k$. We rank candidates by $S(k)$ in descending order and greedily select the highest-scoring views until the context budget is exhausted.

\paragraph{Beyond Geometric Coverage.}
Ranking by $S(k)$ alone, as in prior visibility-based retrieval~\cite{vmem}, optimizes purely for geometric coverage. Coverage, however, does not guarantee informative conditioning: a faraway candidate can observe much of the target surface yet deliver appearance evidence that is too coarse, and several top-ranked candidates frequently explain the same surfaces. We therefore refine the ranking with two lightweight components.
\emph{DeDup} discards candidates whose visible Gaussians largely coincide with those of already selected views, preventing the limited budget from being spent on near-duplicate viewpoints.
\emph{PoseAug} fills the remaining slots with a max-min proximity rule that guards against the complementary failure mode: a target left distant from every chosen context. With $\mathcal{S}$ the current context set and $\mathbf{t}_k$, $\mathbf{t}^{\mathrm{tgt}}_{j}$ the camera centers of candidate $k$ and target $j$, PoseAug locates the most isolated target,
\begin{equation}
j^{\star} = \arg\max_{j \in \{1,\dots,M\}} \, \min_{k \in \mathcal{S}} \bigl\| \mathbf{t}_k - \mathbf{t}^{\mathrm{tgt}}_{j} \bigr\|_2,
\label{eq:poseaug}
\end{equation}
and admits the unselected candidate nearest to it, $k^{\star} = \arg\min_{k \notin \mathcal{S}} \bigl\| \mathbf{t}_k - \mathbf{t}^{\mathrm{tgt}}_{j^{\star}} \bigr\|_2$; the rule repeats until the budget is saturated, so every target retains at least one nearby context. The supplementary material ablates both components, confirming that they are complementary to the visibility ranking.

\section{Experiments}

We evaluate SplatGuide on four benchmarks spanning in-domain and out-of-domain scenes. Comparison experiments show that our full pipeline matches or surpasses ground-truth-pose methods, controlled selection experiments isolate the effect of visibility-aware view selection, and ablations verify that each conditioning signal provides complementary gains.

\paragraph{Implementation Details.}
Our diffusion backbone builds on SEVA~\cite{zhou2025stable}, and we adopt WorldMirror~\cite{liu2025worldmirror} as the reconstruction model. Both the reconstruction backbone $f_{\mathrm{recon}}$ and the VAE remain frozen throughout training; $f_{\mathrm{recon}}$ produces the rendered images and tokens that constitute the conditioning $\mathcal{C}$ but receives no gradient updates. Only the diffusion model is trained, using the Adam optimizer with a learning rate of $1\times10^{-5}$ on 8 H200 GPUs with a batch size of 32. Classifier-free guidance is not used during training. At inference time, we use the DDIM sampler~\cite{song2020denoising} with classifier-free guidance~\cite{ho2022classifier}.

\paragraph{Training and Test Data.}
We train the diffusion model on DL3DV~\cite{ling2024dl3dv} with 10,510 scenes and RealEstate10K~\cite{zhou2018stereo} with 67,477 training videos, both containing mixed indoor and outdoor scenes. We evaluate on two in-domain datasets, RealEstate10K and DL3DV, and two out-of-domain datasets, Mip-NeRF 360~\cite{barron2022mipnerf360} and Tanks and Temples~\cite{Knapitsch2017}, to test generalization. Further details are provided in the supplementary material.






\begin{figure*}[!t]
\centering
\resizebox{0.88\textwidth}{!}{%
\begin{tikzpicture}[
    image/.style = {
        inner sep=0pt, outer sep=0pt,
        minimum width=2cm,
        minimum height=\rowimgheight,
        anchor=north west, text width=2cm, text height=\rowimgheight
    },
    node distance = 1pt and 1pt,
    every node/.style={font={\tiny}},
    label/.style = {font={\footnotesize\bfseries\vphantom{p}},
                    anchor=south,inner sep=0pt},
    inputimg/.style = {
        inner sep=0pt, outer sep=0pt,
        anchor=north east
    }
]

\node[image] (img-00)
  {\includegraphics[height=\rowimgheight]{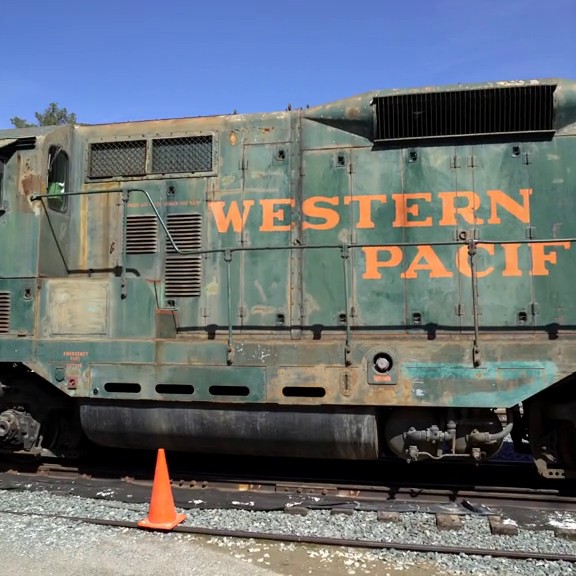}};
\node[image,right=of img-00] (img-01)
  {\includegraphics[height=\rowimgheight]{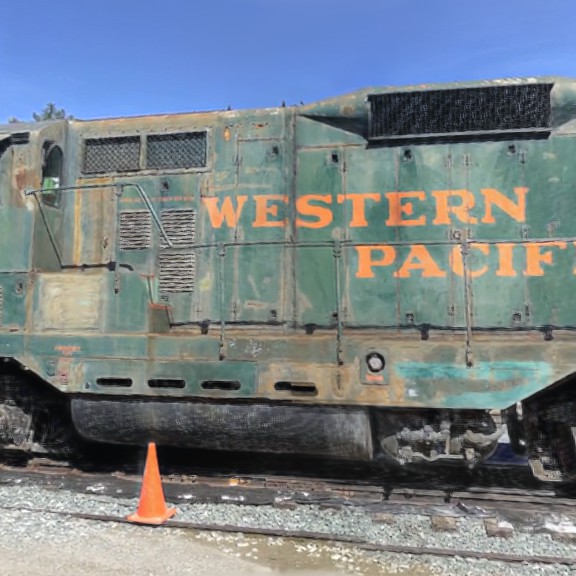}};
\node[image,right=of img-01] (img-02)
  {\includegraphics[height=\rowimgheight]{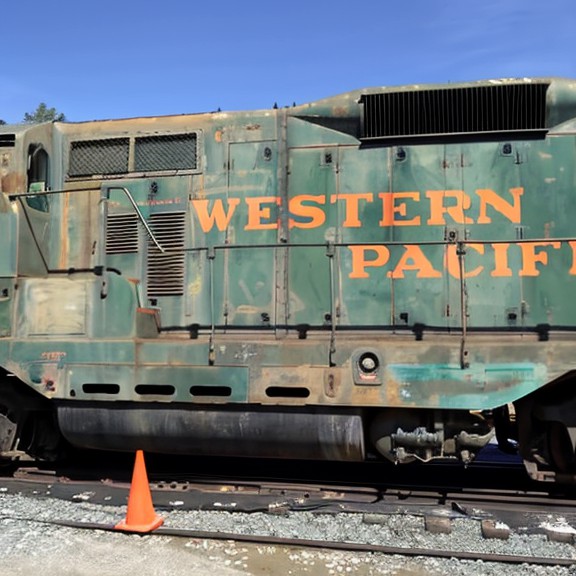}};
\node[image,right=of img-02] (img-03)
  {\includegraphics[height=\rowimgheight]{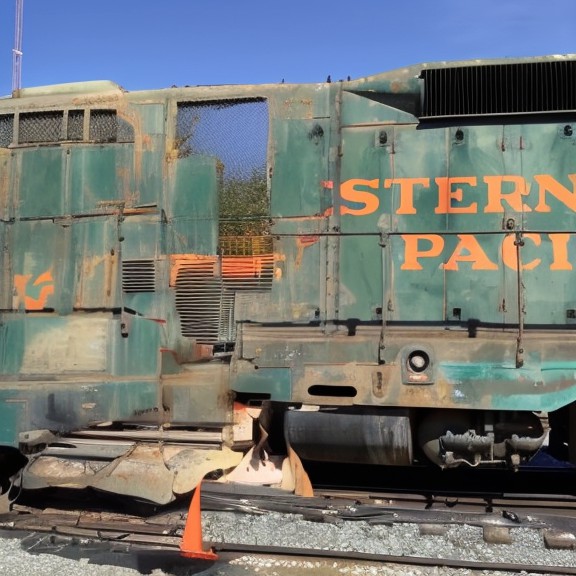}};
\node[anchor=north west, inner sep=0pt, outer sep=0pt,
      minimum height=\rowimgheight] (img-04) at ($(img-03.north east)+(1pt,0)$)
  {\includegraphics[height=\rowimgheight]{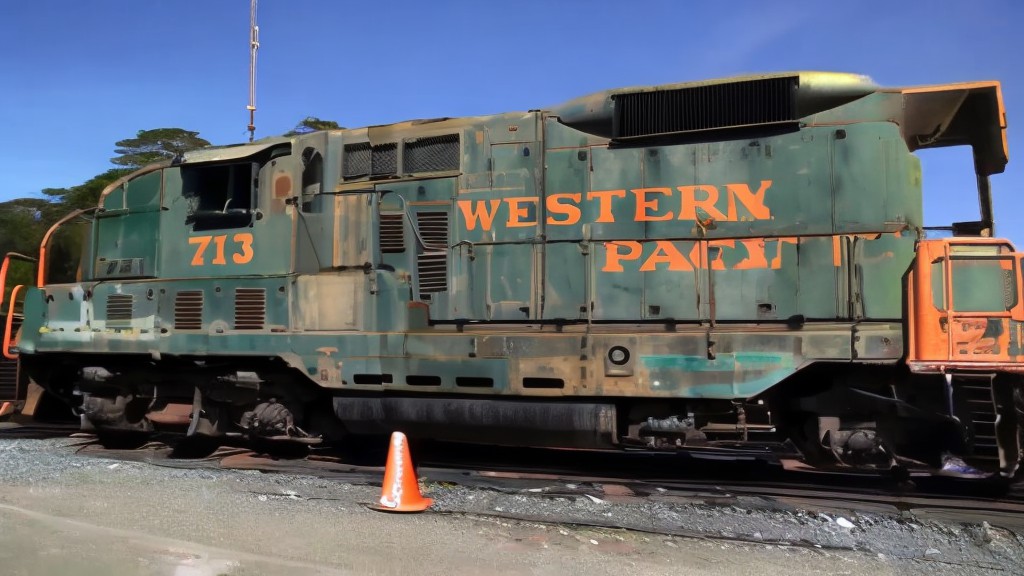}};

\node[image,below=of img-00] (img-10)
  {\includegraphics[height=\rowimgheight]{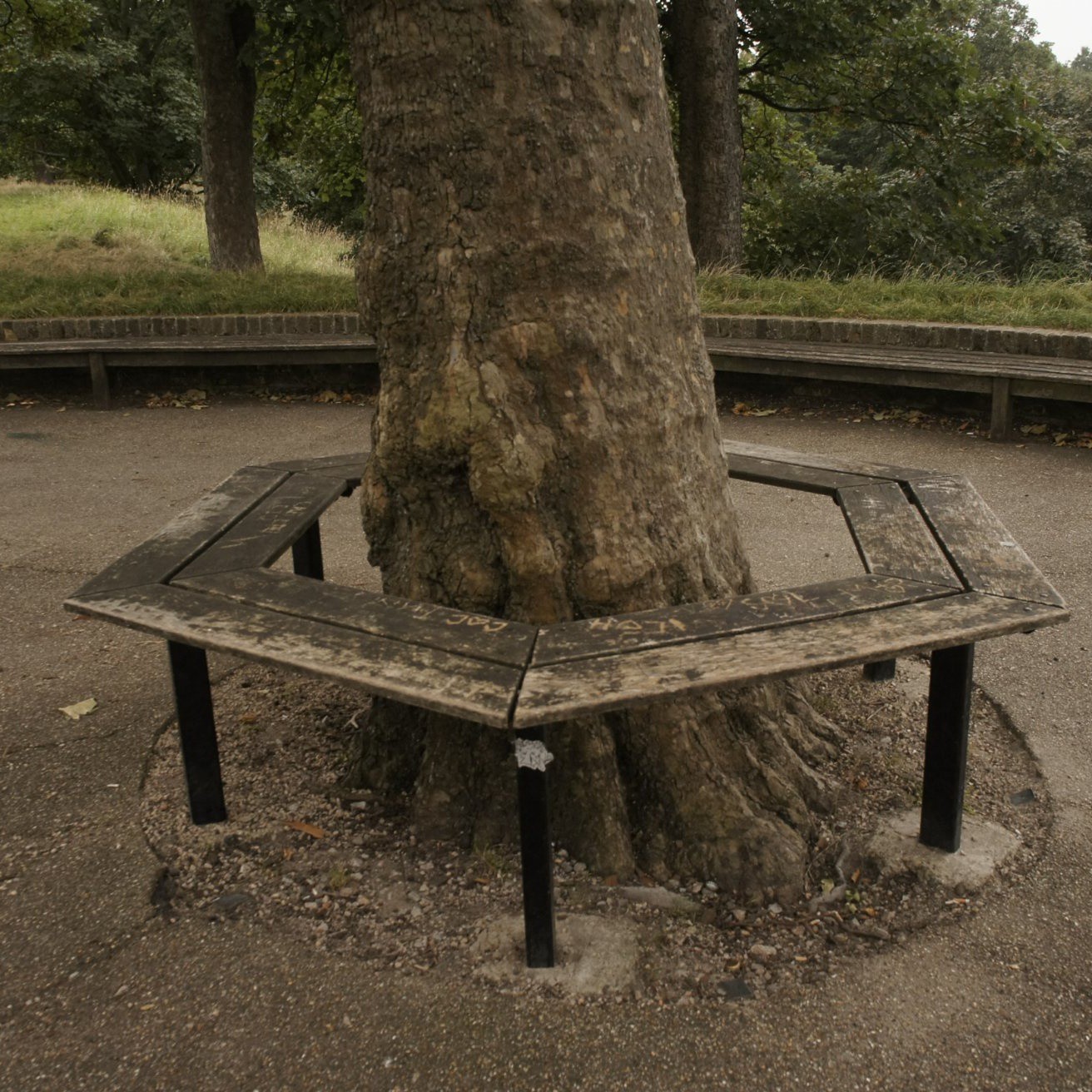}};
\node[image,right=of img-10] (img-11)
  {\includegraphics[height=\rowimgheight]{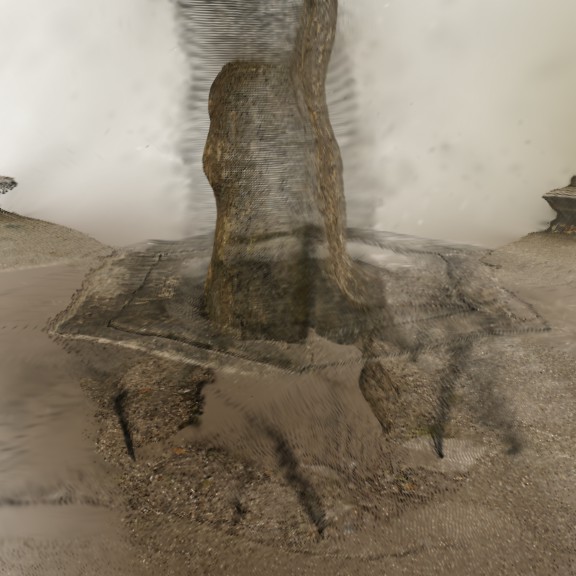}};
\node[image,right=of img-11] (img-12)
  {\includegraphics[height=\rowimgheight]{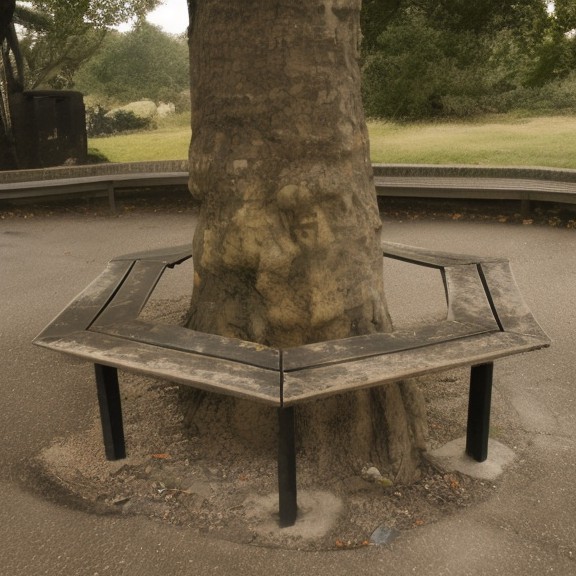}};
\node[image,right=of img-12] (img-13)
  {\includegraphics[height=\rowimgheight]{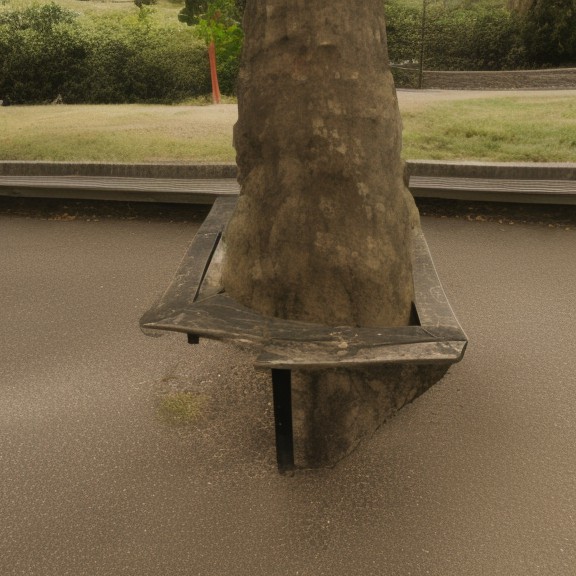}};
\node[anchor=north west, inner sep=0pt, outer sep=0pt,
      minimum height=\rowimgheight] (img-14) at ($(img-13.north east)+(1pt,0)$)
  {\includegraphics[height=\rowimgheight]{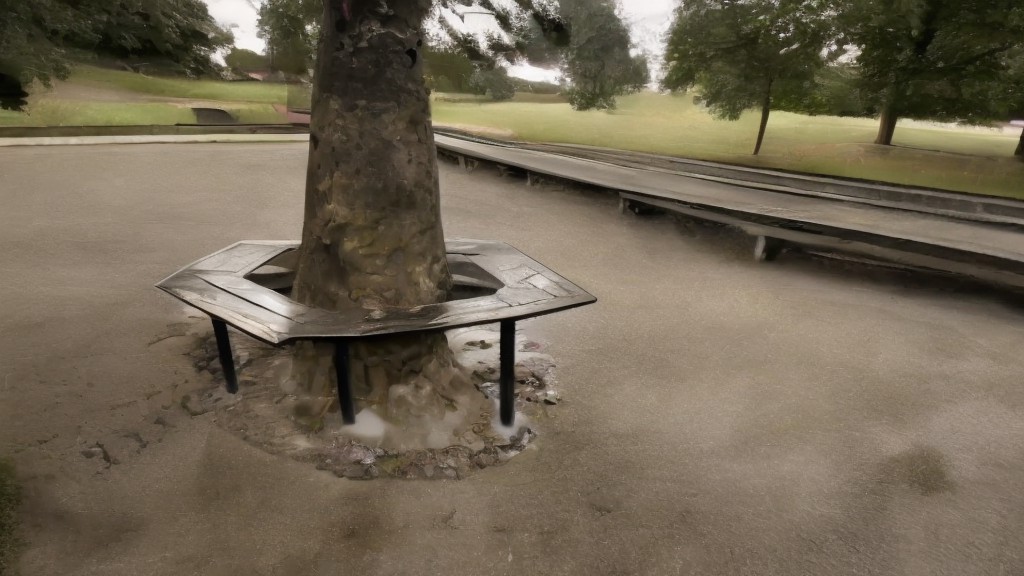}};


\node[image,below=of img-10] (img-30)
  {\includegraphics[height=\rowimgheight]{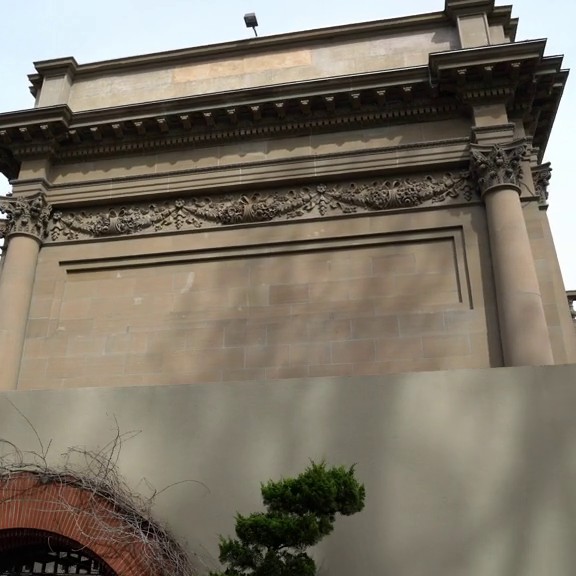}};
\node[image,right=of img-30] (img-31)
  {\includegraphics[height=\rowimgheight]{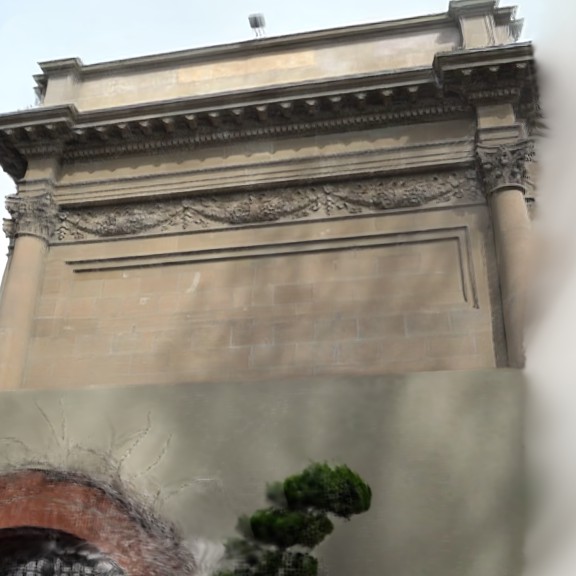}};
\node[image,right=of img-31] (img-32)
  {\includegraphics[height=\rowimgheight]{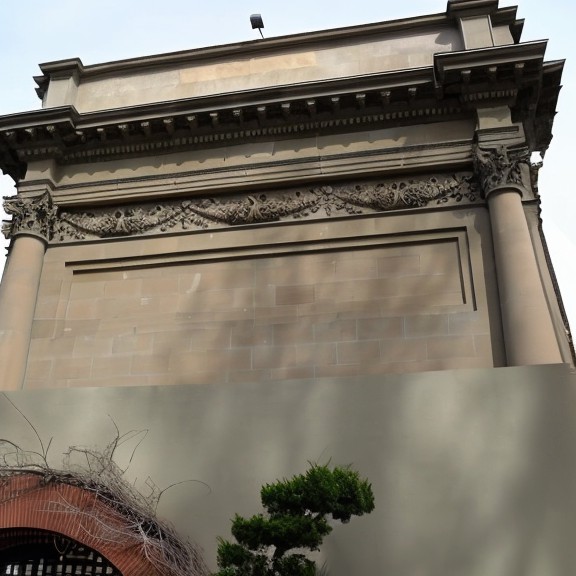}};
\node[image,right=of img-32] (img-33)
  {\includegraphics[height=\rowimgheight]{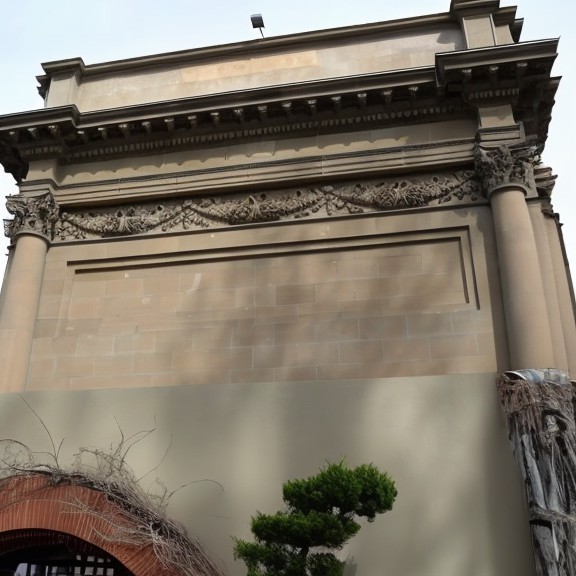}};
\node[anchor=north west, inner sep=0pt, outer sep=0pt,
      minimum height=\rowimgheight] (img-34) at ($(img-33.north east)+(1pt,0)$)
  {\includegraphics[height=\rowimgheight]{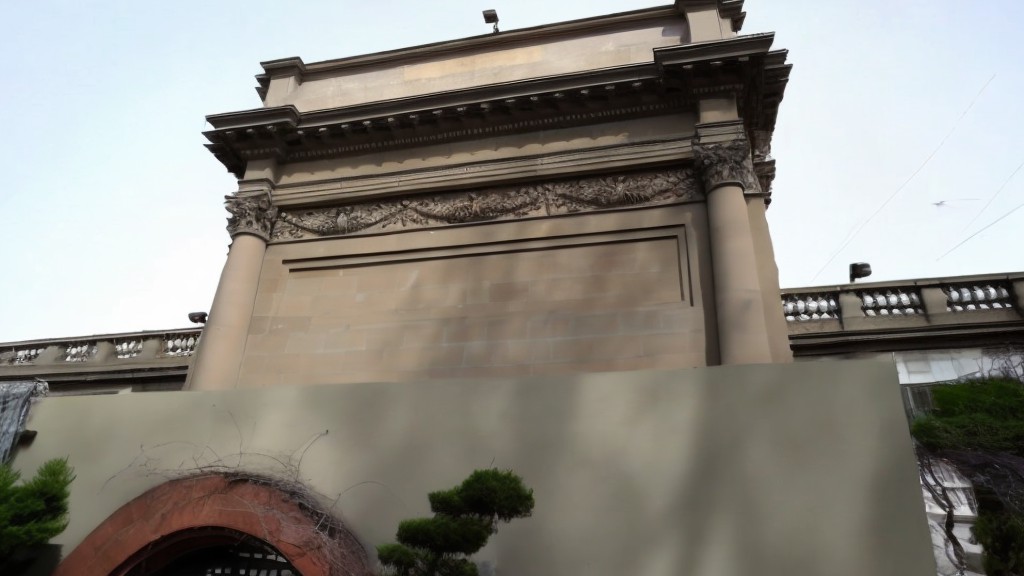}};

\node[inputimg] (in-00a)
  at ($(img-00.north west)+(-1mm,0)$)
  {\includegraphics[height=\inputimgheight]{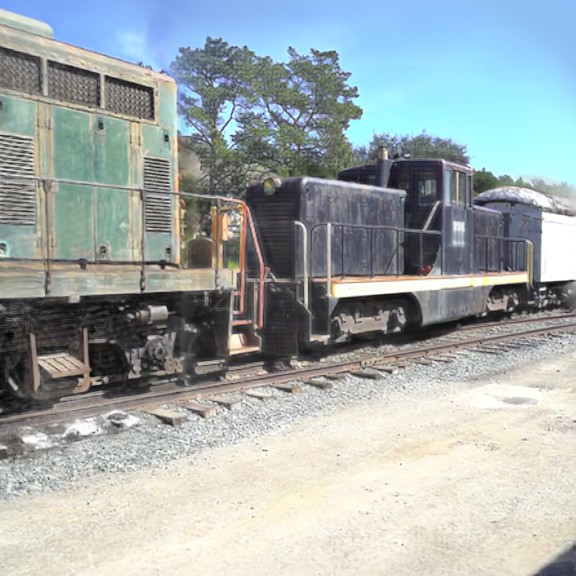}};
\node[inputimg, below=0pt of in-00a] (in-00b)
  {\includegraphics[height=\inputimgheight]{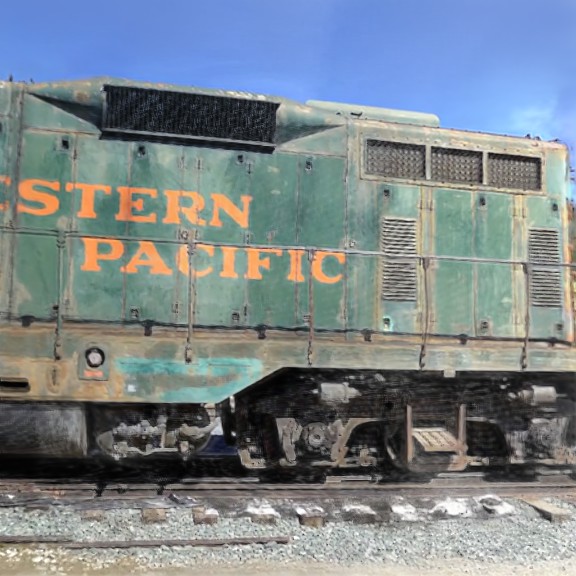}};
\node[inputimg, below=0pt of in-00b] (in-00c)
  {\includegraphics[height=\inputimgheight]{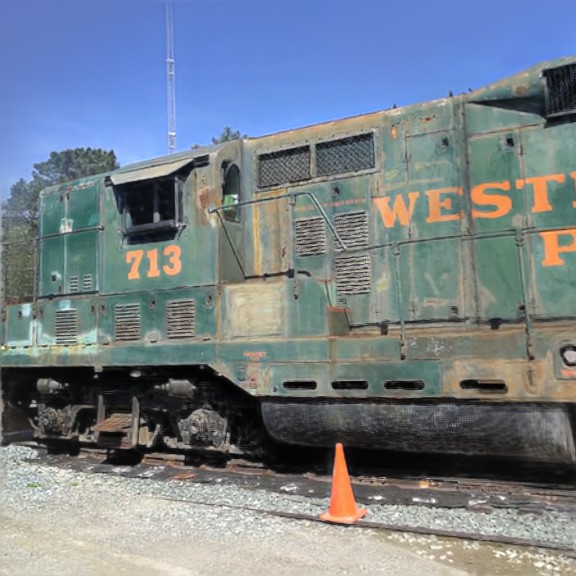}};

\node[inputimg] (in-10a)
  at ($(img-10.north west)+(-1mm,0)$)
  {\includegraphics[height=\inputimgheight]{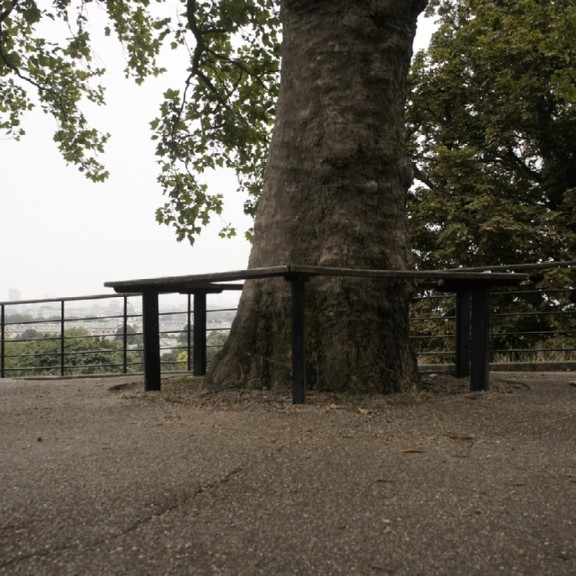}};
\node[inputimg, below=0pt of in-10a] (in-10b)
  {\includegraphics[height=\inputimgheight]{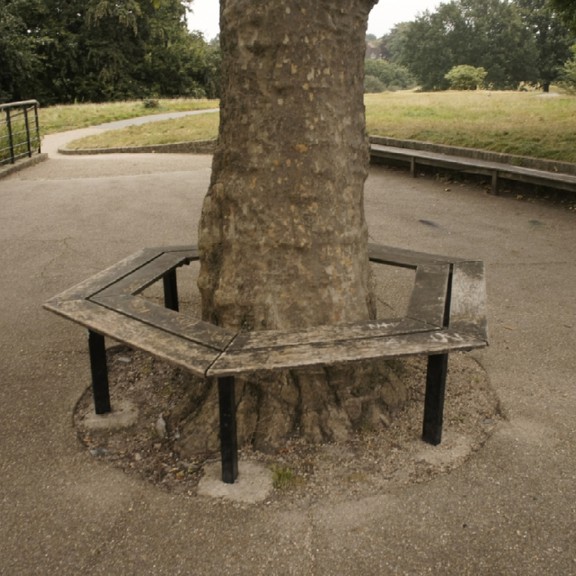}};
\node[inputimg, below=0pt of in-10b] (in-10c)
  {\includegraphics[height=\inputimgheight]{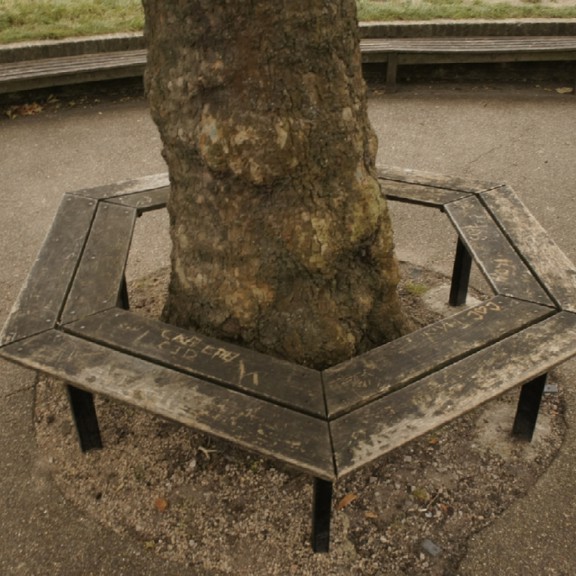}};


\node[inputimg] (in-30a)
  at ($(img-30.north west)+(-1mm,0)$)
  {\includegraphics[height=\inputimgheight]{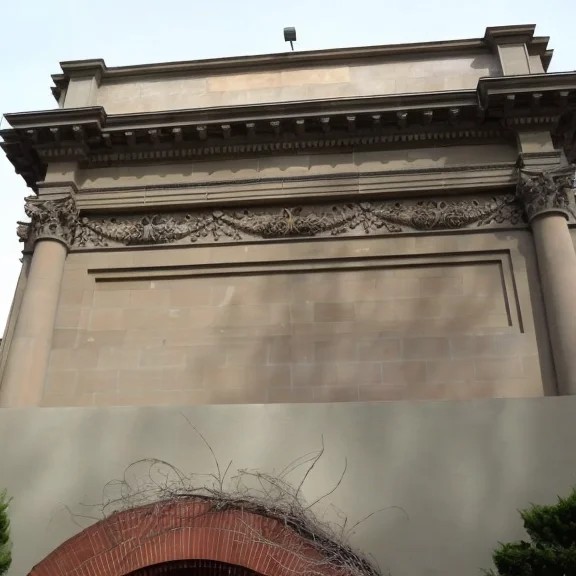}};
\node[inputimg, below=0pt of in-30a] (in-30b)
  {\includegraphics[height=\inputimgheight]{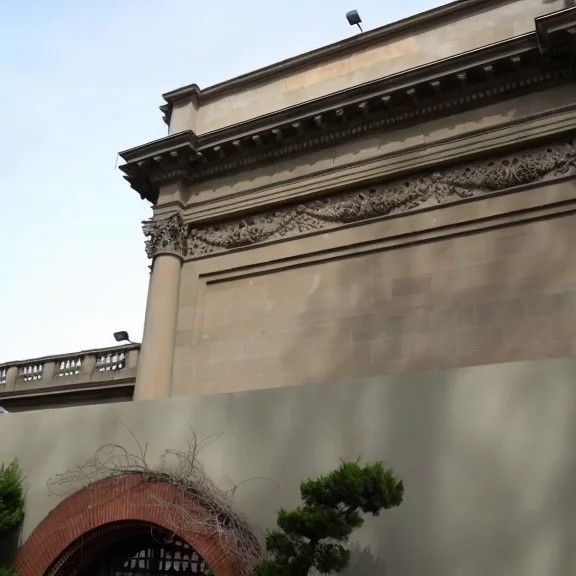}};
\node[inputimg, below=0pt of in-30b] (in-30c)
  {\includegraphics[height=\inputimgheight]{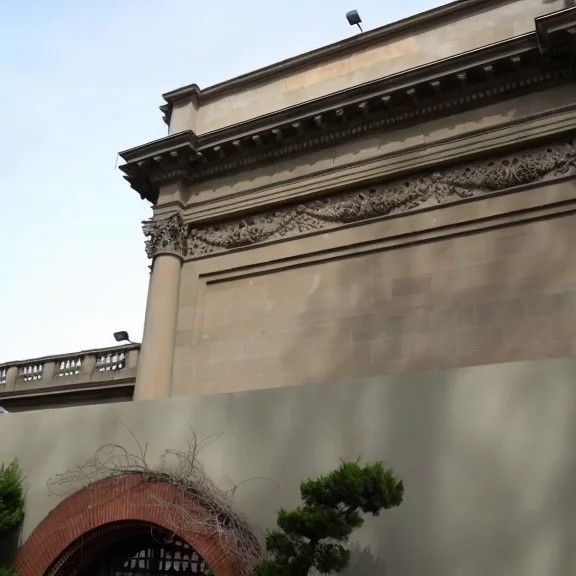}};

\node[label] at (img-00.north) {GT};
\node[label] at (img-01.north) {Render};
\node[label] at (img-02.north) {Ours};
\node[label] at (img-03.north) {SEVA};
\node[label] at (img-04.north) {ViewCrafter};

\end{tikzpicture}
}
\caption{Visual comparison of novel view synthesis. Columns from left to right: ground truth, the coarse 3DGS render serving as our geometric prior, our SplatGuide result, SEVA, and ViewCrafter. SplatGuide produces sharper structures and fewer artifacts than both baselines, benefiting from the explicit geometric guidance provided by the rendered image and injected tokens.}
\label{fig:grid7x4}
\end{figure*}
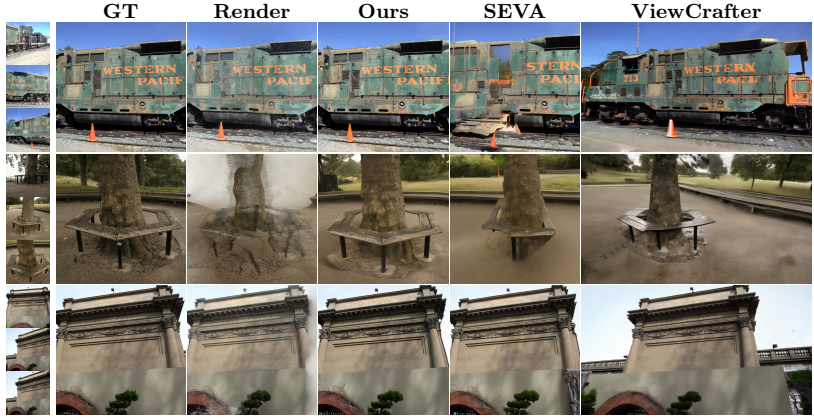

\subsection{Novel View Synthesis Results}

\begin{table*}[!t]
    \centering\footnotesize
    \caption{Quantitative comparison on the in-domain RealEstate10K (top) and DL3DV (bottom) datasets. Best results among unposed methods are in \textbf{bold}, and second-best are \underline{underlined}.}
    \label{tab:in_domain_results}
    \setlength{\tabcolsep}{4pt}
    \resizebox{\textwidth}{!}{
    \begin{tabular}{lc|ccc|ccc|ccc}
    \toprule
    & \textbf{GT} & \multicolumn{3}{c}{3-view} & \multicolumn{3}{c}{6-view} & \multicolumn{3}{c}{9-view} \\
    \cmidrule{3-5}\cmidrule{6-8}\cmidrule{9-11}
    \textbf{Method} & \textbf{pose} & PSNR$\uparrow$ & SSIM$\uparrow$ & LPIPS$\downarrow$ & PSNR$\uparrow$ & SSIM$\uparrow$ & LPIPS$\downarrow$ & PSNR$\uparrow$ & SSIM$\uparrow$ & LPIPS$\downarrow$ \\
    \midrule
    \multicolumn{11}{c}{\textbf{RealEstate10K}}\\
    \midrule
    SEVA & \checkmark & 27.57&0.89&0.07&29.24&0.90&0.06&29.63&0.91&0.05 \\
    AnySplat& \texttimes &19.81&0.71&0.24 &23.23&0.79&0.16 &24.13&0.82&0.15 \\
    WorldMirror & \texttimes & 21.13 & 0.77 & 0.16 & 22.89 & 0.81 & 0.12 & 23.35 & 0.83 & 0.11 \\
    Fillerbuster& \texttimes &19.13&0.63&0.27 &19.81&0.65&0.24 &20.59&0.67&0.22 \\
    ViewCrafter & \texttimes & 20.18&0.74&0.23 &22.97&0.81&0.19 &23.22&0.82&0.17 \\
    SEVA & \texttimes &\underline{23.47}&\underline{0.77}&\underline{0.13}&\underline{26.20}&\underline{0.82}&\underline{0.07}&\underline{27.14}&\underline{0.83}&\underline{0.06}\\
    Ours & \texttimes & \textbf{26.52}&\textbf{0.84}&\textbf{0.07} & \textbf{29.29} &\textbf{0.87} &\textbf{0.04} &\textbf{30.00}&\textbf{0.88}&\textbf{0.04} \\
    \midrule
    \multicolumn{11}{c}{\textbf{DL3DV}}\\
    \midrule
    SEVA & \checkmark & 15.13 & 0.43 & 0.41 & 16.62 & 0.48 & 0.33 & 17.68 & 0.52 & 0.27 \\
    AnySplat & \texttimes & 10.04 & 0.29 & 0.61 & 12.15 & 0.33 & 0.56 & 13.78 & 0.37 & 0.51 \\
    WorldMirror & \texttimes & 13.11 & 0.31 & 0.54 & 13.98 & 0.34 & \underline{0.47} & 15.01 & 0.41 & \underline{0.42} \\
    RayZer & \texttimes & \underline{14.65} & \underline{0.35} & 0.64 & \textbf{16.28} & \underline{0.40} & 0.54 & \textbf{17.75} & \underline{0.46} & 0.46 \\
    Matrix3D & \texttimes & 12.73 & 0.31 & \underline{0.53} & 13.35 & 0.32 & 0.50 & - & - & - \\
    ViewCrafter & \texttimes & 11.16 & 0.28 & 0.61 & - & - & - & - & - & - \\
    SEVA & \texttimes & 12.50 & \underline{0.35} & 0.55 & 13.49 & 0.37 & 0.48 & 14.26 & 0.38 & 0.43 \\
    Ours & \texttimes & \textbf{14.96} & \textbf{0.42} & \textbf{0.39} & \underline{16.21} & \textbf{0.46} & \textbf{0.31} & \underline{16.99} & \textbf{0.48} & \textbf{0.27} \\
    \bottomrule
    \end{tabular}
    }
\end{table*}

\begin{table*}[!t]
    \centering\footnotesize
    \caption{Quantitative comparison on out-of-domain datasets. All methods operate without ground-truth poses. Best results are in \textbf{bold}, and second-best are \underline{underlined}.}
    \label{tab:out_of_domain_results}
    \setlength{\tabcolsep}{3pt}
    \resizebox{\textwidth}{!}{
    \begin{tabular}{l|ccc|ccc|ccc|ccc|ccc}
    \toprule
    & \multicolumn{6}{c|}{\textbf{Tanks and Temples}} & \multicolumn{9}{c}{\textbf{Mip-NeRF 360}} \\
    \cmidrule{2-7}\cmidrule{8-16}
    \textbf{Method} & \multicolumn{3}{c}{3-view} & \multicolumn{3}{c|}{6-view} & \multicolumn{3}{c}{3-view} & \multicolumn{3}{c}{6-view} & \multicolumn{3}{c}{9-view} \\
    \cmidrule{2-4}\cmidrule{5-7}\cmidrule{8-10}\cmidrule{11-13}\cmidrule{14-16}
    & PSNR$\uparrow$ & SSIM$\uparrow$ & LPIPS$\downarrow$ & PSNR$\uparrow$ & SSIM$\uparrow$ & LPIPS$\downarrow$ & PSNR$\uparrow$ & SSIM$\uparrow$ & LPIPS$\downarrow$ & PSNR$\uparrow$ & SSIM$\uparrow$ & LPIPS$\downarrow$ & PSNR$\uparrow$ & SSIM$\uparrow$ & LPIPS$\downarrow$ \\
    \midrule
    \multicolumn{16}{l}{\textit{Regression-based methods:}}\\
    AnySplat &16.04&0.46&0.35&18.00&0.52&0.30 &8.26&0.20&0.66 &10.30&0.23&0.60 &11.42 &0.26 &0.56\\
    WorldMirror & 17.45 & 0.55 & 0.27 & 19.06 & 0.56 & 0.25 & 11.90 & 0.24 & 0.61 & 12.52 & \underline{0.26} & 0.55 & 13.07 & 0.26 & 0.52 \\
    \midrule
    \multicolumn{16}{l}{\textit{Diffusion-based methods:}}\\
    Fillerbuster &17.18&0.46&0.30&18.60&0.52&0.27 & 12.55&0.20&0.62 &13.15&0.20&0.59 &13.82&0.21&0.57\\
    ViewCrafter &18.31&0.51&0.26&20.89&0.61&0.14 & 11.54&0.19&0.63 &11.84&0.20&0.61 &12.25&0.21&0.59\\
    SEVA &\underline{19.45}&\underline{0.58}&\underline{0.15}&\underline{21.59}&\underline{0.64}&\underline{0.10} & \underline{13.89}&\underline{0.25}&\underline{0.49} &\underline{14.80}&\underline{0.26}&\underline{0.43} &\underline{15.07}&\underline{0.27}&\underline{0.41}\\
    Ours &\textbf{20.01}&\textbf{0.60}&\textbf{0.12}&\textbf{21.95}&\textbf{0.67}&\textbf{0.09} & \textbf{14.64}&\textbf{0.27}&\textbf{0.46} & \textbf{15.43} &\textbf{0.28} &\textbf{0.40} &\textbf{16.01}&\textbf{0.30}&\textbf{0.35}\\
    \bottomrule
    \end{tabular}
    }
\end{table*}

Following standard practice~\cite{cat3d,zhou2025stable}, we report PSNR, SSIM, and LPIPS. We compare against regression-based methods, including AnySplat, WorldMirror, and RayZer, as well as diffusion-based methods, including SEVA, Fillerbuster, and ViewCrafter. We note that methods such as Gen3C~\cite{ren2025gen3c} and Geometry Forcing~\cite{wu2025geometryforcing} require ground-truth camera poses and are therefore not directly comparable to our pose-free setting. ViewCrafter uses a separate evaluation setup detailed in the supplementary.

\paragraph{Geometric conditioning bridges the pose gap.}
SEVA (unposed) denotes the official SEVA model~\cite{zhou2025stable} run with DUSt3R-predicted poses provided by the SEVA codebase. As shown in \cref{tab:in_domain_results}, on RealEstate10K, SplatGuide improves over this strongest pose-only diffusion baseline by +3.05~dB at 3 views, and with 9 views it surpasses even the ground-truth-pose SEVA, showing that the complete framework compensates for the accuracy loss of predicted poses when sufficient views are available. The same trend holds on DL3DV, where SplatGuide achieves the best LPIPS across all view counts; RayZer reports higher PSNR at 6 and 9 views but with substantially weaker perceptual quality, \eg, LPIPS of 0.54 vs.\ our 0.31 at 6 views. As \cref{tab:out_of_domain_results} shows, on the out-of-domain Mip-NeRF 360 and Tanks and Temples benchmarks, SplatGuide remains the best among all unposed methods across view counts, confirming that geometric conditioning transfers to diverse scenes without domain-specific tuning. \cref{fig:grid7x4} corroborates these gains visually: SplatGuide produces sharper structures and fewer artifacts than both diffusion baselines.

\paragraph{Generation complements reconstruction.}
SplatGuide surpasses pure reconstruction baselines by large margins, \eg, +5.39~dB over WorldMirror and +6.71~dB over AnySplat on the RealEstate10K 3-view split. This gap confirms that reconstruction models alone can only interpolate observed regions and cannot hallucinate plausible content in unobserved areas, whereas our diffusion-based generation fills these regions with high fidelity guided by the reconstructed geometric prior. Moreover, the advantage persists on out-of-domain scenes, showing that the diffusion model's learned generative prior remains valuable when reconstruction quality degrades on unseen distributions.

\begin{figure}[!t]
\centering
\resizebox{0.75\textwidth}{!}{%
\begin{tikzpicture}[
    image/.style  = {inner sep=0pt, outer sep=1pt, anchor=north west},
    ctxcard/.style = {inner sep=0pt, outer sep=0pt, anchor=north east, draw=white, line width=1pt},
    node distance = 1pt and 1pt,
    every node/.style = {font={\tiny}},
    label/.style  = {font={\large\bfseries\vphantom{p}}, anchor=south, inner sep=1pt},
]

\node [image] (ren-0)
    {\includegraphics[height=\imageheight]{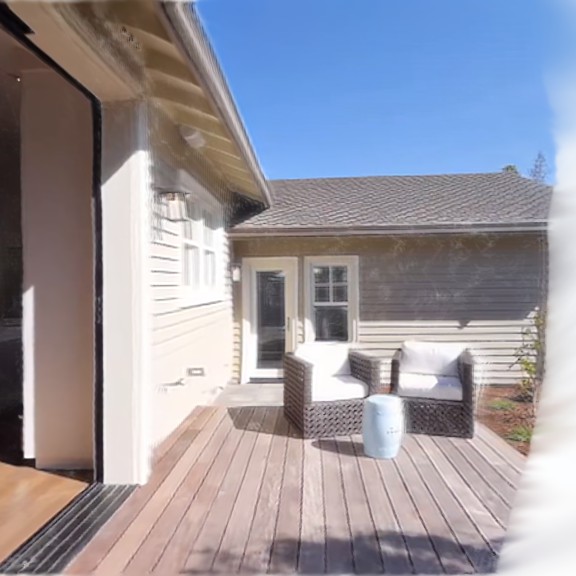}};
\node [image, right=of ren-0] (ren-1)
    {\includegraphics[height=\imageheight]{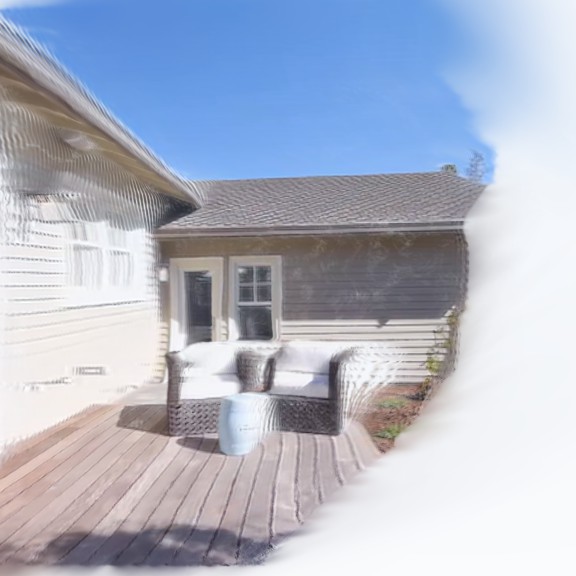}};
\node [image] at ($(ren-1.north east) + (6pt, 0)$) (gen-0)
    {\includegraphics[height=\imageheight]{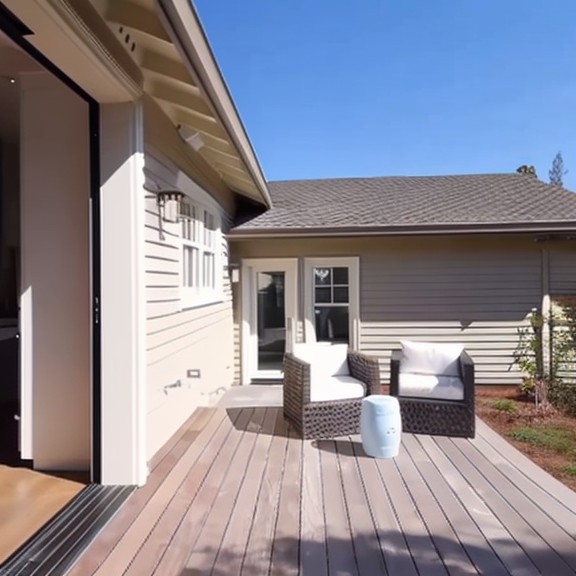}};
\node [image, right=of gen-0] (gen-1)
    {\includegraphics[height=\imageheight]{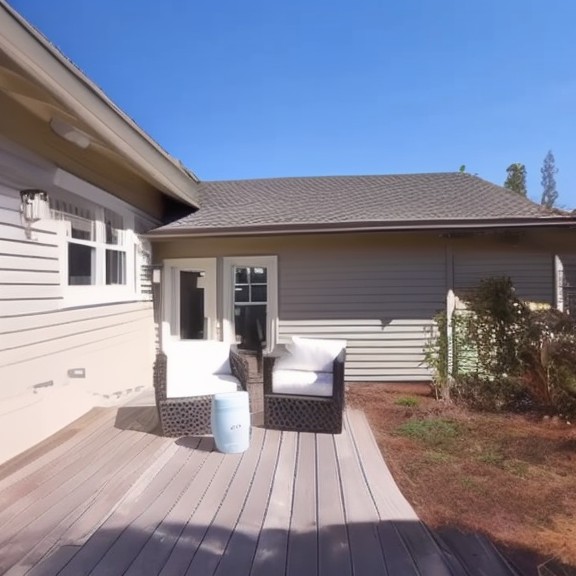}};

\node [ctxcard] at ($(ren-0.north west) + (-6pt, 0)$) (ctx-2)
    {\includegraphics[height=0.8\imageheight]{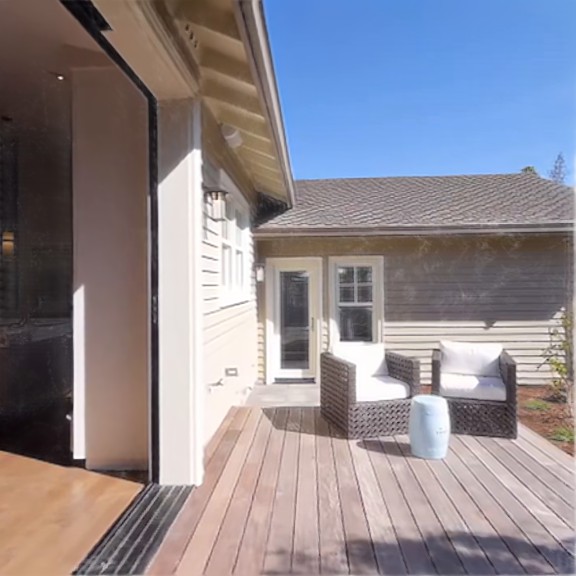}};
\node [ctxcard] at ($(ctx-2.north east) + (-0.1\imageheight, -0.1\imageheight)$) (ctx-1)
    {\includegraphics[height=0.8\imageheight]{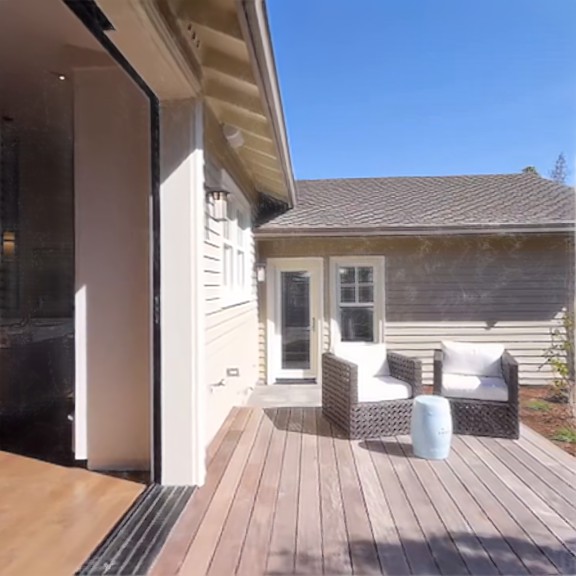}};
\node [ctxcard] at ($(ctx-1.north east) + (-0.1\imageheight, -0.1\imageheight)$) (ctx-0)
    {\includegraphics[height=0.8\imageheight]{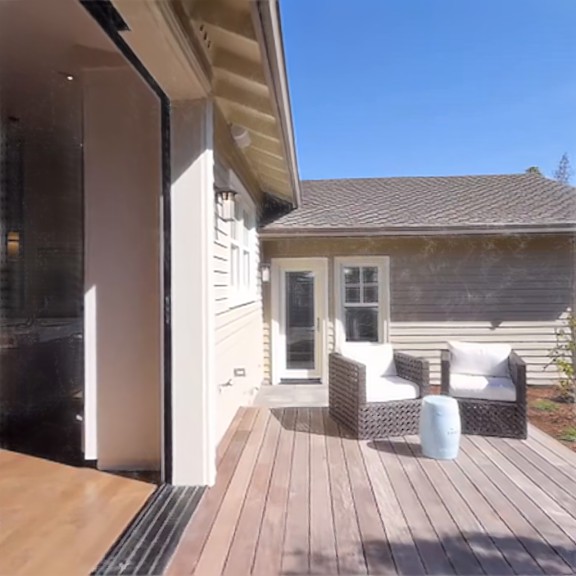}};

\node [image] at ($(ren-0.south west) + (0, -6pt)$) (ren0-r2)
    {\includegraphics[height=\imageheight]{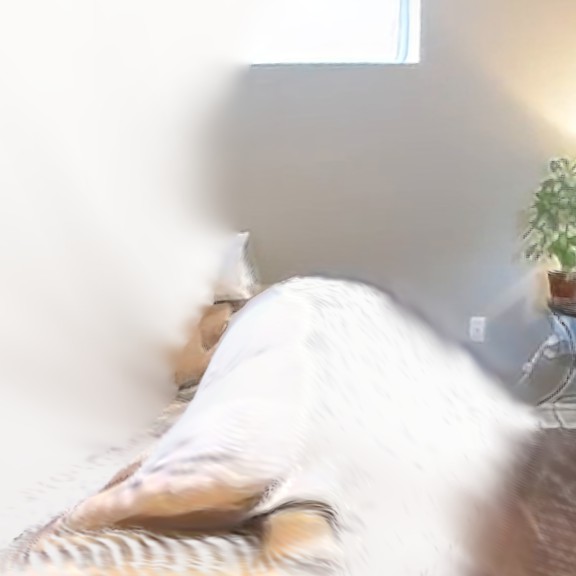}};
\node [image, right=of ren0-r2] (ren1-r2)
    {\includegraphics[height=\imageheight]{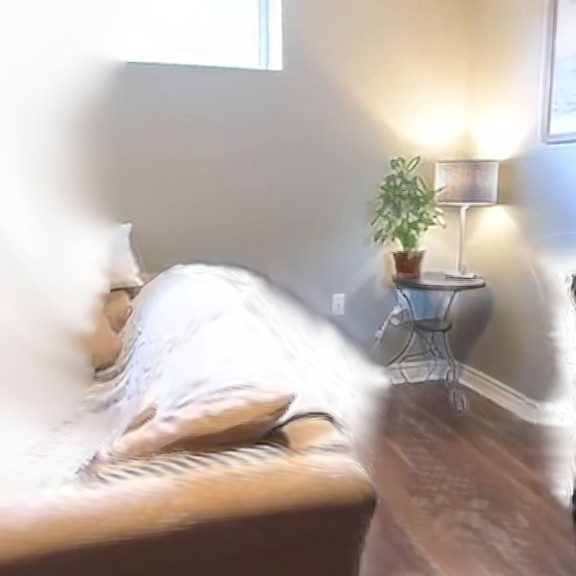}};
\node [image] at ($(ren1-r2.north east) + (6pt, 0)$) (gen0-r2)
    {\includegraphics[height=\imageheight]{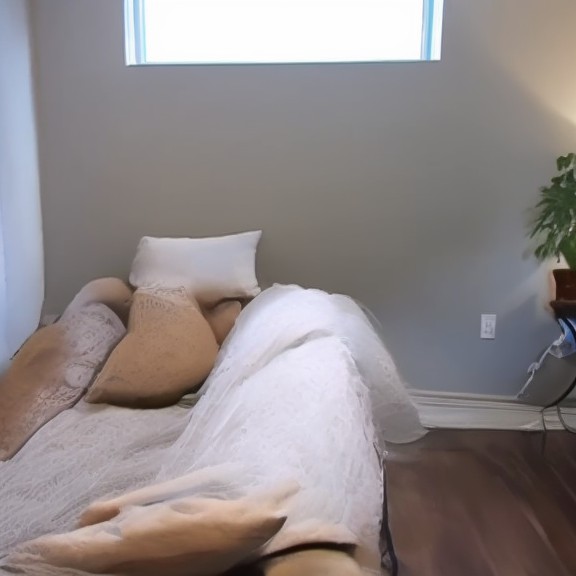}};
\node [image, right=of gen0-r2] (gen1-r2)
    {\includegraphics[height=\imageheight]{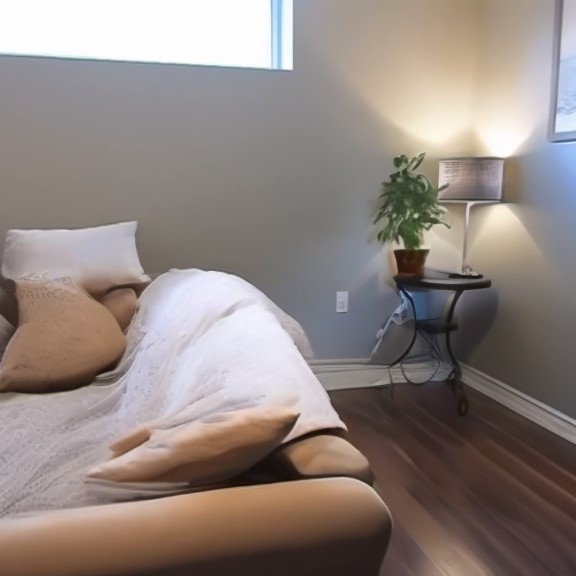}};

\node [ctxcard] at ($(ren0-r2.north west) + (-6pt, 0)$) (ctx1-r2)
    {\includegraphics[height=0.9\imageheight]{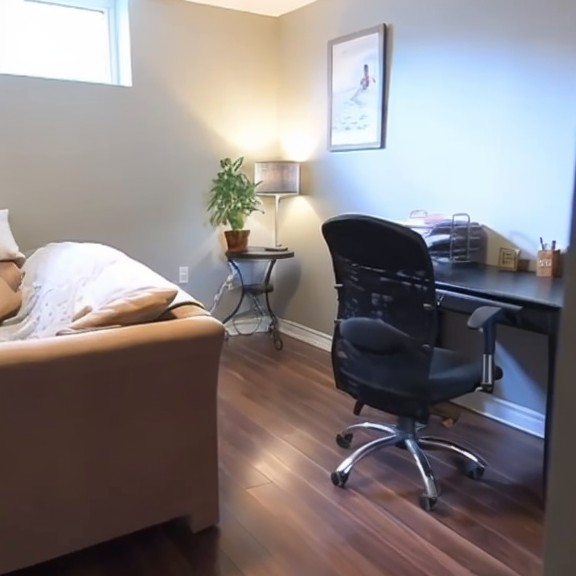}};
\node [ctxcard] at ($(ctx1-r2.north east) + (-0.1\imageheight, -0.1\imageheight)$) (ctx0-r2)
    {\includegraphics[height=0.9\imageheight]{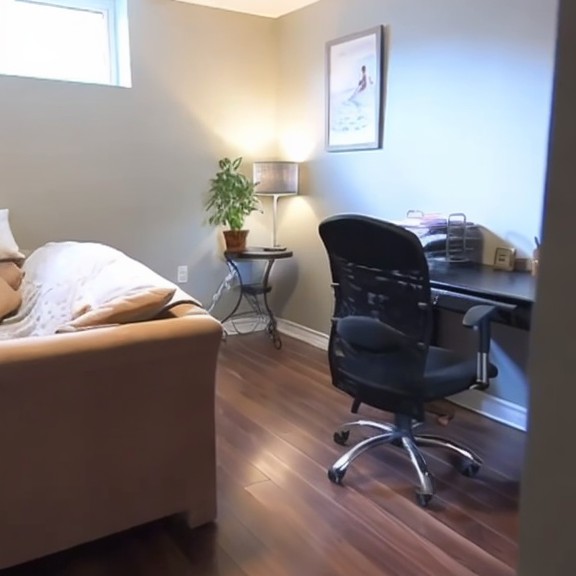}};

\node [label] at ($(ctx-0.west |- ctx-2.north)!0.5!(ctx-2.north east)$) {(a) Contexts};
\node [label] at ($(ren-0.north west)!0.5!(ren-1.north east)$) {(b) Rendered};
\node [label] at ($(gen-0.north west)!0.5!(gen-1.north east)$) {(c) Generated};

\end{tikzpicture}%
}
\caption{View extrapolation under large viewpoint changes. For each scene we show (a) the selected context views, (b) coarse 3DGS renders at target poses far from all context cameras, and (c) the corresponding generated novel views. The degraded renders still anchor the visible layout, which the diffusion model completes into coherent images.}
\label{fig:extrapolation}
\end{figure}
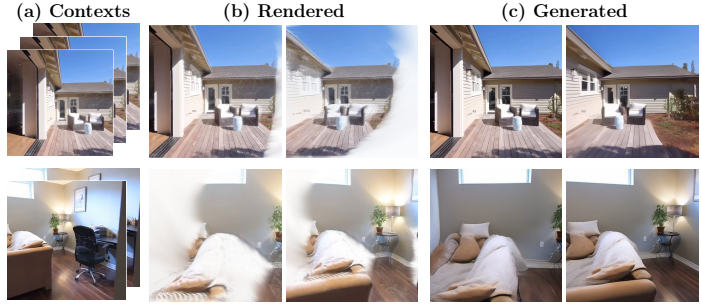

\paragraph{Extrapolation beyond observed viewpoints.}
Generative NVS matters most when targets depart substantially from all reference views, precisely the regime where render-and-refine pipelines are expected to struggle. \cref{fig:extrapolation} examines this setting: as the target leaves the observed trajectory, the 3DGS render turns sparse and incomplete, yet it still pins down the layout of the visible geometry, and the diffusion model completes the remaining content into a coherent novel view.

\subsection{Inference-Time View Selection}

To isolate the effect of selection policy from the generator, we fix the diffusion model and vary only the selection strategy. We compare against four reimplemented baselines: temporal recency \emph{Temporal}~\cite{dfot}, camera-distance ranking \emph{CamDist}~\cite{zhou2025stable}, surfel-based visibility coverage \emph{Surfel}~\cite{vmem}, and field-of-view overlap \emph{FoV}~\cite{worldmem}. All methods draw from the same pool of 32 candidate references under three context budgets: $B{=}6$ for the sparse regime, $B{=}9$ for the mid regime, and $B{=}16$ for the ample regime.

\paragraph{Quantitative Results.}
The results reveal a striking 7~dB range across selection policies on RealEstate10K at $B{=}6$. Our hybrid method achieves the best results across all budgets, with the largest gains in the sparse regime: +1.3~dB over the best existing strategy Surfel and +3.9~dB over CamDist at $B{=}6$. The advantage narrows at $B{=}16$, where the strongest spatial strategies converge within 1~dB: visibility-aware selection matters most precisely when the context budget is tight, the regime most relevant to practical deployment.

\begin{table*}[!t]
\caption{Comparison of view selection policies with a fixed generator. All methods use the same reconstruction and diffusion backbones; only the selection policy differs.}
\centering\footnotesize
\label{tab:selection-main}
\setlength{\tabcolsep}{4pt}
\resizebox{\textwidth}{!}{
\begin{tabular}{lcccccccccc}
\toprule
& \multicolumn{3}{c}{$B{=}6$ (sparse)} & \multicolumn{3}{c}{$B{=}9$ (mid)} & \multicolumn{3}{c}{$B{=}16$ (ample)} \\
\cmidrule(lr){2-4} \cmidrule(lr){5-7} \cmidrule(lr){8-10}
Method 
& PSNR$\uparrow$ & SSIM$\uparrow$ & LPIPS$\downarrow$
& PSNR$\uparrow$ & SSIM$\uparrow$ & LPIPS$\downarrow$
& PSNR$\uparrow$ & SSIM$\uparrow$ & LPIPS$\downarrow$ \\
\midrule
Temporal(DFoT) & 20.88 & 0.71 & 0.22 & 21.50 & 0.75 & 0.18 & 26.53 & 0.86 & 0.10 \\
CamDist(SEVA) & 24.34 & 0.80 & 0.11 & 25.44 & 0.81 & 0.10 & 27.71 & 0.88 & 0.07 \\
FoV(WorldMem) & 26.52 & 0.83 & \underline{0.09} & 27.61 & 0.84 & \underline{0.07} & \underline{29.25} & \underline{0.90} & \underline{0.06} \\
Surfel(Vmem) & \underline{26.93} & \underline{0.84} & \underline{0.09} & \underline{27.86} & \underline{0.85} & 0.08 & 29.12 & \underline{0.90} & 0.07 \\
Ours & \textbf{28.25} & \textbf{0.85} & \textbf{0.06} & \textbf{28.91} & \textbf{0.86} & \textbf{0.06} & \textbf{29.94} & \textbf{0.91} & \textbf{0.05} \\
\bottomrule
\end{tabular}
}
\end{table*}

\begin{figure}[!t]
\centering
\newlength{\selimgwidth}
\setlength{\selimgwidth}{0.115\linewidth}
\newlength{\spysize}
\setlength{\spysize}{0.48\selimgwidth}
\begin{tikzpicture}[
    image/.style = {
        inner sep=0pt, outer sep=0pt,
        anchor=north west
    },
    node distance = 1pt and 1pt,
    label/.style = {
        font={\footnotesize\bfseries},
        anchor=south,
        inner sep=1pt
    },
    spy using outlines={rectangle, magnification=1.8, size=\spysize, line width=0.1pt}
]
\node[image] (s1-gt)
    {\includegraphics[width=\selimgwidth]{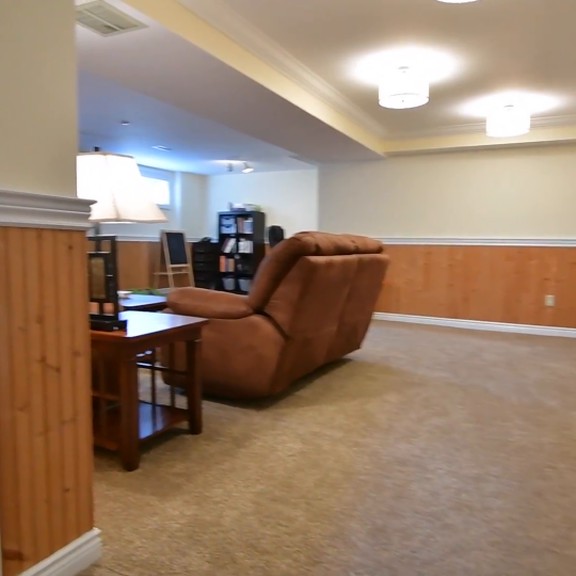}};
\spy [red] on ($(s1-gt.center) + (-0.25\selimgwidth, 0.20\selimgwidth)$) in node [right] at ($(s1-gt.east) + (0, 0.52\spysize)$);
\spy [blue] on ($(s1-gt.center) + (-0.32\selimgwidth, -0.20\selimgwidth)$) in node [right] at ($(s1-gt.east) + (0, -0.52\spysize)$);

\node[image, right=of s1-gt, xshift=1.05\spysize] (s1-ours)
    {\includegraphics[width=\selimgwidth]{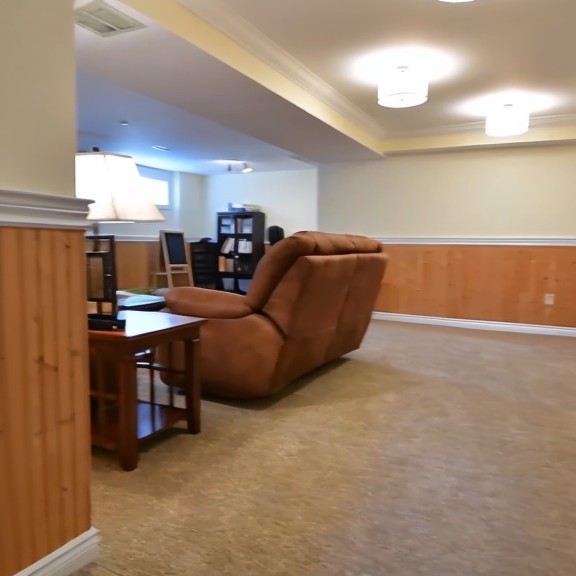}};
\spy [red] on ($(s1-ours.center) + (-0.25\selimgwidth, 0.20\selimgwidth)$) in node [right] at ($(s1-ours.east) + (0, 0.52\spysize)$);
\spy [blue] on ($(s1-ours.center) + (-0.32\selimgwidth, -0.20\selimgwidth)$) in node [right] at ($(s1-ours.east) + (0, -0.52\spysize)$);

\node[image, right=of s1-ours, xshift=1.05\spysize] (s1-seva)
    {\includegraphics[width=\selimgwidth]{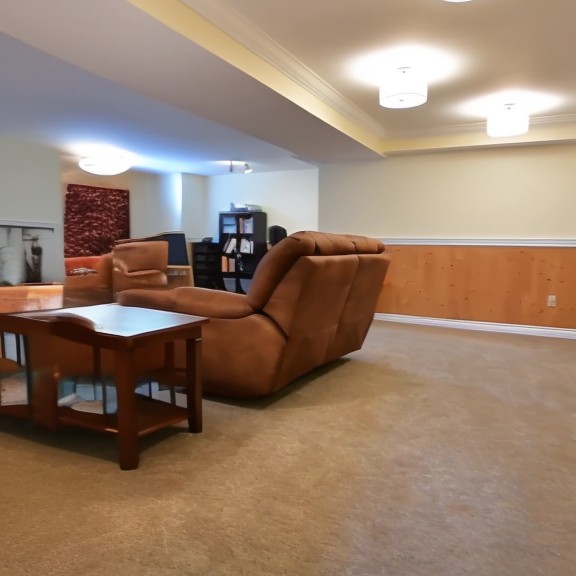}};
\spy [red] on ($(s1-seva.center) + (-0.25\selimgwidth, 0.20\selimgwidth)$) in node [right] at ($(s1-seva.east) + (0, 0.52\spysize)$);
\spy [blue] on ($(s1-seva.center) + (-0.32\selimgwidth, -0.20\selimgwidth)$) in node [right] at ($(s1-seva.east) + (0, -0.52\spysize)$);

\node[image, right=of s1-seva, xshift=1.05\spysize] (s1-vmem)
    {\includegraphics[width=\selimgwidth]{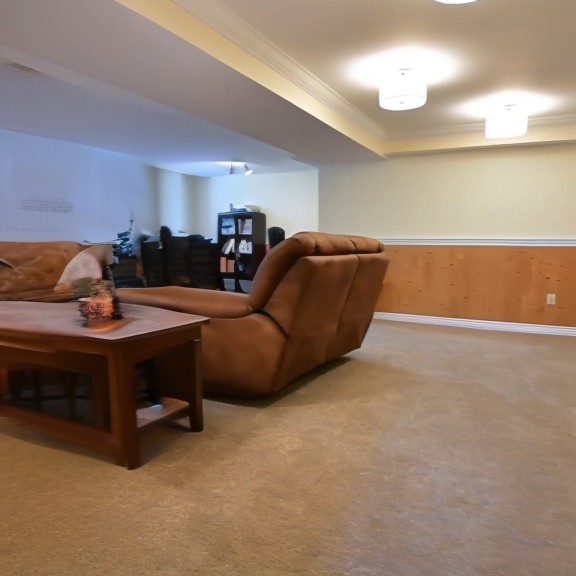}};
\spy [red] on ($(s1-vmem.center) + (-0.25\selimgwidth, 0.20\selimgwidth)$) in node [right] at ($(s1-vmem.east) + (0, 0.52\spysize)$);
\spy [blue] on ($(s1-vmem.center) + (-0.32\selimgwidth, -0.20\selimgwidth)$) in node [right] at ($(s1-vmem.east) + (0, -0.52\spysize)$);

\node[image, right=of s1-vmem, xshift=1.05\spysize] (s1-wm)
    {\includegraphics[width=\selimgwidth]{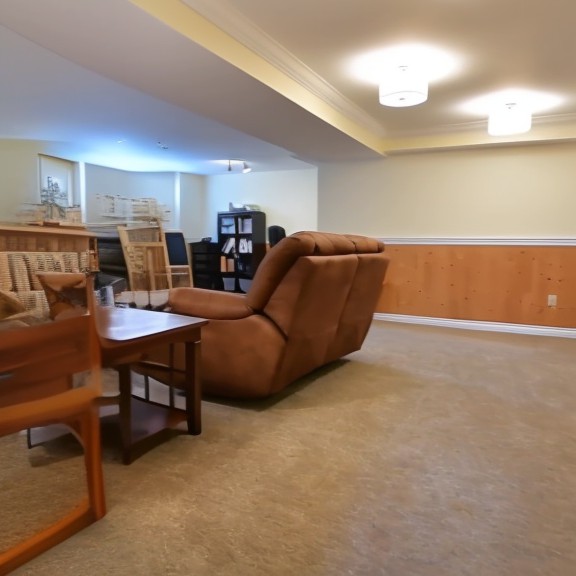}};
\spy [red] on ($(s1-wm.center) + (-0.25\selimgwidth, 0.20\selimgwidth)$) in node [right] at ($(s1-wm.east) + (0, 0.52\spysize)$);
\spy [blue] on ($(s1-wm.center) + (-0.32\selimgwidth, -0.20\selimgwidth)$) in node [right] at ($(s1-wm.east) + (0, -0.52\spysize)$);

\node[image, below=of s1-gt, yshift=-2pt] (s2-gt)
    {\includegraphics[width=\selimgwidth]{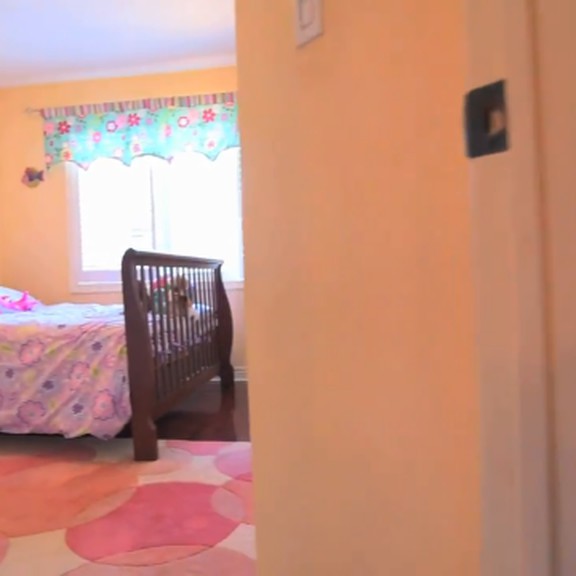}};
\spy [red] on ($(s2-gt.center) + (-0.32\selimgwidth, 0.22\selimgwidth)$) in node [right] at ($(s2-gt.east) + (0, 0.52\spysize)$);
\spy [blue] on ($(s2-gt.center) + (-0.15\selimgwidth, -0.18\selimgwidth)$) in node [right] at ($(s2-gt.east) + (0, -0.52\spysize)$);

\node[image, right=of s2-gt, xshift=1.05\spysize] (s2-ours)
    {\includegraphics[width=\selimgwidth]{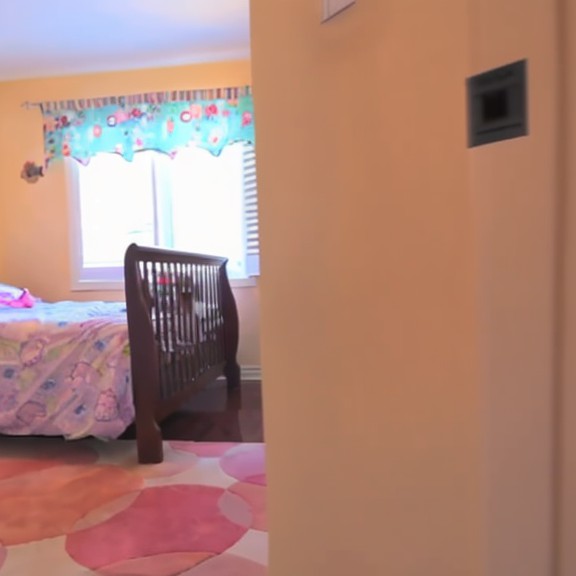}};
\spy [red] on ($(s2-ours.center) + (-0.32\selimgwidth, 0.22\selimgwidth)$) in node [right] at ($(s2-ours.east) + (0, 0.52\spysize)$);
\spy [blue] on ($(s2-ours.center) + (-0.15\selimgwidth, -0.18\selimgwidth)$) in node [right] at ($(s2-ours.east) + (0, -0.52\spysize)$);

\node[image, right=of s2-ours, xshift=1.05\spysize] (s2-seva)
    {\includegraphics[width=\selimgwidth]{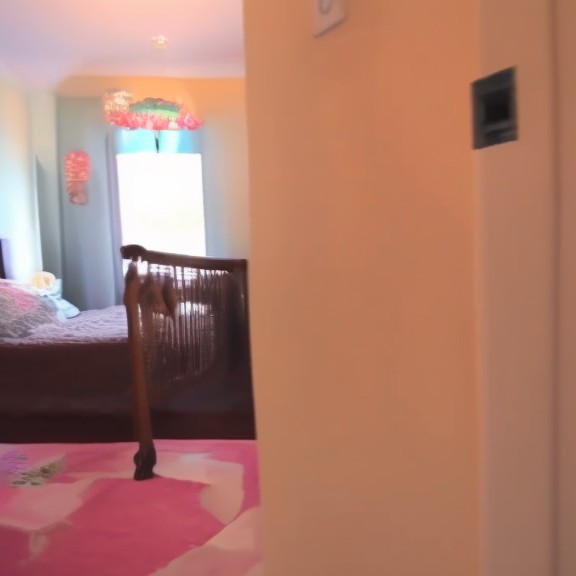}};
\spy [red] on ($(s2-seva.center) + (-0.32\selimgwidth, 0.22\selimgwidth)$) in node [right] at ($(s2-seva.east) + (0, 0.52\spysize)$);
\spy [blue] on ($(s2-seva.center) + (-0.15\selimgwidth, -0.18\selimgwidth)$) in node [right] at ($(s2-seva.east) + (0, -0.52\spysize)$);

\node[image, right=of s2-seva, xshift=1.05\spysize] (s2-vmem)
    {\includegraphics[width=\selimgwidth]{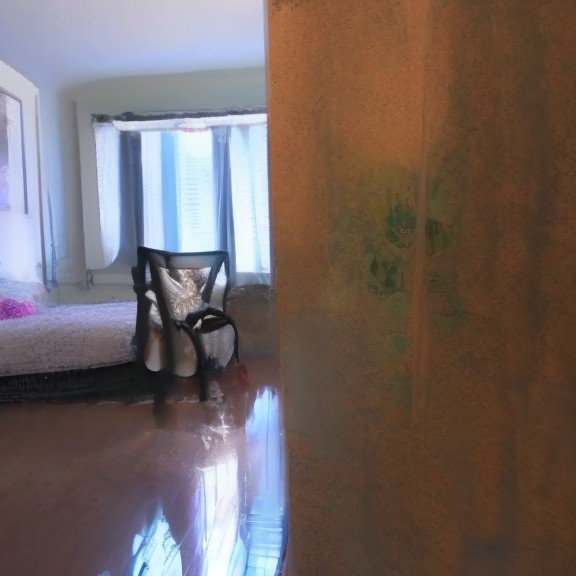}};
\spy [red] on ($(s2-vmem.center) + (-0.32\selimgwidth, 0.22\selimgwidth)$) in node [right] at ($(s2-vmem.east) + (0, 0.52\spysize)$);
\spy [blue] on ($(s2-vmem.center) + (-0.15\selimgwidth, -0.18\selimgwidth)$) in node [right] at ($(s2-vmem.east) + (0, -0.52\spysize)$);

\node[image, right=of s2-vmem, xshift=1.05\spysize] (s2-wm)
    {\includegraphics[width=\selimgwidth]{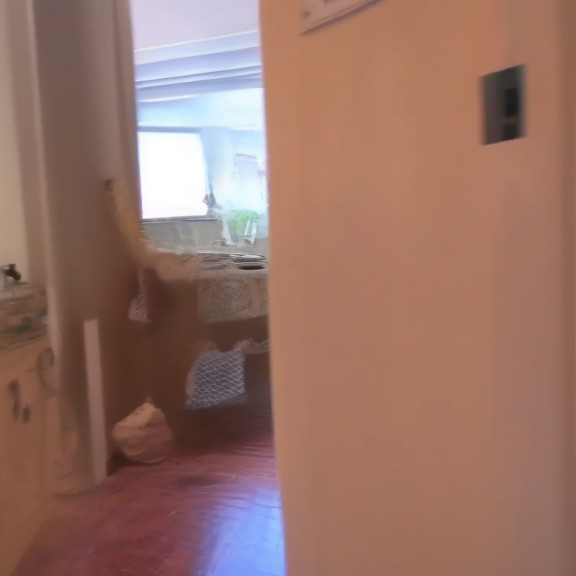}};
\spy [red] on ($(s2-wm.center) + (-0.32\selimgwidth, 0.22\selimgwidth)$) in node [right] at ($(s2-wm.east) + (0, 0.52\spysize)$);
\spy [blue] on ($(s2-wm.center) + (-0.15\selimgwidth, -0.18\selimgwidth)$) in node [right] at ($(s2-wm.east) + (0, -0.52\spysize)$);

\draw[dashed,gray,thick]
  ($ (s2-gt.south west) + (-0.05,-0.08) $)
  --
  ($ (s2-wm.south east) + (1.05\spysize,-0.08) $);

\node[image, below=of s2-gt, yshift=-5pt] (s3-gt)
    {\includegraphics[width=\selimgwidth]{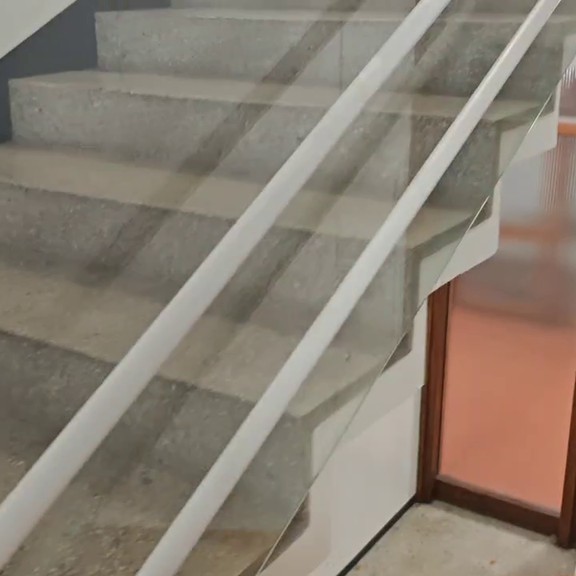}};
\spy [red] on ($(s3-gt.center) + (0.20\selimgwidth, 0.12\selimgwidth)$) in node [right] at ($(s3-gt.east) + (0, 0.52\spysize)$);
\spy [blue] on ($(s3-gt.center) + (-0.16\selimgwidth, -0.16\selimgwidth)$) in node [right] at ($(s3-gt.east) + (0, -0.52\spysize)$);

\node[image, right=of s3-gt, xshift=1.05\spysize] (s3-ours)
    {\includegraphics[width=\selimgwidth]{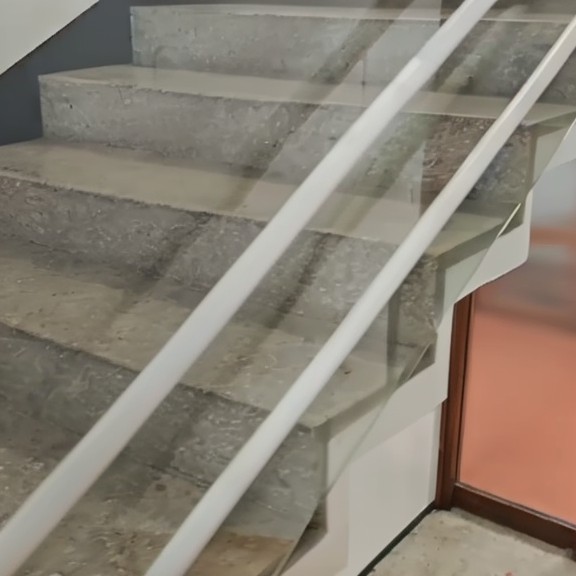}};
\spy [red] on ($(s3-ours.center) + (0.20\selimgwidth, 0.12\selimgwidth)$) in node [right] at ($(s3-ours.east) + (0, 0.52\spysize)$);
\spy [blue] on ($(s3-ours.center) + (-0.16\selimgwidth, -0.16\selimgwidth)$) in node [right] at ($(s3-ours.east) + (0, -0.52\spysize)$);

\node[image, right=of s3-ours, xshift=1.05\spysize] (s3-seva)
    {\includegraphics[width=\selimgwidth]{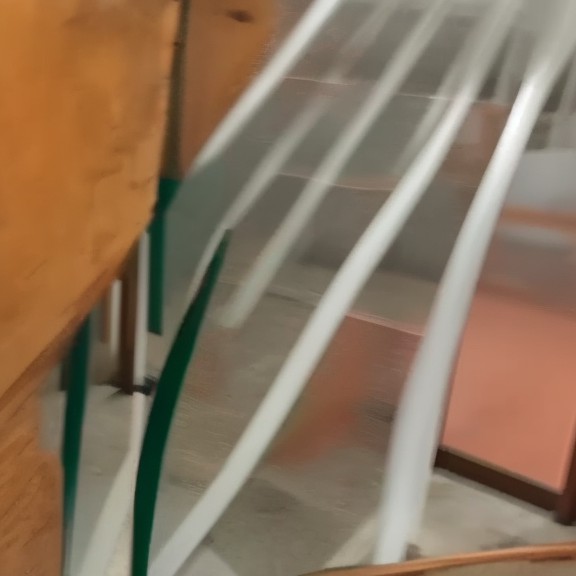}};
\spy [red] on ($(s3-seva.center) + (0.20\selimgwidth, 0.12\selimgwidth)$) in node [right] at ($(s3-seva.east) + (0, 0.52\spysize)$);
\spy [blue] on ($(s3-seva.center) + (-0.16\selimgwidth, -0.16\selimgwidth)$) in node [right] at ($(s3-seva.east) + (0, -0.52\spysize)$);

\node[image, right=of s3-seva, xshift=1.05\spysize] (s3-vmem)
    {\includegraphics[width=\selimgwidth]{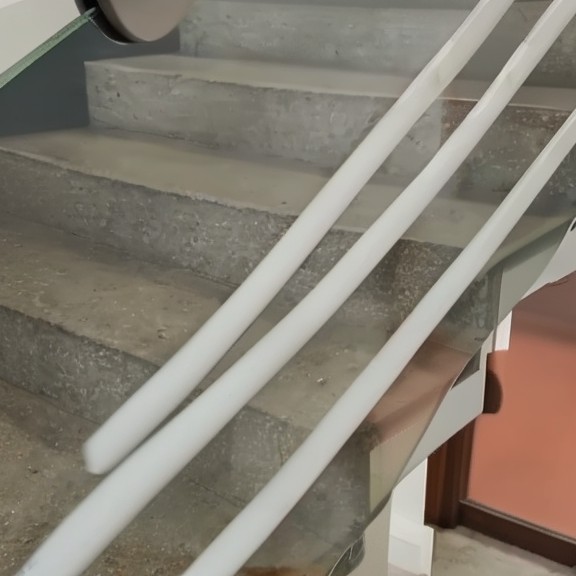}};
\spy [red] on ($(s3-vmem.center) + (0.20\selimgwidth, 0.12\selimgwidth)$) in node [right] at ($(s3-vmem.east) + (0, 0.52\spysize)$);
\spy [blue] on ($(s3-vmem.center) + (-0.16\selimgwidth, -0.16\selimgwidth)$) in node [right] at ($(s3-vmem.east) + (0, -0.52\spysize)$);

\node[image, right=of s3-vmem, xshift=1.05\spysize] (s3-wm)
    {\includegraphics[width=\selimgwidth]{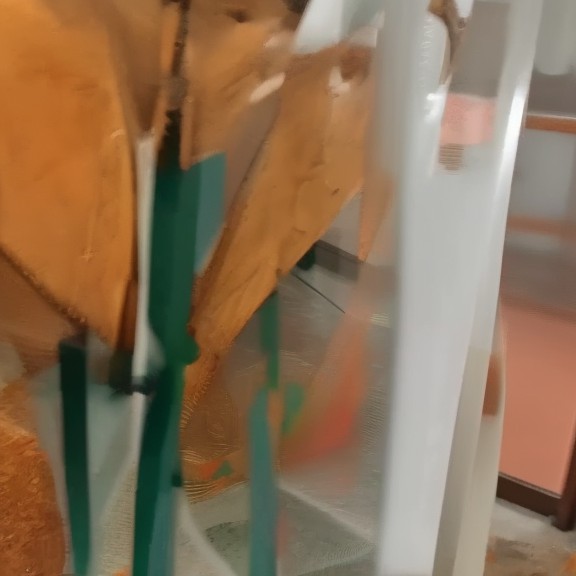}};
\spy [red] on ($(s3-wm.center) + (0.20\selimgwidth, 0.12\selimgwidth)$) in node [right] at ($(s3-wm.east) + (0, 0.52\spysize)$);
\spy [blue] on ($(s3-wm.center) + (-0.16\selimgwidth, -0.16\selimgwidth)$) in node [right] at ($(s3-wm.east) + (0, -0.52\spysize)$);

\node[image, below=of s3-gt, yshift=-2pt] (s4-gt)
    {\includegraphics[width=\selimgwidth]{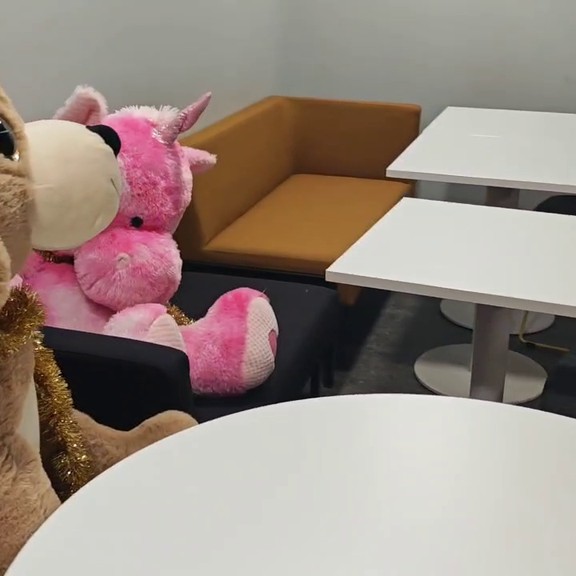}};
\spy [red] on ($(s4-gt.center) + (-0.17\selimgwidth, 0.14\selimgwidth)$) in node [right] at ($(s4-gt.east) + (0, 0.52\spysize)$);
\spy [blue] on ($(s4-gt.center) + (-0.19\selimgwidth, -0.36\selimgwidth)$) in node [right] at ($(s4-gt.east) + (0, -0.52\spysize)$);

\node[image, right=of s4-gt, xshift=1.05\spysize] (s4-ours)
    {\includegraphics[width=\selimgwidth]{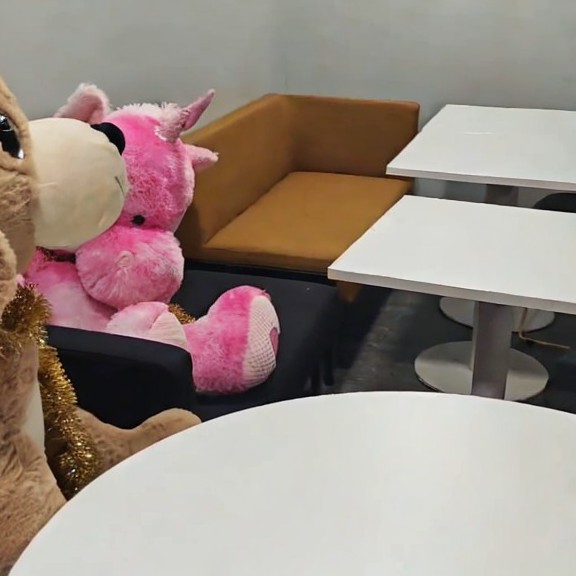}};
\spy [red] on ($(s4-ours.center) + (-0.17\selimgwidth, 0.14\selimgwidth)$) in node [right] at ($(s4-ours.east) + (0, 0.52\spysize)$);
\spy [blue] on ($(s4-ours.center) + (-0.19\selimgwidth, -0.36\selimgwidth)$) in node [right] at ($(s4-ours.east) + (0, -0.52\spysize)$);

\node[image, right=of s4-ours, xshift=1.05\spysize] (s4-seva)
    {\includegraphics[width=\selimgwidth]{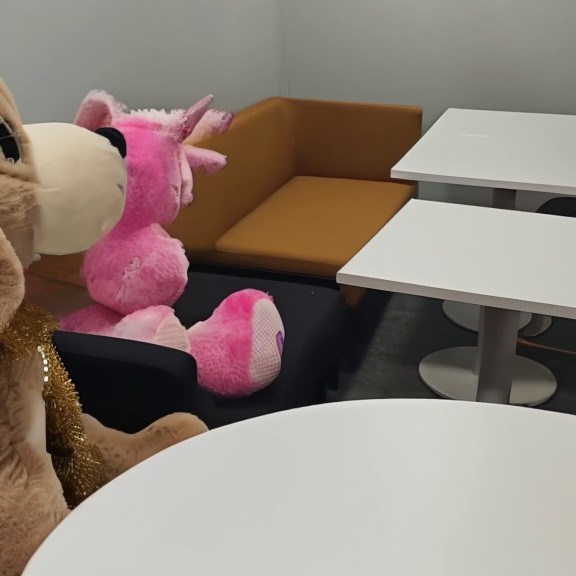}};
\spy [red] on ($(s4-seva.center) + (-0.17\selimgwidth, 0.14\selimgwidth)$) in node [right] at ($(s4-seva.east) + (0, 0.52\spysize)$);
\spy [blue] on ($(s4-seva.center) + (-0.19\selimgwidth, -0.36\selimgwidth)$) in node [right] at ($(s4-seva.east) + (0, -0.52\spysize)$);

\node[image, right=of s4-seva, xshift=1.05\spysize] (s4-vmem)
    {\includegraphics[width=\selimgwidth]{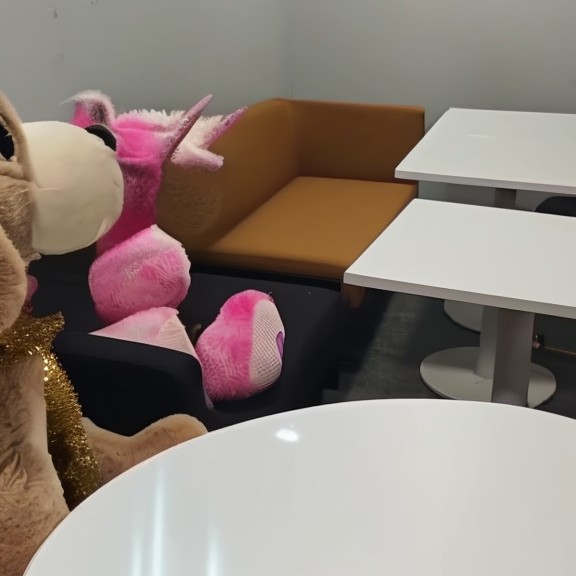}};
\spy [red] on ($(s4-vmem.center) + (-0.17\selimgwidth, 0.14\selimgwidth)$) in node [right] at ($(s4-vmem.east) + (0, 0.52\spysize)$);
\spy [blue] on ($(s4-vmem.center) + (-0.19\selimgwidth, -0.36\selimgwidth)$) in node [right] at ($(s4-vmem.east) + (0, -0.52\spysize)$);

\node[image, right=of s4-vmem, xshift=1.05\spysize] (s4-wm)
    {\includegraphics[width=\selimgwidth]{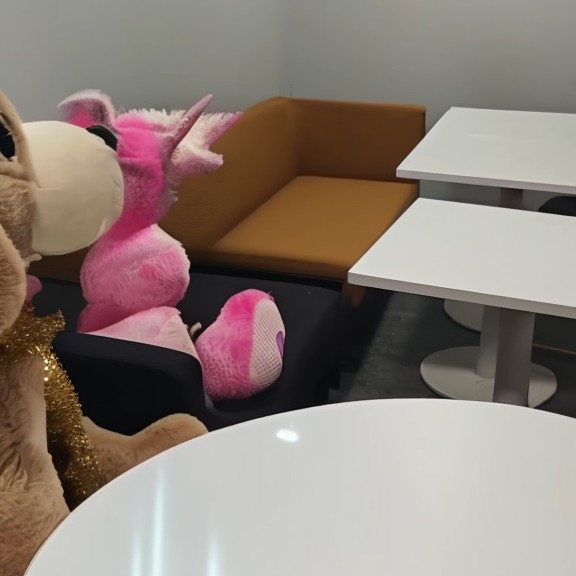}};
\spy [red] on ($(s4-wm.center) + (-0.17\selimgwidth, 0.14\selimgwidth)$) in node [right] at ($(s4-wm.east) + (0, 0.52\spysize)$);
\spy [blue] on ($(s4-wm.center) + (-0.19\selimgwidth, -0.36\selimgwidth)$) in node [right] at ($(s4-wm.east) + (0, -0.52\spysize)$);

\node[label, xshift=0.52\spysize] at (s1-gt.north) {Ground Truth};
\node[label, xshift=0.52\spysize] at (s1-ours.north) {Ours};
\node[label, xshift=0.52\spysize] at (s1-seva.north) {SEVA};
\node[label, xshift=0.52\spysize] at (s1-vmem.north) {VMem};
\node[label, xshift=0.52\spysize] at (s1-wm.north) {WorldMem};

\end{tikzpicture}
\caption{Qualitative comparison of view-selection policies. Top two rows: our method recovers fine details by selecting complementary views. Bottom two rows: our visibility-aware selection suppresses occluder-dominated candidates and reveals hidden geometry.}
\label{fig:selection}
\end{figure}
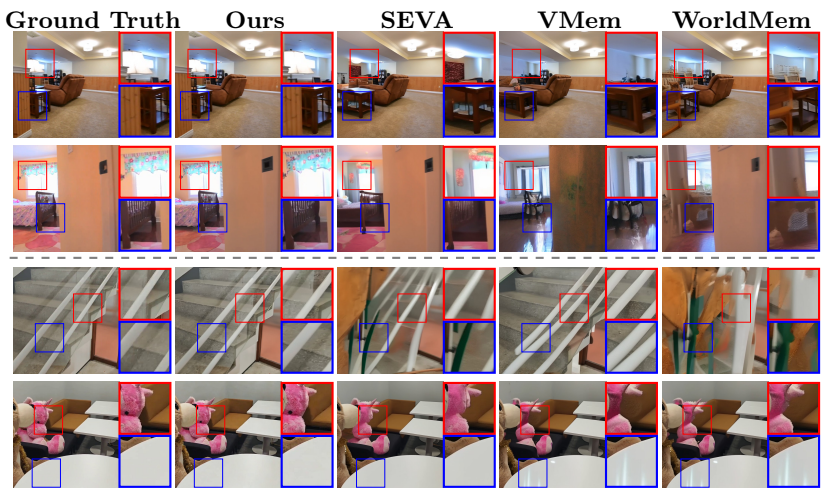

\paragraph{Qualitative Observations.}
\cref{fig:selection} illustrates two representative failure modes. In the top two rows, pose-based baselines select clusters of nearly identical views, wasting the budget on redundant content and causing texture bleeding near depth discontinuities, while the surfel-based VMem selector over-concentrates on already well-covered surfaces; our method selects complementary views, producing sharper edges and fewer missing regions. In the bottom two rows, strong foreground occluders dominate the frames chosen by pose-only methods, whereas our first-hit voting suppresses such candidates and retrieves views that see around the obstruction. These patterns are consistent across scenes and align with the quantitative gaps in \cref{tab:selection-main}.

\subsection{Ablation of Geometric Conditioning}
\begin{table}[!t]
    \centering
    \small
    \caption{Ablation study on geometric conditioning on DL3DV with 9 views. We progressively add rendering images, camera tokens and register tokens to the baseline to measure each component's contribution.}
    \label{tab:ablation_structure}
    \begin{tabular}{l c c c}
        \toprule
        Method & PSNR $\uparrow$ & SSIM $\uparrow$ & LPIPS $\downarrow$ \\
        \midrule
        Baseline + predicted pose & 15.64 & 0.33 & 0.38 \\
        + rendering image & 16.30 & 0.35 & 0.33 \\
        + rendering image + cam \& reg & \textbf{16.63} & \textbf{0.36} & \textbf{0.32} \\
        \bottomrule
    \end{tabular}
\end{table}
\cref{tab:ablation_structure} presents a staged ablation that progressively adds geometric conditioning components. To reduce computational cost, all variants are trained and evaluated on a representative subset of DL3DV with 9 reference views.

Starting from the predicted-pose baseline, adding the rendering image alone yields a +0.66~dB gain, confirming that rendering serves as the primary geometric anchor. Further adding camera and register tokens yields the best overall quality at 16.63 PSNR and 0.32 LPIPS, a 16\% relative LPIPS reduction from the baseline. Token-level guidance thus complements pixel-level rendering rather than replacing it: renderings anchor the spatial layout, while tokens supply global scene context that resolves texture and style ambiguities.


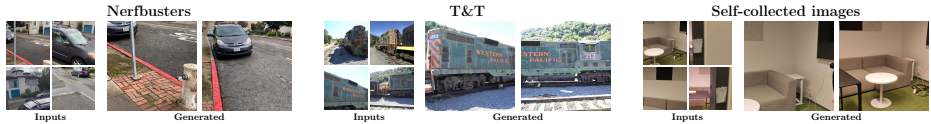
\begin{figure}[!t]
\centering
\resizebox{\textwidth}{!}{%
\begin{tikzpicture}[
    image/.style  = {inner sep=0pt, outer sep=1pt, anchor=north west},
    node distance = 1pt and 1pt,
    every node/.style = {font={\tiny}},
    label/.style  = {font={\Large\bfseries\vphantom{p}}, anchor=south, inner sep=1pt},
    sublabel/.style = {font={\footnotesize\bfseries}, anchor=north, inner sep=2pt},
]

\node [image] (s1c1) {\includegraphics[height=0.48\imageheight]{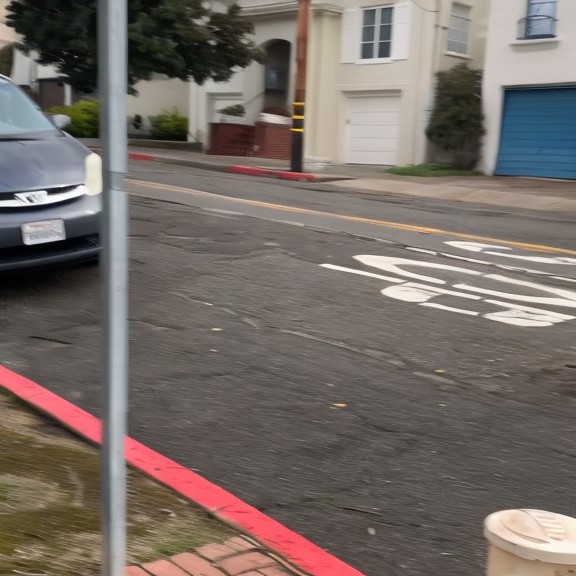}};
\node [image, right=of s1c1] (s1c2)
    {\includegraphics[height=0.48\imageheight]{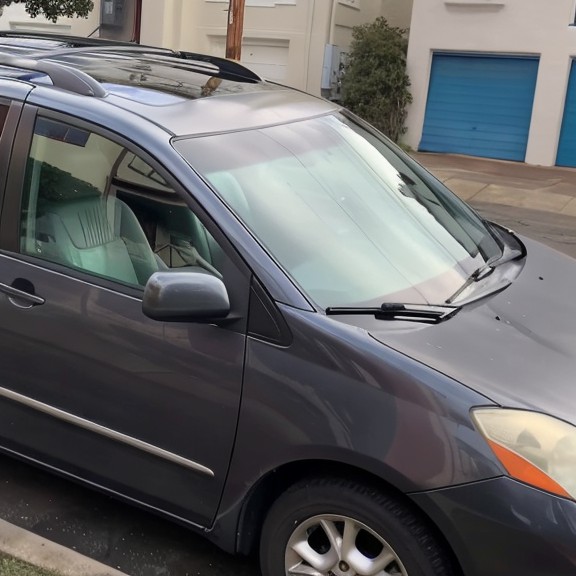}};
\node [image, below=of s1c1] (s1c3)
    {\includegraphics[height=0.48\imageheight]{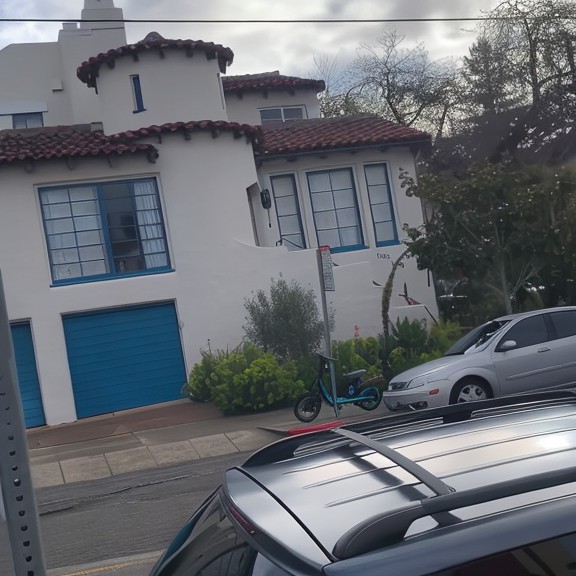}};
\node [image, right=of s1c3] (s1c4)
    {\includegraphics[height=0.48\imageheight]{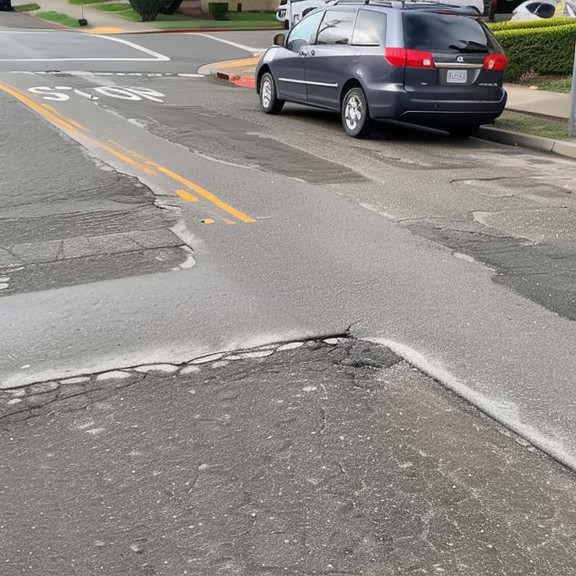}};
\node [image] at ($(s1c2.north east) + (10pt, 0)$) (s1gen-1)
    {\includegraphics[height=\imageheight]{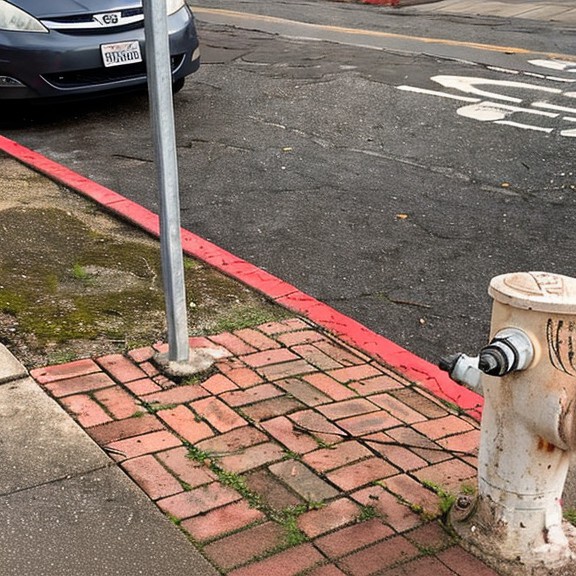}};
\node [image] at ($(s1gen-1.north east) + (4pt, 0)$) (s1gen-3)
    {\includegraphics[height=\imageheight]{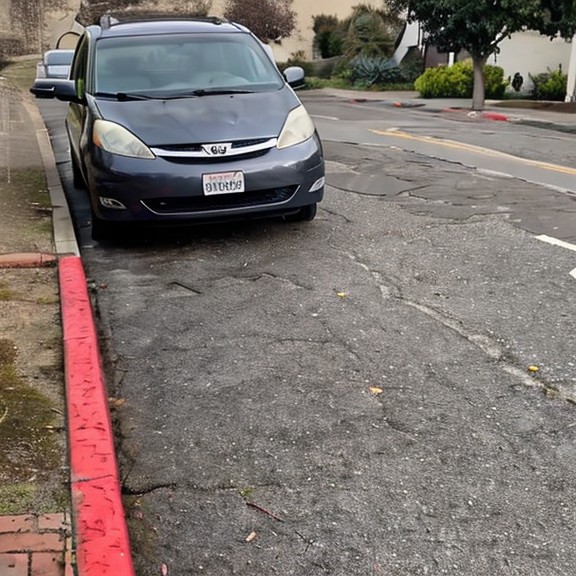}};

\node [image] at ($(s1gen-3.north east) + (30pt, 0)$) (s2c1)
    {\includegraphics[height=0.48\imageheight]{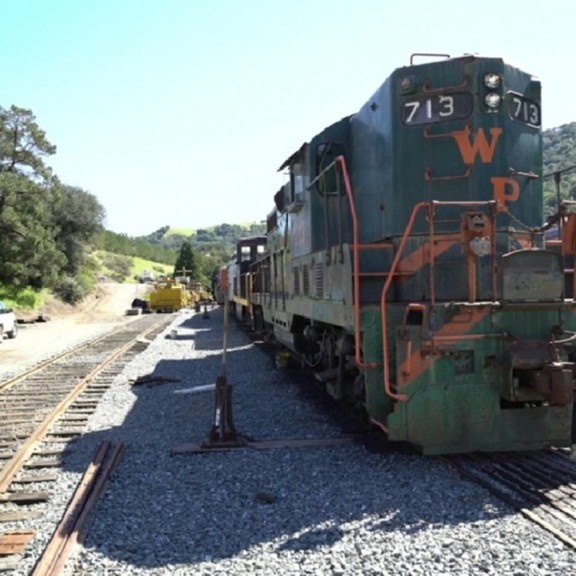}};
\node [image, right=of s2c1] (s2c2)
    {\includegraphics[height=0.48\imageheight]{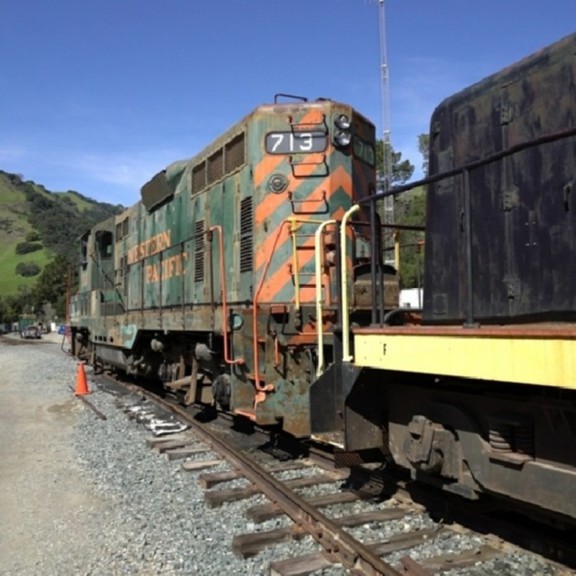}};
\node [image, below=of s2c1] (s2c3)
    {\includegraphics[height=0.48\imageheight]{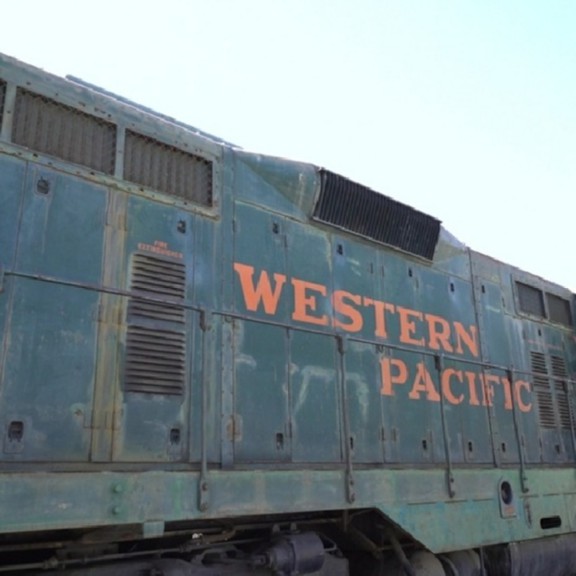}};
\node [image, right=of s2c3] (s2c4)
    {\includegraphics[height=0.48\imageheight]{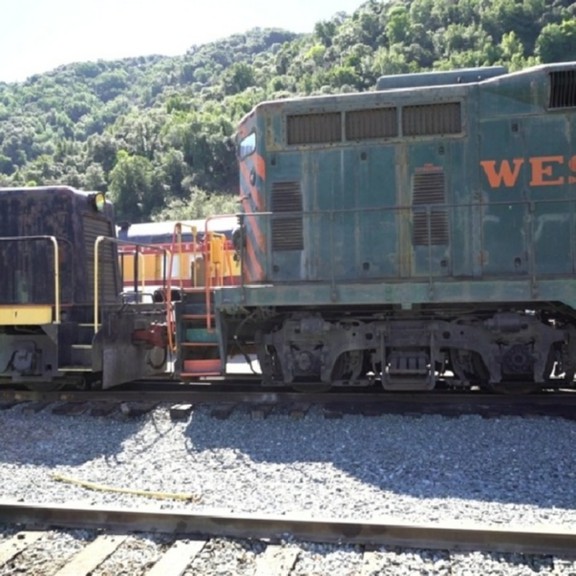}};
\node [image] at ($(s2c2.north east) + (10pt, 0)$) (s2gen-1)
    {\includegraphics[height=\imageheight]{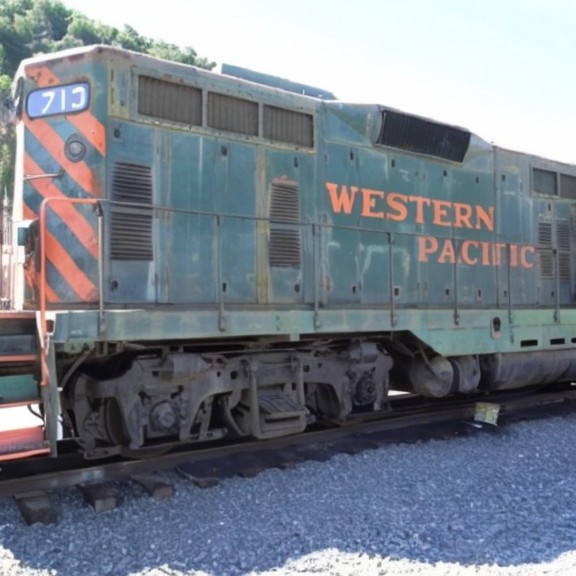}};
\node [image] at ($(s2gen-1.north east) + (4pt, 0)$) (s2gen-3)
    {\includegraphics[height=\imageheight]{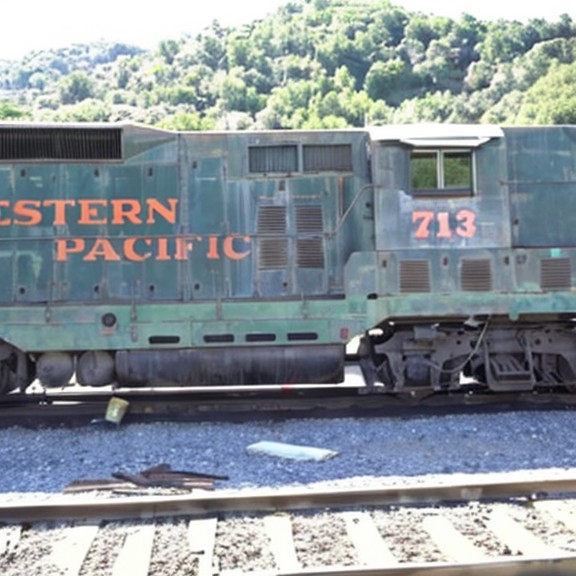}};

\node [image] at ($(s2gen-3.north east) + (30pt, 0)$) (s3c1)
    {\includegraphics[height=0.48\imageheight]{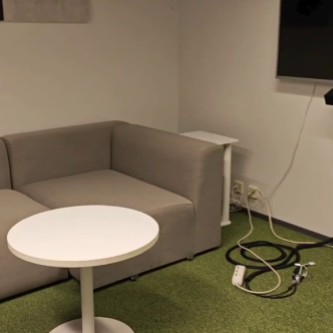}};
\node [image, right=of s3c1] (s3c2)
    {\includegraphics[height=0.48\imageheight]{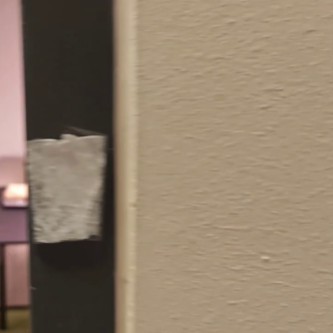}};
\node [image, below=of s3c1] (s3c3)
    {\includegraphics[height=0.48\imageheight]{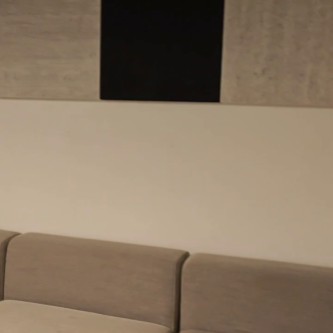}};
\node [image, right=of s3c3] (s3c4)
    {\includegraphics[height=0.48\imageheight]{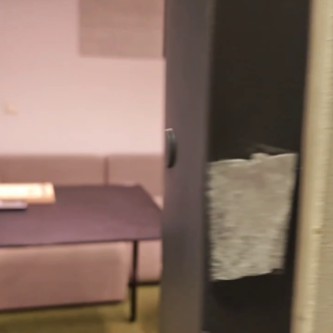}};
\node [image] at ($(s3c2.north east) + (10pt, 0)$) (s3gen-1)
    {\includegraphics[height=\imageheight]{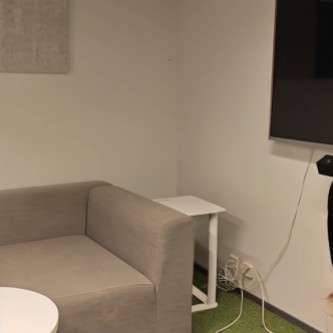}};
\node [image] at ($(s3gen-1.north east) + (4pt, 0)$) (s3gen-3)
    {\includegraphics[height=\imageheight]{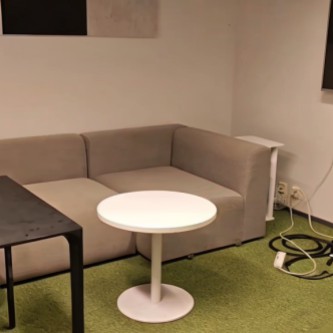}};

\node [label] at ($(s1c1.north west)!0.5!(s1gen-3.north east)$) {Nerfbusters};
\node [label] at ($(s2c1.north west)!0.5!(s2gen-3.north east)$) {T\&T};
\node [label] at ($(s3c1.north west)!0.5!(s3gen-3.north east)$) {Self-collected images};

\node [sublabel] at ($(s1c3.south east)!0.5!(s1c4.south west)$) {Inputs};
\node [sublabel] at ($(s1gen-1.south east)!0.5!(s1gen-3.south west)$) {Generated};
\node [sublabel] at ($(s2c3.south east)!0.5!(s2c4.south west)$) {Inputs};
\node [sublabel] at ($(s2gen-1.south east)!0.5!(s2gen-3.south west)$) {Generated};
\node [sublabel] at ($(s3c3.south east)!0.5!(s3c4.south west)$) {Inputs};
\node [sublabel] at ($(s3gen-1.south east)!0.5!(s3gen-3.south west)$) {Generated};

\end{tikzpicture}%
}
\caption{Novel view synthesis on large-scale scenes with 100+ input images. Each group shows four example inputs (left $2{\times}2$ grid) and two generated novel views (right), on Nerfbusters~\cite{warburg2023nerfbusters}, Tanks-and-Temples, and self-collected casual captures.}
\label{fig:large-scale}
\end{figure}

\paragraph{Scaling to Larger Candidate Pools.}
Casual captures routinely yield pools far larger than the 32 candidates used above, stressing both the selector and the palette encoding, whose colors for a growing view count $V$ lie ever closer in RGB space. On long-trajectory Tanks-and-Temples scenes from the Long-LRM split~\cite{ziwen2025llrm}, we fix the budget to $B{=}9$ and enlarge the pool from 32 to 128 candidates. As \cref{tab:selection-scale} shows, quality rises monotonically with pool size, by +1.15~dB PSNR overall, with no palette-decoding failures even at $V{=}128$: with more candidates available, each target finds contexts nearby, so the generator bridges a shorter extrapolation gap. \cref{fig:large-scale} illustrates this regime on casually captured scenes and public benchmarks with over 100 input images.

\begin{table}[!t]
\centering
\small
\caption{Scaling the candidate pool on long-trajectory Tanks-and-Temples scenes (Long-LRM split) with a fixed context budget $B{=}9$. Larger pools consistently improve quality without degrading palette-encoded index recovery.}
\label{tab:selection-scale}
\setlength{\tabcolsep}{8pt}
\begin{tabular}{lcccc}
\toprule
Pool size $V$ & 32 & 64 & 96 & 128 \\
\midrule
PSNR$\uparrow$    & 16.06 & 16.40 & 16.87 & \textbf{17.21} \\
SSIM$\uparrow$    & 0.41  & 0.42  & 0.43  & \textbf{0.47} \\
LPIPS$\downarrow$ & 0.28  & 0.25  & 0.25  & \textbf{0.23} \\
\bottomrule
\end{tabular}
\end{table}

\subsection{Generalizability to Different Reconstruction Models}
\label{sec:reconstruction_dependency}

To test whether the framework depends on a specific reconstruction backbone, we replace WorldMirror~\cite{liu2025worldmirror} with AnySplat~\cite{jiang2025anysplat} zero-shot, without any fine-tuning of the diffusion model, and evaluate on the identical RealEstate10K~\cite{zhou2018stereo} test split with all other components held constant.

\begin{table}[!t]
    \centering
    \caption{Zero-shot generalization across reconstruction backbones on RealEstate10K. Our diffusion model, trained exclusively with WorldMirror priors, successfully transfers to AnySplat without retraining.}
    \label{tab:dependency_reconstruction_model}
    \setlength{\tabcolsep}{3pt}
    \resizebox{\linewidth}{!}{
    \begin{tabular}{lccccccccc}
    \toprule
    & \multicolumn{3}{c}{\textit{3-view}} & \multicolumn{3}{c}{\textit{6-view}} & \multicolumn{3}{c}{\textit{9-view}} \\
    \cmidrule(lr){2-4} \cmidrule(lr){5-7} \cmidrule(lr){8-10}
    Method & PSNR$\uparrow$ & SSIM$\uparrow$ & LPIPS$\downarrow$ & PSNR$\uparrow$ & SSIM$\uparrow$ & LPIPS$\downarrow$ & PSNR$\uparrow$ & SSIM$\uparrow$ & LPIPS$\downarrow$ \\
    \midrule
    Ours (AnySplat) & \textbf{26.56} & \underline{0.83} & \underline{0.08} & \textbf{29.97} & \textbf{0.87} & \underline{0.05} & \textbf{31.29} & \textbf{0.88} & \textbf{0.04} \\
    Ours (WorldMirror) & \underline{26.52} & \textbf{0.84} & \textbf{0.07} & \underline{29.29} & \textbf{0.87} & \textbf{0.04} & \underline{30.00} & \textbf{0.88} & \textbf{0.04} \\
    \bottomrule
    \end{tabular}
    }
\end{table}

As shown in \cref{tab:dependency_reconstruction_model}, the AnySplat variant achieves competitive or superior performance across all view counts, with gains of up to +1.29~dB at 9 views. Any feed-forward model producing camera poses and a pixel-aligned 3DGS scene can thus serve as a drop-in replacement, and a stronger backbone translates directly into better generation quality: the modular separation between reconstruction and generation lets the framework absorb future advances in feed-forward reconstruction without retraining.

\paragraph{Portability of the Conditioning Interface.}
The conditioning interface is largely backbone-agnostic: rendered images enter through channel-wise concatenation in the latent space, which any latent diffusion model supports by expanding its input convolution with zero-initialized weights, and reconstruction tokens enter through cross-attention layers present in most U-Net and DiT architectures. The visibility-aware selector runs entirely upstream of the generator, so it transfers to any backbone with a bounded context budget. Only the Pl\"ucker-and-mask input layout and the context-window length follow SEVA's design; we adopt SEVA for its strong pose-conditioned prior, and validating other multi-view diffusion backbones is left to future work.

\section{Conclusion}

We present SplatGuide, a framework for pose-free novel view synthesis that repurposes a single 3DGS reconstruction as a unified interface for pixel-level rendering, feature-level token guidance, and visibility-aware view selection. By systematically bridging the information disconnect between reconstruction and generation, SplatGuide closes the gap between predicted and ground-truth poses, even surpassing the ground-truth-pose baseline on RealEstate10K with sufficient input views, and demonstrates that visibility-aware selection is a first-order design decision for scalable generation.
A key strength of our design is its modularity: reconstruction and generation are fully decoupled, so each component can be upgraded independently, as validated by our zero-shot backbone substitution experiment, and future advances in feed-forward reconstruction translate directly into higher-quality synthesis.

\paragraph{Limitations.}
Our failure modes are correlated: renderings, tokens, and source-view indices all come from the same reconstruction $\mathcal{G}$, so wherever the feed-forward backbone degrades, such as on textureless surfaces, wide baselines, repetitive structures, or dynamic content, all three signals degrade at once and the generator has no independent geometric cue left. Dynamic scenes are the sharpest case, as motion both corrupts the reconstruction and breaks the first-hit correspondence behind the view-index map.

\subsubsection*{Acknowledgements.}
We acknowledge funding by Nokia Technologies, the Research Council of Finland (projects 339730, 352788, 353138, 353139, 362407, 362408, 362409, 372999, 373778, 373780, 373997, 373999), and the Finnish Doctoral Program Network in Artificial Intelligence, AI-DOC (decision number VN/3137/2024-OKM-6). We acknowledge CSC -- IT Center for Science, Finland, and the Aalto Science-IT project for the computational resources.
\newpage

\bibliographystyle{splncs04}
\bibliography{main}


\end{document}


\title{SplatGuide: Supplementary Material}
\titlerunning{SplatGuide: Supplementary Material}
\author{}
\authorrunning{}
\institute{}
\maketitle

\appendix
\renewcommand{\thetable}{A\arabic{table}}
\renewcommand{\thefigure}{A\arabic{figure}}

\section{Additional Implementation Details}
We adopt a dual-resolution strategy: inputs are resized to $448 \times 448$ for the reconstruction backbone to balance computational cost, while the diffusion model and 3DGS rendering operate at $576 \times 576$ to match the SEVA baseline.
For benchmark evaluation, we first align the target camera poses with the reconstructed scene. We use WorldMirror to place the reference and target cameras in the same coordinate system. We reconstruct the 3DGS using the reference RGBs and the aligned reference poses. We then render the 3DGS at the aligned target poses. Target RGBs are used only for camera alignment and metric computation. They are not used to build the 3DGS or provide features to the generation model.
We initialize the diffusion model from the pre-trained SEVA weights, with all newly added layers zero-initialized and the VAE frozen throughout training. We train for 25,000 steps on 8 H200 GPUs. At inference time, we use the DDIM sampler with 50 steps and a classifier-free guidance scale of 2.0, following the default SEVA configuration. We set the context window length to $T{=}21$ following SEVA, where $T$ is the total number of reference and target views, unless otherwise specified.

\section{Additional Evaluation Details}

\paragraph{RayZer and Matrix3D}
We evaluate RayZer~\cite{jiang2025rayzer} using provided checkpoints trained on DL3DV with 16 context views at $256 \times 256$ resolution. Since its image-index embeddings make it sensitive to view counts, we pad our inputs to match the required 16 views.
While RayZer achieves high PSNR, these metrics are inflated by the low evaluation resolution and do not reflect superior quality. Visual analysis uncovers significant mosaic-like artifacts, indicating that input padding fails to resolve the model's structural sensitivity to mismatches between training and testing view counts. We also evaluate Matrix3D~\cite{lu2025matrix3d}, which processes input images at $896 \times 896$ resolution and generates outputs at $512 \times 512$. Due to its maximum context limit of 8 views, we restrict our evaluation to the 3-view and 6-view splits. Unlike our fully unposed framework, Matrix3D requires ground-truth camera intrinsics as input; we therefore provide these intrinsics during testing.

\paragraph{Camera Pose Error}
\cref{tab:camera pose error} reports camera pose accuracy using Rotation Error ${\bf R}_{err}$ and Translation Error ${\bf T}_{err}$. We first align the predicted poses ${\cal P}_{\rm pred}$ with the ground truth ${\cal P}_{\rm gt}$ by setting the first frame to identity and rescaling predicted translations to match the ground-truth scale. The scale factor is the median ratio of ground-truth to predicted translation norms, computed over reference frames only. After alignment, $\mathbf{R}_{err}$ measures the mean geodesic distance between rotation matrices and $\mathbf{T}_{err}$ measures the mean Euclidean distance between translation vectors over all $N$ frames:

\begin{align}
{\bf R}_{err} &= \frac{1}{N} \sum_{i=1}^{N} \frac{180}{\pi} \arccos\left(\frac{\text{tr}(\mathbf{R}_{\text{gt}}^{(i)T} \mathbf{R}_{\text{pred}}^{(i)}) - 1}{2}\right) \\
{\bf T}_{err} &= \frac{1}{N} \sum_{i=1}^{N} ||\mathbf{t}_{\text{pred}}^{(i)} - \mathbf{t}_{\text{gt}}^{(i)}||_2
\end{align}

\begin{table}[!t]
    \centering
    \small
    \caption{Camera pose error comparison between DUSt3R and WorldMirror as reconstruction backbones.}
    \label{tab:camera pose error}
    \begin{tabular}{l c c}
        \toprule
        Method & ${\bf R}_{err}$($^\circ$) $\downarrow$ & ${\bf T}_{err}$
        $\downarrow$ \\
        \midrule
        DUSt3R &  \underline{0.69} & \underline{0.06} \\
        WorldMirror & \textbf{0.27} & \textbf{0.02}\\
        \bottomrule
    \end{tabular}
\end{table}

\section{Additional Qualitative Results}

\begin{figure}[!t]
\centering
\newlength{\supimgwidth}
\setlength{\supimgwidth}{0.16\linewidth}
\newlength{\supspysize}
\setlength{\supspysize}{0.48\supimgwidth}
\begin{tikzpicture}[
    image/.style = {
        inner sep=0pt,
        outer sep=0pt,
        anchor=north west
    },
    node distance = 1pt and 1pt,
    label/.style = {
        font={\footnotesize\bfseries},
        anchor=south,
        inner sep=1pt
    },
    spy using outlines={rectangle, magnification=1.8, size=\supspysize, line width=0.1pt}
]

\node[image] (img-00)
  {\includegraphics[width=\supimgwidth]{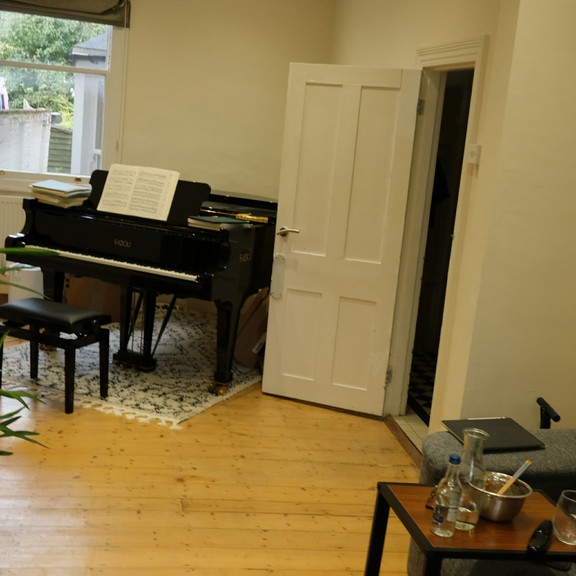}};
\spy [red] on ($(img-00.center) + (0.08\supimgwidth, 0.30\supimgwidth)$) in node [right] at ($(img-00.east) + (0, 0.52\supspysize)$);
\spy [blue] on ($(img-00.center) + (0.35\supimgwidth, -0.32\supimgwidth)$) in node [right] at ($(img-00.east) + (0, -0.52\supspysize)$);

\node[image, right=of img-00, xshift=1.05\supspysize] (img-01)
  {\includegraphics[width=\supimgwidth]{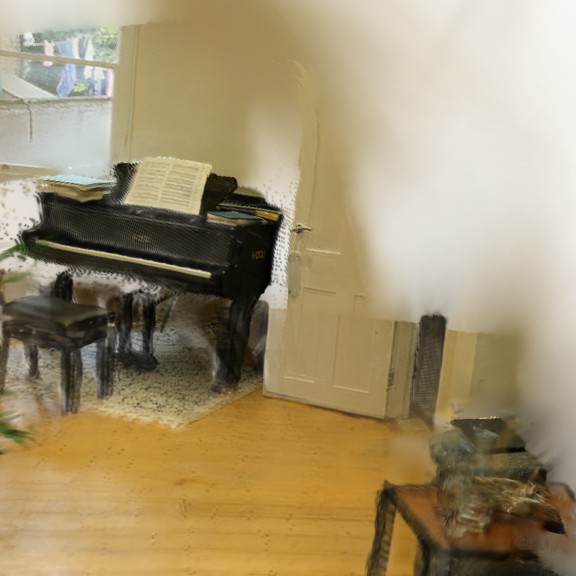}};
\spy [red] on ($(img-01.center) + (0.08\supimgwidth, 0.30\supimgwidth)$) in node [right] at ($(img-01.east) + (0, 0.52\supspysize)$);
\spy [blue] on ($(img-01.center) + (0.35\supimgwidth, -0.32\supimgwidth)$) in node [right] at ($(img-01.east) + (0, -0.52\supspysize)$);

\node[image, right=of img-01, xshift=1.05\supspysize] (img-02)
  {\includegraphics[width=\supimgwidth]{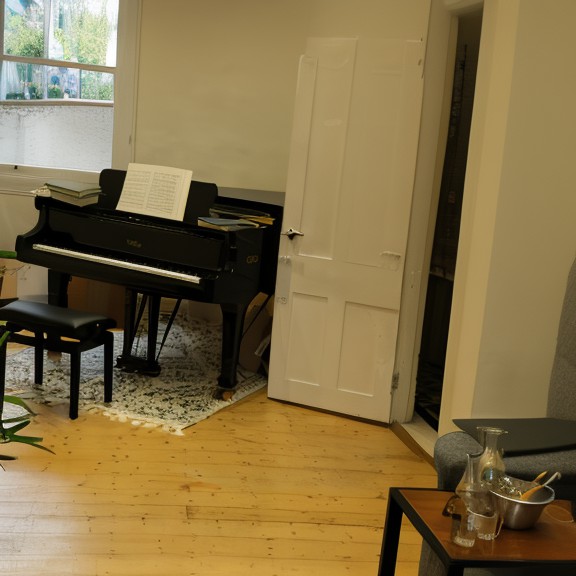}};
\spy [red] on ($(img-02.center) + (0.08\supimgwidth, 0.30\supimgwidth)$) in node [right] at ($(img-02.east) + (0, 0.52\supspysize)$);
\spy [blue] on ($(img-02.center) + (0.35\supimgwidth, -0.32\supimgwidth)$) in node [right] at ($(img-02.east) + (0, -0.52\supspysize)$);

\node[image, right=of img-02, xshift=1.05\supspysize] (img-03)
  {\includegraphics[width=\supimgwidth]{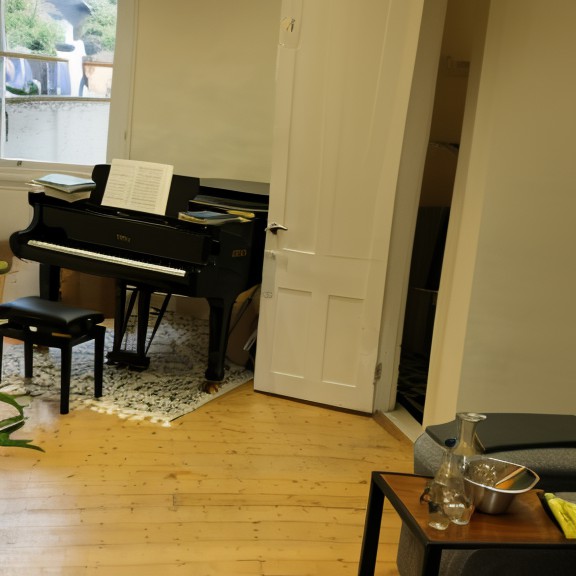}};
\spy [red] on ($(img-03.center) + (0.08\supimgwidth, 0.30\supimgwidth)$) in node [right] at ($(img-03.east) + (0, 0.52\supspysize)$);
\spy [blue] on ($(img-03.center) + (0.35\supimgwidth, -0.32\supimgwidth)$) in node [right] at ($(img-03.east) + (0, -0.52\supspysize)$);

\node[image, below=of img-00] (img-10)
  {\includegraphics[width=\supimgwidth]{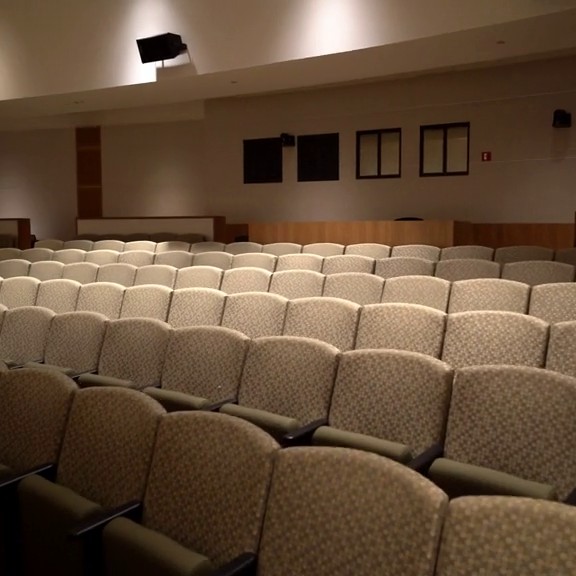}};
\spy [red] on ($(img-10.center) + (-0.16\supimgwidth, -0.14\supimgwidth)$) in node [right] at ($(img-10.east) + (0, 0.52\supspysize)$);
\spy [blue] on ($(img-10.center) + (-0.36\supimgwidth, -0.36\supimgwidth)$) in node [right] at ($(img-10.east) + (0, -0.52\supspysize)$);

\node[image, right=of img-10, xshift=1.05\supspysize] (img-11)
  {\includegraphics[width=\supimgwidth]{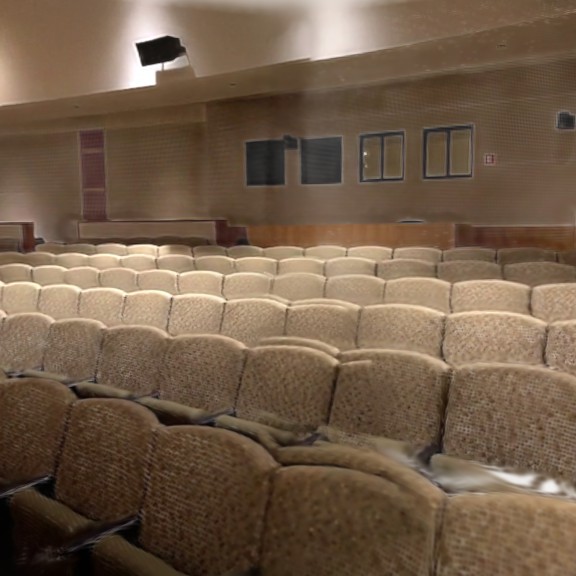}};
\spy [red] on ($(img-11.center) + (-0.16\supimgwidth, -0.14\supimgwidth)$) in node [right] at ($(img-11.east) + (0, 0.52\supspysize)$);
\spy [blue] on ($(img-11.center) + (-0.36\supimgwidth, -0.36\supimgwidth)$) in node [right] at ($(img-11.east) + (0, -0.52\supspysize)$);

\node[image, right=of img-11, xshift=1.05\supspysize] (img-12)
  {\includegraphics[width=\supimgwidth]{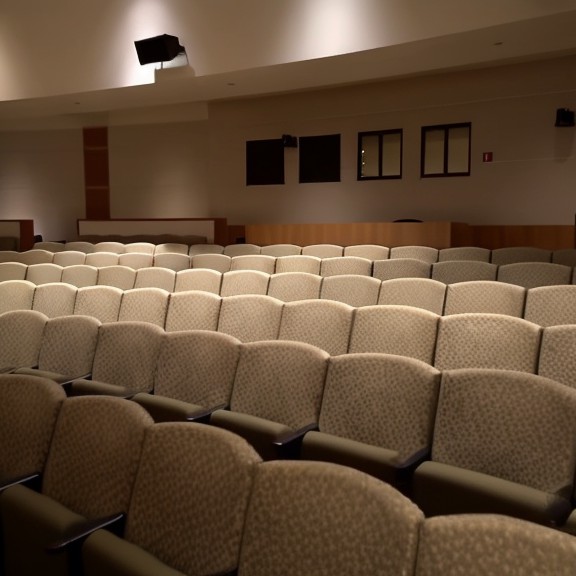}};
\spy [red] on ($(img-12.center) + (-0.16\supimgwidth, -0.14\supimgwidth)$) in node [right] at ($(img-12.east) + (0, 0.52\supspysize)$);
\spy [blue] on ($(img-12.center) + (-0.36\supimgwidth, -0.36\supimgwidth)$) in node [right] at ($(img-12.east) + (0, -0.52\supspysize)$);

\node[image, right=of img-12, xshift=1.05\supspysize] (img-13)
  {\includegraphics[width=\supimgwidth]{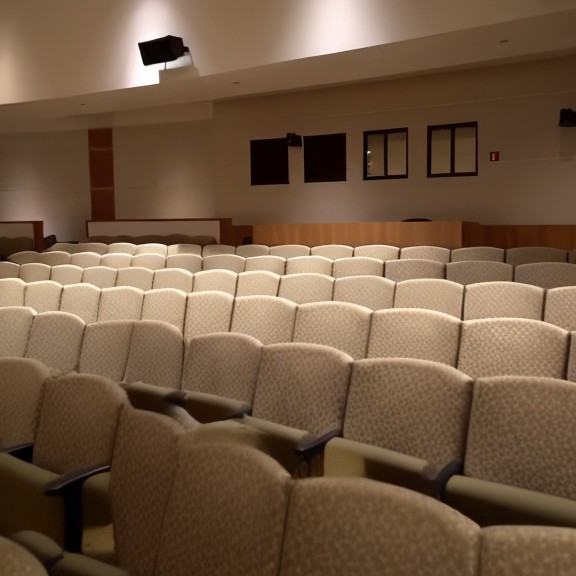}};
\spy [red] on ($(img-13.center) + (-0.16\supimgwidth, -0.14\supimgwidth)$) in node [right] at ($(img-13.east) + (0, 0.52\supspysize)$);
\spy [blue] on ($(img-13.center) + (-0.36\supimgwidth, -0.36\supimgwidth)$) in node [right] at ($(img-13.east) + (0, -0.52\supspysize)$);

\node[image, below=of img-10] (img-20)
  {\includegraphics[width=\supimgwidth]{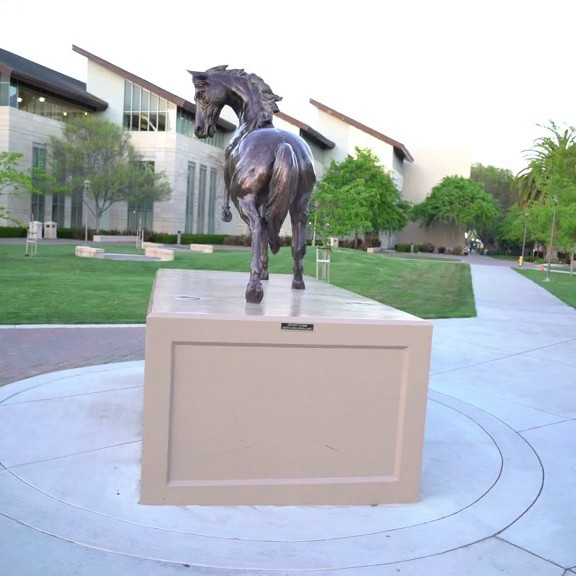}};
\spy [red] on ($(img-20.center) + (-0.10\supimgwidth, 0.15\supimgwidth)$) in node [right] at ($(img-20.east) + (0, 0.52\supspysize)$);
\spy [blue] on ($(img-20.center) + (0.20\supimgwidth, -0.15\supimgwidth)$) in node [right] at ($(img-20.east) + (0, -0.52\supspysize)$);

\node[image, right=of img-20, xshift=1.05\supspysize] (img-21)
  {\includegraphics[width=\supimgwidth]{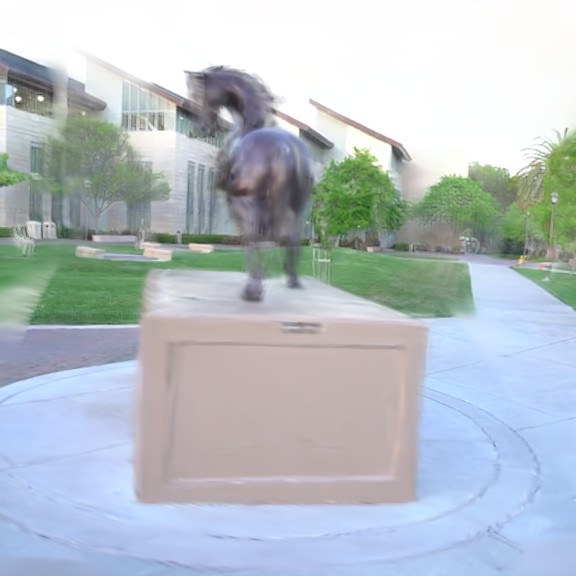}};
\spy [red] on ($(img-21.center) + (-0.10\supimgwidth, 0.15\supimgwidth)$) in node [right] at ($(img-21.east) + (0, 0.52\supspysize)$);
\spy [blue] on ($(img-21.center) + (0.20\supimgwidth, -0.15\supimgwidth)$) in node [right] at ($(img-21.east) + (0, -0.52\supspysize)$);

\node[image, right=of img-21, xshift=1.05\supspysize] (img-22)
  {\includegraphics[width=\supimgwidth]{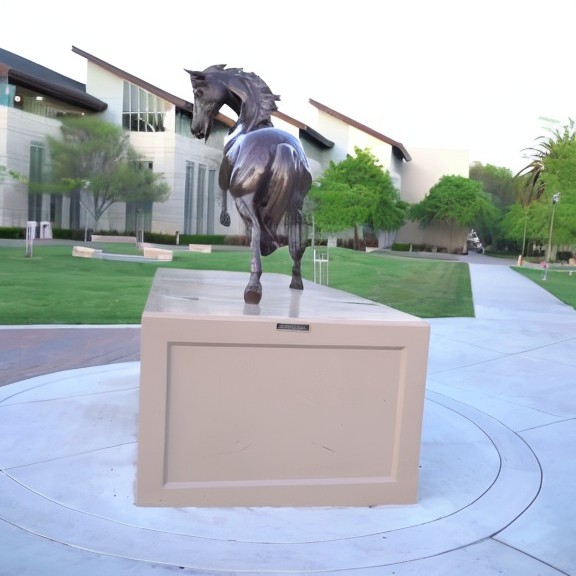}};
\spy [red] on ($(img-22.center) + (-0.10\supimgwidth, 0.15\supimgwidth)$) in node [right] at ($(img-22.east) + (0, 0.52\supspysize)$);
\spy [blue] on ($(img-22.center) + (0.20\supimgwidth, -0.15\supimgwidth)$) in node [right] at ($(img-22.east) + (0, -0.52\supspysize)$);

\node[image, right=of img-22, xshift=1.05\supspysize] (img-23)
  {\includegraphics[width=\supimgwidth]{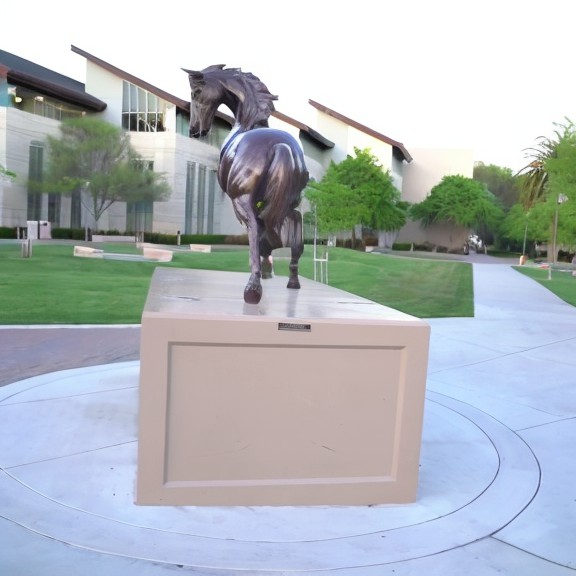}};
\spy [red] on ($(img-23.center) + (-0.10\supimgwidth, 0.15\supimgwidth)$) in node [right] at ($(img-23.east) + (0, 0.52\supspysize)$);
\spy [blue] on ($(img-23.center) + (0.20\supimgwidth, -0.15\supimgwidth)$) in node [right] at ($(img-23.east) + (0, -0.52\supspysize)$);

\node[image, below=of img-20] (img-30)
  {\includegraphics[width=\supimgwidth]{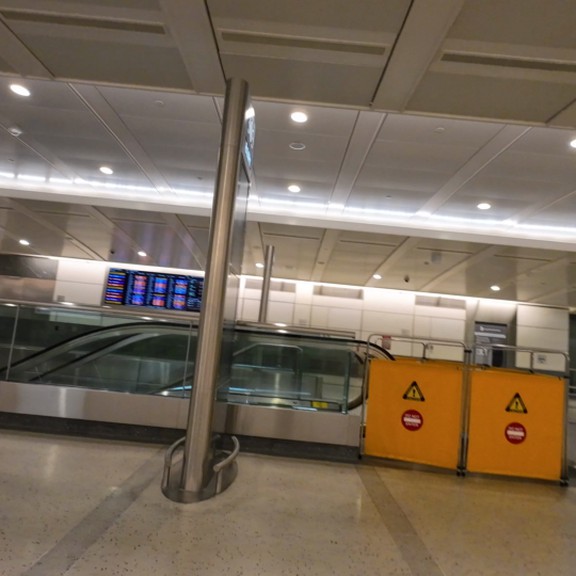}};
\spy [red] on ($(img-30.center) + (-0.15\supimgwidth, 0.15\supimgwidth)$) in node [right] at ($(img-30.east) + (0, 0.52\supspysize)$);
\spy [blue] on ($(img-30.center) + (0.14\supimgwidth, -0.18\supimgwidth)$) in node [right] at ($(img-30.east) + (0, -0.52\supspysize)$);

\node[image, right=of img-30, xshift=1.05\supspysize] (img-31)
  {\includegraphics[width=\supimgwidth]{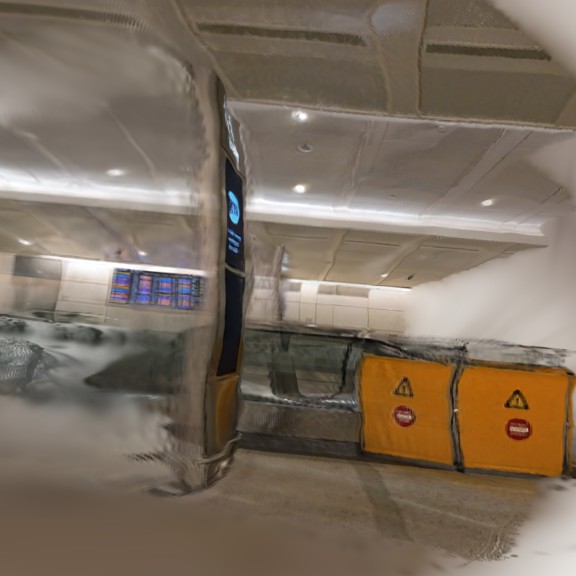}};
\spy [red] on ($(img-31.center) + (-0.15\supimgwidth, 0.15\supimgwidth)$) in node [right] at ($(img-31.east) + (0, 0.52\supspysize)$);
\spy [blue] on ($(img-31.center) + (0.14\supimgwidth, -0.18\supimgwidth)$) in node [right] at ($(img-31.east) + (0, -0.52\supspysize)$);

\node[image, right=of img-31, xshift=1.05\supspysize] (img-32)
  {\includegraphics[width=\supimgwidth]{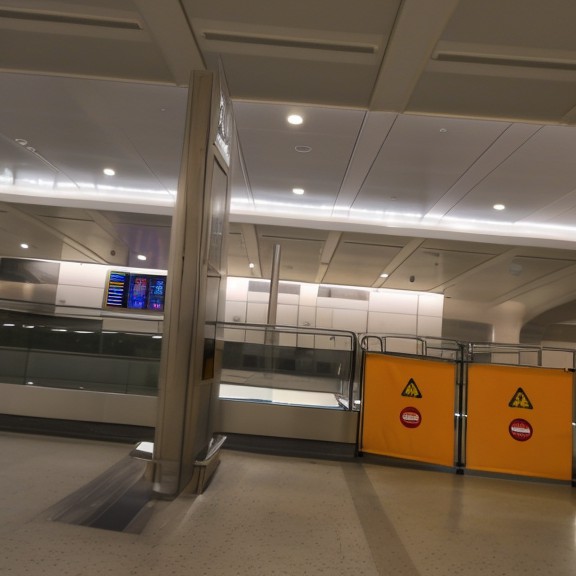}};
\spy [red] on ($(img-32.center) + (-0.15\supimgwidth, 0.15\supimgwidth)$) in node [right] at ($(img-32.east) + (0, 0.52\supspysize)$);
\spy [blue] on ($(img-32.center) + (0.14\supimgwidth, -0.18\supimgwidth)$) in node [right] at ($(img-32.east) + (0, -0.52\supspysize)$);

\node[image, right=of img-32, xshift=1.05\supspysize] (img-33)
  {\includegraphics[width=\supimgwidth]{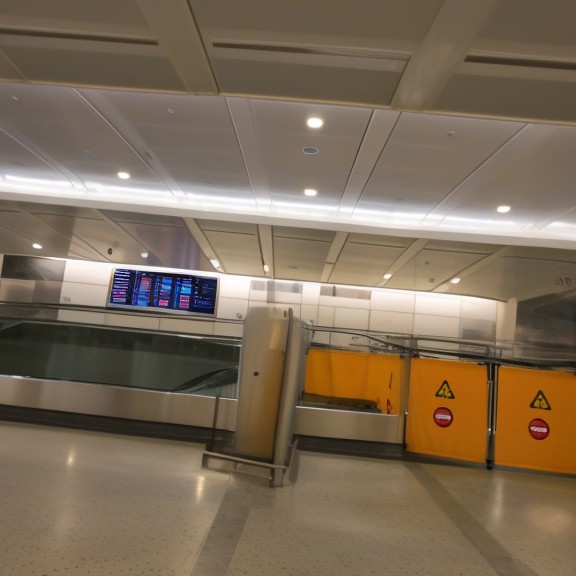}};
\spy [red] on ($(img-33.center) + (-0.15\supimgwidth, 0.15\supimgwidth)$) in node [right] at ($(img-33.east) + (0, 0.52\supspysize)$);
\spy [blue] on ($(img-33.center) + (0.14\supimgwidth, -0.18\supimgwidth)$) in node [right] at ($(img-33.east) + (0, -0.52\supspysize)$);

\node[label, xshift=0.52\supspysize] at (img-00.north) {Ground Truth};
\node[label, xshift=0.52\supspysize] at (img-01.north) {Render};
\node[label, xshift=0.52\supspysize] at (img-02.north) {Ours};
\node[label, xshift=0.52\supspysize] at (img-03.north) {SEVA};

\end{tikzpicture}
\caption{Qualitative Results on the MipNeRF360, Tanks and Temples, and DL3DV Datasets. Red and blue boxes highlight detailed regions with corresponding magnified views showing texture fidelity and geometric accuracy.}
\label{fig:sup_vis}
\end{figure}

We provide more qualitative visualizations of novel view synthesis results across diverse datasets. \cref{fig:sup_vis} shows generation quality on MipNeRF 360, Tanks and Temples, and DL3DV datasets, with red and blue zoom boxes highlighting detailed regions to assess texture fidelity and geometric accuracy.

\section{Inference Time View Selection}

In this section, we provide additional analysis of the inference-time view selection policies introduced in the main paper. Unless otherwise specified, we use 32 candidate reference views and a 6-view context budget ($B{=}6$) at each generation, matching the sparse regime in the main paper.

\paragraph{View-Index Rendering Implementation.}
As described in the main paper, we render a view-index map by downsampling the reconstructed 3D Gaussians and encoding each source view with a distinct palette color. We apply a hard depth filter before color decoding. At each pixel, we keep the closest valid Gaussian and ignore all other Gaussians. Pixels with no valid Gaussian are excluded from voting. Thus, colors from different source views are not blended in the index map. We downsample to approximately 10\% of the original Gaussian count to reduce computational overhead while preserving coarse geometric visibility. The palette colors $\{c_k\}_{k=1}^{V}$ are generated by uniform sampling in HSV color space and converting to RGB, ensuring sufficient perceptual distance between view indices for robust nearest-color lookup during index recovery.

\paragraph{Curated challenging dataset for view selection analysis.}
Standard benchmarks mostly feature simple scene geometry and smooth camera motion, so different selection rules often choose very similar views. To stress-test selection policies, we collect a small dataset of short handheld video sequences in everyday indoor and outdoor environments. From each video we uniformly subsample a fixed number of frames as candidate reference views and select a few target views for evaluation.

The captured scenes exhibit three recurring geometric patterns.
\begin{itemize}
    \item \textbf{Occlusion}: small objects are partially or fully blocked by foreground structures, so that nearby views see only the occluder while slightly different viewpoints reveal the object.
    \item \textbf{Corridor}: long walkways with repeated structures and strong perspective, where many neighboring views become redundant if selected simultaneously.
    \item \textbf{Staircase}: multi-level geometry with railings and noticeable height changes, where small camera shifts significantly alter which surfaces are visible.
\end{itemize}

\begin{table}[!t]
\centering
\small
\caption{Quantitative comparison of view-selection policies on our captured
real-world scenes with a 6-view context budget. Metrics are
averaged over all target views.}
\label{tab:supp_selection_quant}
\setlength{\tabcolsep}{4pt}
\begin{tabular}{lccc}
\toprule
Method   & PSNR$\uparrow$ & SSIM$\uparrow$ & LPIPS$\downarrow$ \\
\midrule
CamDist  & 18.37 & 0.58 & 0.31 \\
Surfel   & \underline{18.52} & \underline{0.59} & \underline{0.30} \\
FoV      & 18.27 & 0.58 & \underline{0.30} \\
Ours     & \textbf{19.30} & \textbf{0.61} & \textbf{0.26} \\
\bottomrule
\end{tabular}
\end{table}

\paragraph{Quantitative comparison on captured scenes.}
\cref{tab:supp_selection_quant} reports a quantitative comparison of selection policies on these captured sequences. On scenes with strong occlusions and depth variation, purely pose-based methods such as CamDist and FoV perform the weakest. The Surfel-based method brings a slight improvement, and our Gaussian-visibility selection achieves the best overall image quality.

\paragraph{Ablation on selection variants on RealEstate10K.}
Our hybrid selector combines Gaussian-visibility scores $S(k)$ with two additional components: \emph{PoseAug}, which fills remaining context budget slots with views ranked by camera distance to the target pose, and \emph{DeDup}, a lightweight spatial-diversity heuristic that discourages selecting nearly-identical views by encouraging coverage across different image regions.

\begin{table}[!t]
\centering
\small
\caption{Ablation of selection variants on RealEstate10K with 32 candidates and a
6-view context budget.}
\label{tab:re10k_selection_ablation}
\setlength{\tabcolsep}{4pt}
\begin{tabular}{lccc}
\toprule
Method                      & PSNR$\uparrow$ & SSIM$\uparrow$ & LPIPS$\downarrow$ \\
\midrule
Ours                        &\textbf{28.25}                &\textbf{0.85 }               &\textbf{0.06 }                   \\
w/o PoseAug                 &26.50                &0.83                &0.08                    \\
w/o DeDup                   &\underline{27.94}                &\underline{0.84}                &\textbf{0.06}                    \\
w/o PoseAug + DeDup         &25.13                &0.81                &0.11                    \\
\bottomrule
\end{tabular}
\end{table}

To isolate the contribution of each component, we conduct an ablation on RealEstate10K with 32 candidates and a 6-view context budget, using the same fixed generator as in the main paper.
\cref{tab:re10k_selection_ablation} shows that both PoseAug and DeDup provide consistent gains. Removing PoseAug alone causes a 1.75~dB drop, while removing both components degrades PSNR by 3.12~dB, confirming that the pose-based fallback and diversity filtering are complementary to visibility-based ranking.

\textit{DeDup spatial filtering mechanism.}
The DeDup component partitions the target image into a $2 \times 2$ grid of tiles and computes tile-restricted visibility scores $S_{\mathrm{vis}}^{(b)}(k)$ for each tile $b$. During greedy selection, we choose views that maximize the marginal gain:
\begin{equation}
\Delta(k) = \lambda_{\mathrm{global}} S_{\mathrm{vis}}(k) + \lambda_{\mathrm{tile}} \sum_{b=1}^{4} \max\{0, S_{\mathrm{vis}}^{(b)}(k) - C_b\}
\end{equation}
where $C_b = \max_{u \in \mathcal{S}} S_{\mathrm{vis}}^{(b)}(u)$ tracks the best current coverage of tile $b$ among already-selected views $\mathcal{S}$. This prevents selecting redundant views that observe the same regions while encouraging spatial diversity across the image.

\section{Inference Cost Analysis}
\label{sec:inference_cost}

\begin{table}[!t]
\centering
\small
\caption{Inference cost on a single RTX 4090 with batch size 1 and a candidate
pool of 32 views. Sampling times for SEVA and for our model are measured under
the same DDIM configuration. Rendering and selection together account for less
than $0.01\%$ of our end-to-end latency. The reconstruction cost is not exclusive
to our method: the unposed SEVA baseline also requires a reconstruction pass to
obtain poses, so the two pipelines differ mainly in sampling time.}
\label{tab:inference_cost}
\setlength{\tabcolsep}{4pt}
\begin{tabular}{llc}
\toprule
Stage & Component & Time (s) \\
\midrule
Reconstruction  & WorldMirror forward pass                & 3.73 \\
\midrule
Selection       & 3DGS rendering (RGB + index map)        & 0.005 \\
                & full selection incl.\ DeDup and PoseAug & 0.006 \\
\midrule
Generation      & SEVA sampling (baseline)                & 62.9 \\
                & SplatGuide sampling (ours)              & 71.8 \\
\midrule
\multicolumn{2}{l}{End-to-end (ours)}                     & 75.5 \\
\bottomrule
\end{tabular}
\end{table}

\cref{tab:inference_cost} reports a stage-wise breakdown of inference cost. Two observations follow.

First, the visibility-aware selector is essentially free. Rendering the view-index map and aggregating votes over a pool of 32 candidates takes 0.006~s in total, four orders of magnitude below the sampling cost, because it reuses the 3DGS scene and the rasterizer already required for pixel-level conditioning and adds no separate 3D data structure. This substantiates the claim that occlusion-aware selection is obtained at negligible cost: the accuracy gains over pose-based and surfel-based policies reported in the main paper are not purchased with compute.

Second, the geometric conditioning itself is not free, and we report its cost explicitly. Sampling rises from 62.9~s to 71.8~s, an increase of 8.9~s or roughly $14\%$, arising from the extra rendered-latent channel group and the additional cross-attention layers that consume reconstruction tokens. The 3.73~s reconstruction pass is the other component of our end-to-end cost, but it is not an overhead unique to our method, since the unposed SEVA baseline likewise depends on a reconstruction pass for pose estimation. We consider the sampling increase a favorable trade: the same reconstruction pass simultaneously supplies pixel-level conditioning, feature-level tokens, and the selection signal, so one forward pass is amortized across all three uses rather than paid for separately.

We report wall-clock time rather than FLOPs because the dominant cost is iterative DDIM sampling, whose latency is governed by the number of sequential denoising steps and is therefore not captured by a single-pass FLOP count. All measurements above fit on a single consumer card, and the selector adds no persistent state beyond the downsampled Gaussian set, which is roughly $10\%$ of the full reconstruction.

\section{ViewCrafter Evaluation Details}

The official ViewCrafter codebase does not provide an evaluation pipeline for multi-view input scenarios. To enable a fair comparison, we explored two reproduction approaches.

\textbf{Approach 1, All-view reconstruction:} We input all reference and target images into DUSt3R~\cite{dust3r_cvpr24} to obtain a complete point cloud and camera poses for all views. Subsequently, we remove the point cloud corresponding to target views while retaining only the reference view point cloud for subsequent point rendering. As shown in~\cref{tab:viewcrafter_comparison}, this approach yields results closely aligned with those reported for ViewCrafter in the SEVA paper, suggesting that SEVA may have adopted a similar evaluation strategy.

However, this approach has a potential issue: although target point clouds are removed after reconstruction, using all views during the reconstruction phase yields more accurate poses and geometry for reference views, particularly in sparse-view scenarios. This introduces information leakage from target views, creating an unfair advantage that does not reflect the true capability of synthesizing novel views from reference views alone.

\textbf{Approach 2, Test-time alignment:} To ensure a fair comparison, we adopt the test-time camera pose alignment strategy, where reconstruction is performed using only reference views. Similar stricter protocol is applied identically to our method and all baselines, and all results reported in the main paper are based on this consistent setup.

Our method outperforms ViewCrafter across all datasets under both evaluation protocols, confirming that the improvements are not artifacts of the evaluation setup.

\begin{table}[!t]
    \centering
    \small
    \caption{Comparison of ViewCrafter results under different evaluation protocols on RealEstate10K benchmark. Here we use 3 input-views as an example.}
    \label{tab:viewcrafter_comparison}
    \begin{tabular}{lccc}
    \toprule
    Method & PSNR $\uparrow$ & SSIM $\uparrow$ & LPIPS $\downarrow$ \\
    \midrule
    ViewCrafter (SEVA) & \underline{22.81} & \underline{0.83} & \underline{0.16} \\
    ViewCrafter  (All-view) & 22.80 & 0.81 & \underline{0.16} \\
    ViewCrafter  (Test-align) & 20.18 & 0.74 & 0.23 \\
    \midrule
    Ours & \textbf{26.52} & \textbf{0.84} & \textbf{0.07} \\
    \bottomrule
    \end{tabular}
\end{table}

\section{Evaluation Dataset Split}
We adhere to established evaluation protocols across all benchmarks to ensure fair comparison. For RealEstate10K and Mip-NeRF 360, we use the test splits defined in ReconFusion~\cite{wu2023reconfusion} and the official release~\cite{barron2022mipnerf360}, respectively.
For Tanks and Temples, we follow ViewCrafter's test scene selection but extend the evaluation from single-input to 3-view and 6-view settings. For DL3DV, we adopt the 20-scene subset introduced in Long-LRM~\cite{ziwen2025llrm}, with strict separation from the training data. The specific DL3DV scenes selected for evaluation are:

\begin{footnotesize}
\begin{itemize}
    \item \texttt{\seqsplit{0bfdd020cf475b9c68e4b469d1d1a2d0cad303eefe8b78fb2307855afdaac8be}}
    \item \texttt{\seqsplit{6d81c5ab0d480fd43d78b75ff372a8113ad38e2c03f1d69627c009883054d4c2}}
    \item \texttt{\seqsplit{8cb2e97d26a639f05a571476240a8fa86988e6853f0f13cc05830d1578002aad}}
    \item \texttt{\seqsplit{093ef327b4e4f9d4ee52c02a354a53558a8652157fb0d58f3b4a708734afb334}}
    \item \texttt{\seqsplit{119fd56d3797e2d349ca64ddcc5851463cd13b5974b5b2e4566ed5cf7e02e6c1}}
    \item \texttt{\seqsplit{165f5af8bfe32f70595a1c9393a6e442acf7af019998275144f605b89a306557}}
    \item \texttt{\seqsplit{183dd248f6a86e07c5adf9de8ee2d0abe45b1216331c03678e89634c2e9b1c7f}}
    \item \texttt{\seqsplit{0569e83fdc248a51fc0ab082ce5e2baff15755c53c207f545e6d02d91f01d166}}
    \item \texttt{\seqsplit{918c8dad730c3b804306c5da8486124be4aa0612e85fb825338fd350c912e1b0}}
    \item \texttt{\seqsplit{8324b3ca22085040c2a0ecb7284e0cdf776b1f846b73a7c0df893587cb4a45f8}}
    \item \texttt{\seqsplit{35317e621976e87f0c143e66fc61fb8cddb4ff134304da7a00e32ac1983105b4}}
    \item \texttt{\seqsplit{35872363e17af5d173b6a0b09fcf5de94627ad5dc5f8a9ad4c579f3e70b4797a}}
    \item \texttt{\seqsplit{41036716da7efda334c1d434c4141d15642e0e02f881a01b6c8c36f8bea64c45}}
    \item \texttt{\seqsplit{493816813d2d6d248eb3c2b0b77b63e54235266e9a06e270fd0d282f13960493}}
    \item \texttt{\seqsplit{0853979305f7ecb80bd8fc2c8df916410d471ef04ed5f1a64e9651baa41d7695}}
    \item \texttt{\seqsplit{1264931635e127fb905c8953cbc2deadd0c763e633af7fbd9405a61ca849710c}}
    \item \texttt{\seqsplit{a17a984ca90a9b5840fdf85b15104b0d18e25975981c1aa90fcdfd6eeeb285f3}}
    \item \texttt{\seqsplit{a62c330f5403e2e41a82a74c4e865b705c5706843b992fae2fe2e538b122d984}}
    \item \texttt{\seqsplit{adf35184a12d4cfa3f4248b87aa5adb4f39f179df460d6d76136e13d37299a2a}}
    \item \texttt{\seqsplit{e5684b3292bfd77db297839fc37ee4cce7fd59775af1a6a4827e3b4f59c036d3}}
\end{itemize}
\end{footnotesize}

\bibliographystyle{splncs04}
\bibliography{main}